%% file: main.tex
\documentclass[sigconf]{acmart}

\copyrightyear{2023}
\acmYear{2023}
\setcopyright{rightsretained}
\acmConference[CCS '23]{Proceedings of the 2023 ACM SIGSAC Conference on Computer and Communications Security}{November 26--30, 2023}{Copenhagen, Denmark}
\acmBooktitle{Proceedings of the 2023 ACM SIGSAC Conference on Computer and Communications Security (CCS '23), November 26--30, 2023, Copenhagen, Denmark}\acmDOI{10.1145/3576915.3616605}
\acmISBN{979-8-4007-0050-7/23/11}
\usepackage{fancyhdr}
\usepackage{algorithmic}
\usepackage{amsmath,amsthm}
\usepackage{balance}
\usepackage{booktabs}
\usepackage{caption}
\usepackage{enumitem}
\usepackage{graphicx}
\usepackage{multicol}
\usepackage{multirow}
\usepackage{caption}
\usepackage{subcaption}
\usepackage{threeparttable}
\usepackage{tikz}
\usepackage{url}
\usepackage{xcolor}
\usepackage{xspace}
\usepackage{bbding}
\usepackage{float}
\usepackage{makecell}
\usepackage[]{footmisc}
\usepackage{appendix}

\usetikzlibrary{arrows,shapes,positioning,shadows,trees}

\usepackage{grffile}

\usepackage{tikz}
\usetikzlibrary{positioning, shapes.geometric}

\tikzset{
box/.style ={
rectangle, 
rounded corners =5pt, 
minimum width =25pt, 
minimum height =22pt, 
inner sep=4pt, 
draw=black 
}}

\newcommand{\ignore}[1]{{}}

\newcommand{\eg}{\textit{e.g.}\xspace}
\newcommand{\ie}{\textit{i.e.}\xspace}
\newcommand{\etal}{\textit{et al.}\xspace}
\newcommand{\etc}{\textit{etc}\xspace}
\newtheorem{mydef}{Definition}

\begin{document}

\title{Good-looking but Lacking Faithfulness: Understanding Local Explanation Methods through Trend-based Testing}

\author{Jinwen He}
\email{hejinwen@iie.ac.cn}
\affiliation{
  \institution{SKLOIS, IIE, CAS$^{\dag}$}
  \institution{School of Cyber Security, UCAS$^{\ddag}$}
  \city{Beijing}
  \country{China}
}

\author{Kai Chen$^*$}
\email{chenkai@iie.ac.cn}
\affiliation{
  \institution{SKLOIS, IIE, CAS$^{\dag}$}
  \institution{School of Cyber Security, UCAS$^{\ddag}$}
  \city{Beijing}
  \country{China}
}

\author{Guozhu Meng}
\email{mengguozhu@iie.ac.cn}
\affiliation{
  \institution{SKLOIS, IIE, CAS$^{\dag}$}
  \institution{School of Cyber Security, UCAS$^{\ddag}$}
  \city{Beijing}
  \country{China}
}

\author{Jiangshan Zhang}
\email{zhangjiangshan@iie.ac.cn}
\affiliation{
  \institution{SKLOIS, IIE, CAS$^{\dag}$}
  \institution{School of Cyber Security, UCAS$^{\ddag}$}
  \city{Beijing}
  \country{China}
}

\author{Congyi Li}
\email{licongyi@iie.ac.cn}
\affiliation{
  \institution{SKLOIS, IIE, CAS$^{\dag}$}
  \institution{School of Cyber Security, UCAS$^{\ddag}$}
  \city{Beijing}
  \country{China}
}

\thanks{$*$ Corresponding author}

\thanks{${\dag}$ Institute of Information Engineering,  Chinese Academy of Sciences}
\thanks{${\ddag}$ University of Chinese Academy of Sciences}

\renewcommand{\shortauthors}{Jinwen He et al.}

\begin{abstract}
While enjoying the great achievements brought by deep learning (DL), people are also worried about the decision made by DL models, since the high degree of non-linearity of DL models makes the decision extremely difficult to understand. Consequently, attacks such as adversarial attacks are easy to carry out, but difficult to detect and explain, which has led to a boom in the research on local explanation methods for explaining model decisions. In this paper, we evaluate the faithfulness of explanation methods and find that traditional tests on faithfulness encounter the random dominance problem, \ie, the random selection performs the best, especially for complex data. To further solve this problem, we propose three trend-based faithfulness tests and empirically demonstrate that the new trend tests can better assess faithfulness than traditional tests on image, natural language and security tasks. We implement the assessment system and evaluate ten popular explanation methods.
Benefiting from the trend tests, we successfully assess the explanation methods on complex data for the first time, bringing unprecedented discoveries and inspiring future research. Downstream tasks also greatly benefit from the tests. For example, model debugging equipped with faithful explanation methods performs much better for detecting and correcting accuracy and security problems. 
\end{abstract}



\maketitle

\input{sections/introduction}

\input{sections/background}
\input{sections/evaluation2}
\input{sections/measurement}
\input{sections/application}
\input{sections/relatedwork}
\input{sections/discussion}
\input{sections/conclusion}

\section*{acknowledgements}
We thank all the anonymous reviewers for their constructive feedback.
The IIE authors are supported in part by NSFC (92270204), Beijing Natural Science Foundation (No.M22004), Youth Innovation Promotion Association CAS, Beijing Nova Program, a research grant from Huawei and the Anhui Department of Science and Technology under Grant 202103a05020009.

\bibliographystyle{ACM-Reference-Format}
\balance
\bibliography{ref}

\renewcommand\thesection{\Alph{subsection}}
\input{sections/appendix}
\end{document}

%% file: sections/introduction.tex
\section{Introduction}\label{sec:introduction} 

In the past ten years, with rapid advances in the field of deep learning (DL), data-driven approaches have drawn lots of attention. They have made great progress in many fields, including computer vision~\cite{lenet5, resnet}, speech recognition~\cite{seq2seq, bilstm}, natural language processing~\cite{birnn, transformer}, etc. One of the main benefits of data-driven approaches is that, without needing to know a theory, a machine learning algorithm can be used to analyze a problem using data alone. However, on the other side of the coin, DL models are hard to explain without the theory. Neither can researchers understand why the DL models make a decision. A well-known problem is adversarial examples (AEs), which mislead a DL model by adding human-imperceptible perturbations to the natural data~\cite{FGSM}. These perturbations are imperceptible by humans, but impact the decision of the model.  
\begin{figure}
    \setlength{\abovecaptionskip}{4pt}
    \setlength{\belowcaptionskip}{0pt}
    \centering
	\includegraphics[width=0.45\textwidth, trim={10 0 0 0},clip]{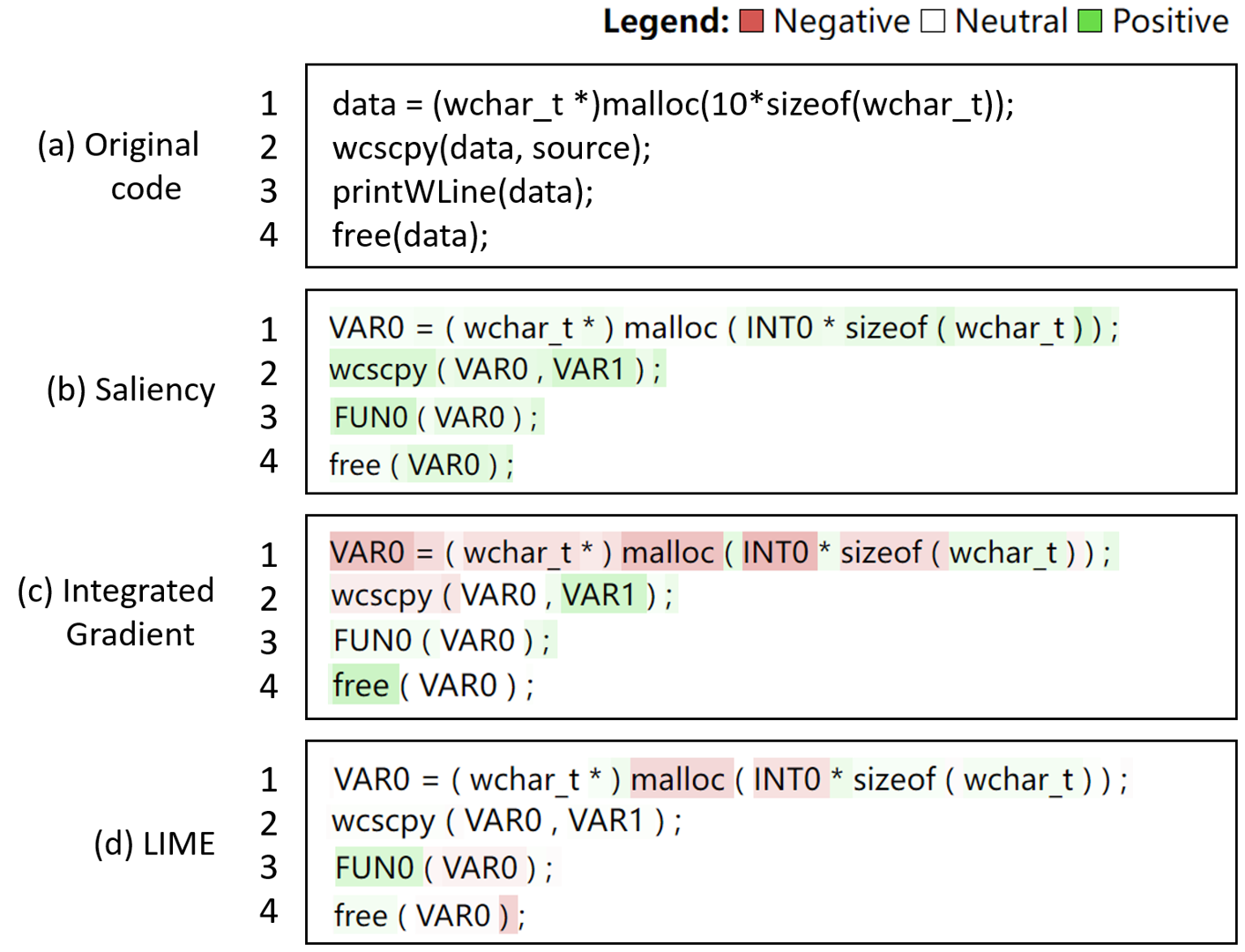}
	\caption{The importance of words identified by three explanation methods. The darker the color, the higher the contribution score.}\label{fig:demo_figure_1}
    \vspace{-4mm}
\end{figure}
To fill the gap between model decisions and human cognition, researchers develop various techniques to explain the prediction results~\cite{Grad, DeepLIFT, LIME}. Obviously, an ideal technique should explain a model’s predictions in a \textit{human-understandable} and \textit{model-faithful} manner~\cite{humanstudy,BAM}. That is, the explanation should be meaningful to humans and correspond to the model's behavior in the vicinity of the instance being predicted. 
The risks of deep learning models further propel the advance of explanation methods, which are popularly used to build secure and trustworthy models~\cite{ieeexaisurvey}, such as model debugging~\cite{modeldebug,DSL}, understanding attacks~\cite{interpretadversarialattack, explanationpoison} and defenses~\cite{explanationadversarialrobustness} of DL models.

In this paper, we compare popular local explanation methods theoretically and experimentally. Specifically, we implement ten typical methods for comparison. 
Figure~\ref{fig:demo_figure_1} compares the results of Saliency map~\cite{Grad}, Integrated Gradient~\cite{IG} and LIME~\cite{LIME} on a vulnerability detection model trained with VulDeePecker dataset~\cite{Vuldeepecker}. 
The contribution of ``wcscpy'' in the second line differs among the three explanation methods. In Figure~\ref{fig:demo_figure_1}(b), ``wcscpy'' has positive contribution, while in Figure~\ref{fig:demo_figure_1}(c), ``wcscpy'' has negative contribution. In Figure~\ref{fig:demo_figure_1}(d), ``wcscpy'' has almost no contribution. It is observed that the similarity between the results of different explanation methods is small.
Thus, it is highly needed to assess the faithfulness of explanation methods, which is also highly challenging.
The main difficulty lies in the lack of ground truth, where contemporary assessments cannot accurately determine the consistency of the explanation with model prediction. 
Most of these methods rely on the hypothesis to assess explanations that the perturbations imposed to more important features can positively make a larger change to the model prediction. 
However, this hypothesis suffers from one significant limit, undermining the faithfulness assessment. This limit is dubbed as \emph{random dominance}.

\begin{figure}
    \setlength{\abovecaptionskip}{4pt}
    \setlength{\belowcaptionskip}{0pt}
    \centering
	\includegraphics[width=0.46\textwidth]{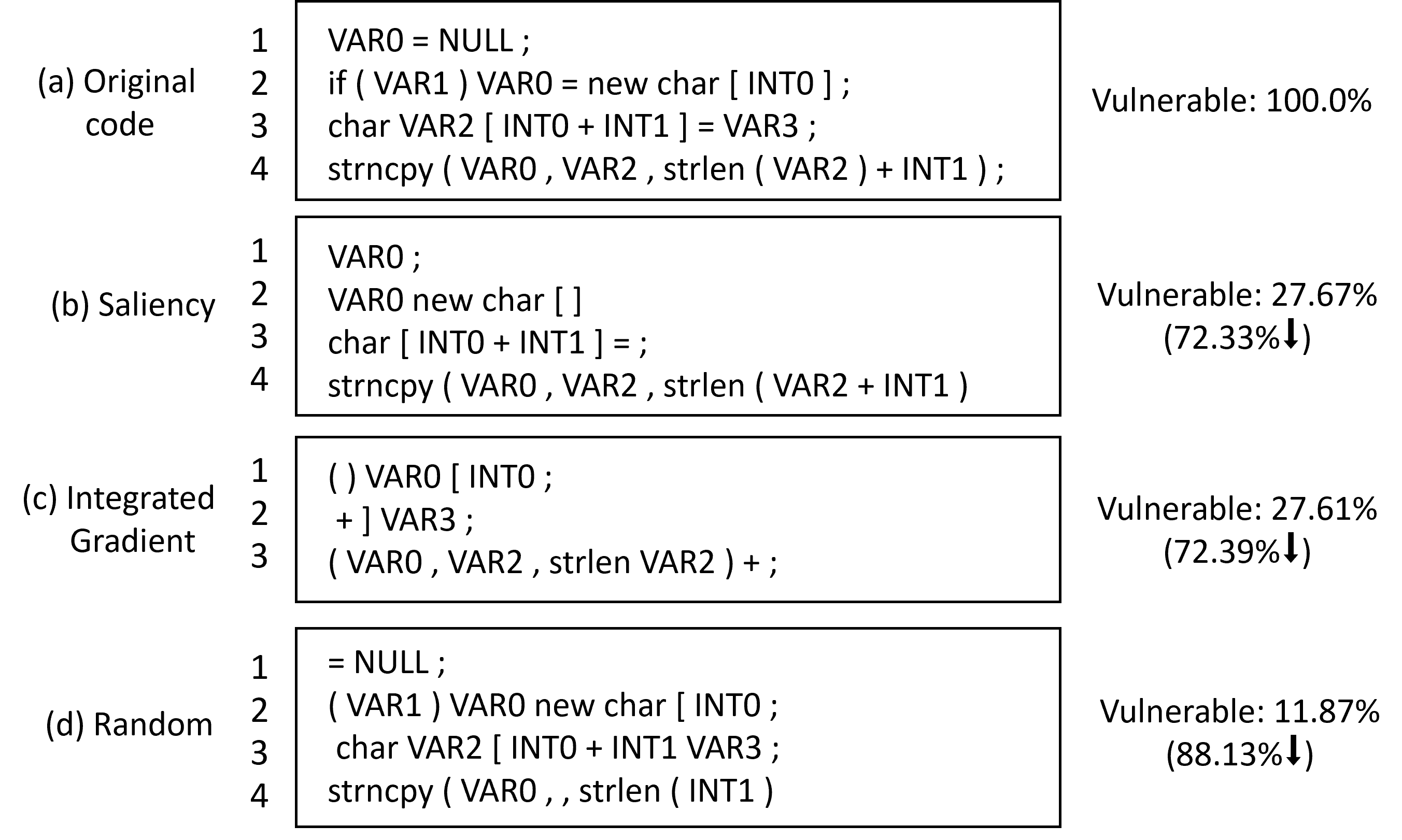}
	\caption{The percentage of score decline after removing 20\% of the most important or randomly selected words. The random method shows the most significant drop in the prediction score.}\label{fig:demo_figure_2}
	\vspace{-4mm}
\end{figure}

\noindent\textbf{Random dominance in model explanation.} 
Take the assessment method-- feature reduction~\cite{RealtimeSaliency,LEMNA,EuroSPSurvey,appofinterpretability} as an example, where the difference in prediction scores is measured when important features of the input are deleted. 
In Figure~\ref{fig:demo_figure_2}, deleting the outputs by Saliency (Figure~\ref{fig:demo_figure_2}(b)) lowers the prediction score by 72.33\%, and deleting Integrated Gradient's output (Figure~\ref{fig:demo_figure_2}(c)) reduces the score by 72.39\%. 
Figure~\ref{fig:demo_figure_2} shows the remaining features after removing important features. From the results, the important features tagged by the two methods are very different, but the prediction scores drop a lot for both methods. Surprisingly, if we randomly delete 20\% of the input (Figure~\ref{fig:demo_figure_2}(d)), the score can be reduced by 88.13\%, even larger than the two explanation methods. The random method can never be a good explanation.

To solve the problem, we design three new trend tests for explanation assessment: the evolving-model-with-backdoor test (EMBT), partial-trigger test (PTT), and evolving-model test (EMT). 
Instead of destructing important features, we gradually evolve either a model or a sample, and form a series of test pairs $\langle model, sample\rangle$. It enables the models and samples to stay in distribution since the model can continuously learn from the samples during evolution, and the evolution of samples is limited within the cognition scope of the model. 
We employ the probability and loss function as an indicator to quantify model behaviors and then calculate the correlation with explanation results. 
Based on these trends, we perform extensive evaluations and analysis of various explanation methods through trend tests and traditional tests. Specifically, we explore the following research questions: 
\begin{itemize}[leftmargin=*]
\item \textbf{RQ1:} How well do the traditional tests work? What are the advantages of trend tests over traditional tests? (See Section~\ref{subsec:traditionvstrend})
\item \textbf{RQ2:} What factors affect the faithfulness of explanation methods?  (See Section~\ref{subsec:influencingfactors})
\item \textbf{RQ3:} Do downstream applications such as model debugging work better when using the explanation method chosen by trend tests? (See Section~\ref{sec:application})
\end{itemize}

Through the evaluation, we have the opportunity to assess the explanation methods and gain unprecedented findings. We find that all explanation methods seem to be unable to handle complex data, as indicated by traditional assessment tests. However, our newly designed tests report that some methods (e.g., Integrated Gradient~\cite{IG} and Integrated SmoothGrad-Squared~\cite{IG,SmoothGrad}) can work well. The reason is mainly due to the random dominance problem existing in the traditional tests, which leads to the wrong results of the evaluation report. 
Furthermore, model complexity seems less important to the explanation methods' faithfulness than data complexity; but the parameters used by the explanation methods are essential. Some researchers are in favor of the parameters that can generate more explainable features (to humans) but ignore faithfulness. 
Our trend tests can address this problem by suggesting the most suitable parameters from candidate ones, resulting in the best faithfulness. Moreover, trend tests are applicable to multiple types of models for various tasks, such as images, natural language, security applications, etc. Finally, we demonstrate the effectiveness of trend tests using a popular downstream application, model debugging. For a given DL model, trend tests recommend explanation methods with higher faithfulness to better debug the model, making it secure and trustworthy.

\noindent \textbf{Contributions.} Our main contributions are as follows:
\begin{itemize}[leftmargin=*] 
\item We develop three novel trend tests (EMBT, PTT, and EMT) to handle the random dominance problem. They are experimentally proven to be effective in measuring the faithfulness of an explanation method and getting rid of the random dominance problem. All the code and extra analysis are released for further research: https://github.com/JenniferHo97/XAI-TREND-TEST. 
\item Through the experiments, we identify the limitations of previous assessment methods and quantify the influence of multiple factors (\ie, data complexity, model complexity, parameters) over explanation results.
\item We demonstrate that trend tests can recommend more faithful explanation methods for model debugging and thus better detect spurious correlations in DL models. 
\end{itemize}

%% file: sections/background.tex
\section{Background}

\subsection{Explanation on DNN}

The high degree of non-linearity of DL models makes it difficult to understand the decision, so security cannot be guaranteed~\cite{FGSM}.
Such dilemma motivates research on explanation techniques for DL models~\cite{interpretvulnerabilitydetection,interpretvuldetectionmodels}, aiming to explain DL models’ decisions~\cite{modeldebug} and understand adversarial attacks~\cite{robustnessanalysisondl,interpretadversarialattack} as well as defenses~\cite{robustcodegeneration}, thereby paving the way for building secure and trustworthy models.
Explanation methods can be categorized as global explanation and local explanation in terms of the analysis object~\cite{ieeexaisurvey}.
In this paper, we focus on local explanation methods.
Without loss of generality, we define the explanation method for input as follows.
\begin{mydef} \textbf{(Local Explanation)}
	Given a model $\mathcal{F}:\mathcal{X}\rightarrow{\mathcal{Y}}$, an explanation method $I: (\mathbf{X}, \mathcal{F})\rightarrow{\mathbf{\phi}}$. For any test input $\mathbf{X}$, the explanation method gives the importance score $\mathbf{\phi}$ for each feature of $\mathbf{X}$, where $\mathbf{\phi}$ has the same dimensions as $\mathbf{X}$.
\end{mydef}

Assuming that $\{x_1, x_2, \cdots, x_n\}$ is the feature set of instance $\textbf{X}$ and $\{\phi_1, \phi_2, \cdots, \phi_n\}$ is the importance score set of the explanation $\phi$, $x_i$ is important for the explanation if $\phi_i\geq \epsilon$ where $1\leq i \leq n$ and $\epsilon$ is often empirically configured. 
Local explanation methods can be either white-box or black-box methods. If one explanation method is dependent on the hyper-parameters and weights of the model,
it is a white-box method. Otherwise, it is a black-box method. Saliency map~\cite{Grad} is a typical white-box method, which computes gradients of the input. 
Although simple and easy to implement, the Saliency map suffers from the gradient saturation problem and is sensitive to noise. Integrated gradient (IG)~\cite{IG} moderates the gradient saturation problem by considering the straight-line path from the baseline to the input and computes the gradients at all points along the path. 
SmoothGrad~\cite{SmoothGrad} tries to reduce the sensitivity of the gradient by adding Gaussian noise to the input and then calculating the average of the gradients. SmoothGrad-Squared (SG-SQ)~\cite{retrain}, VarGrad (VG)~\cite{vargrad} and Integrated SmoothGrad-Squared (SG-SQ-IG)~\cite{retrain} are common variants of the above methods.
Deep Learning Important FeaTures (DeepLIFT)~\cite{DeepLIFT} alleviates the gradient saturation by using the difference between the input and the reference point to explain the importance of input features.
The black-box methods are perturbation-based. Kernel SHAP~\cite{SHAP} and LIME~\cite{LIME} mutate the input randomly. LIME~\cite{LIME} leverages superpixel segmentation~\cite{slicsuperpixels} to improve efficiency in image tasks. Occlusion~\cite{Occlusion} uses a moving square to generate perturbed input. Occlusion~\cite{Occlusion} directly uses the target classification probability as the metric. The lower the probability caused by the mutated input, the more important the features. Based on the local linearity assumption of the neural network decision boundary, LIME~\cite{LIME} trains a surrogate linear model using the perturbed data and labels. The weights of the linear model reflect the importance of the feature.
SHAP~\cite{SHAP}, derived from cooperative game theory, calculates Shapley values as feature importance.

\subsection{Relationship between explanations, models and humans}
An explanation system usually includes the interaction between explanation methods, models, and humans. Prior work that assesses faithfulness falls into two types: human-understandable and model-faithful. The human-understandable assessments focus on the correlation between explanation methods and human cognition~\cite{humanstudy}. Unfortunately, explanation methods cannot reveal all the knowledge learned by the model precisely. Therefore, it has not yet reached the stage where we can assess the correlation between explanation methods and human cognition.
Under such circumstances, we should evaluate the explanation methods in a model-faithful way. The model-faithful assessments focus on the correlation between the explanation method and the model~\cite{LEMNA}. A common way is to mask some important features tagged by the explanation method and then observe the decline in the model prediction probability. The more the probability decreases, the more important the masked features are. However, randomly masking some features may also cause a significant decrease in model prediction probability. We refer to this as the random dominance problem. To overcome this problem, we propose trend tests, which use in-distribution data and are applicable in more scenarios.
After the model-faithful assessment, the user can select a faithful explanation method to explain the model, fix the bias and improve the security and trustworthiness of the model. Ultimately, consistency in explanation methods, model decisions, and human cognition can be achieved.

%% file: sections/evaluation2.tex
\section{Design of Faithfulness Tests}\label{sec:evaluation}

In this section, we first provide a high-level definition of faithfulness. Then we briefly introduce traditional evaluation methods and design three trend-based tests, \ie, the evolving-model-with-backdoor test (EMBT), partial-trigger test (PTT), and evolving-model test (EMT), to assess the faithfulness of explanation methods.

\subsection{Problem Definition}

\begin{figure}
    \setlength{\abovecaptionskip}{4pt}
    \setlength{\belowcaptionskip}{0pt}
    \centering
	\includegraphics[width=0.35\textwidth]{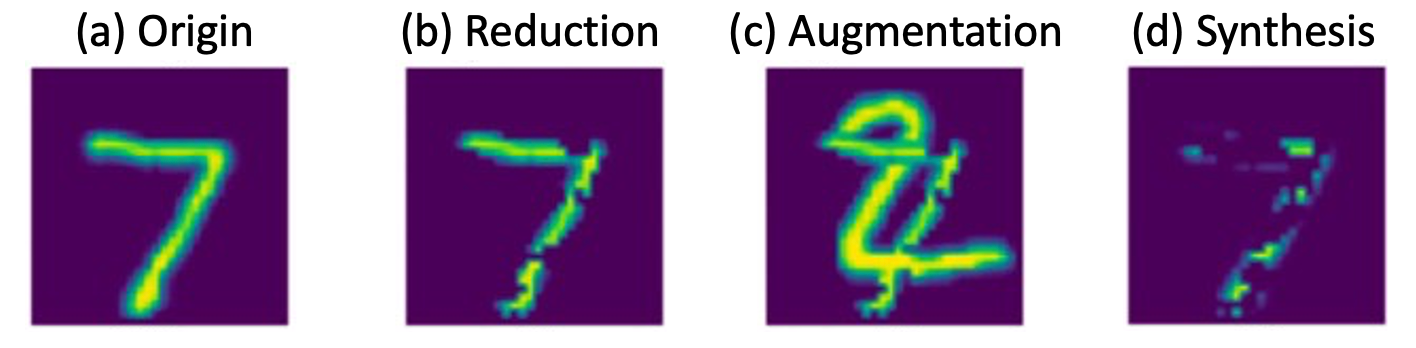}
	\caption{Examples of synthesis test, augmentation test and reduction test. Features with the top 10\% importance scores tagged by the explanation method are important features.}\label{fig:test_vis}
	\vspace{-4mm}
\end{figure}

A local explanation is faithful if its identified features in the input are what the model relies on for making the decision. However, it is non-trivial to evaluate the faithfulness of an explanation method, as indicated in Figure~\ref{fig:demo_figure_1} and ~\ref{fig:demo_figure_2}. 
The formal definition of faithfulness varies across studies~\cite{LEMNA,stablexai}. In this section, we first review the definition of the traditional faithfulness tests from previous work and then present our new trend tests in the next section. Below we use $\rho$ to denote faithfulness.
\vspace{2mm}
\noindent\textbf{Traditional Faithfulness Tests.}
There are three common tests for the local explanation, \ie, synthesis test, augmentation test and reduction test~\cite{LEMNA, EuroSPSurvey}. These tests are widely used as SOTA methods in recent research~\cite{stablexai,robustnessxai}.
The intuition of these tests is to modify an input guided by explanation results and observe the change of the target label's posterior probability by the model, \ie, $\mathcal{F}_{target}(\mathbf{X})$.
In the synthesis test, we only retain the important features $\hat{\mathbf{X}}$ (\ie, $\{x_i ~| ~\phi_i \geq \epsilon \}$) of the test sample $\mathbf{X}$ marked by the explanation methods and add them into an all-black image $\mathbf{X'}$ to form a synthetic sample. Then the difference of target label scores between the synthesis test sample and the all-black image could indicate the faithfulness of the explanation methods, denoted by  $\rho_{syn}$. This can be computed as: $\rho_{syn}(\mathcal{F}, I) = \mathcal{F}_{target}(\mathbf{X'} \oplus\hat{\mathbf{X}}) - \mathcal{F}_{target}(\mathbf{X'})$,
where $\oplus$ denotes element-wise addition. Figure~\ref{fig:test_vis}(a) and (b) show an example of the original test sample and the corresponding synthesis test sample, respectively. Intuitively, $\rho_{syn}$ will increase after important features are added to the all-black image.
In the augmentation test, we randomly select an augmentation sample $\mathbf{X''}$ with a different label from the test samples from the test set. Then we add $\hat{\mathbf{X}}$ to the augmentation sample (see Figure~\ref{fig:test_vis}(c)) and observe the change of the prediction score: $\rho_{aug}(\mathcal{F}, I) = \mathcal{F}_{target}(\mathbf{X''} \oplus \hat{\mathbf{X}}) - \mathcal{F}_{target}(\mathbf{X''})$. 
If important features are accurately recognized, $\rho_{aug}$ is expected to increase.
In the reduction test, we remove important features from the test sample (see Figure~\ref{fig:test_vis}(d)) and observe: $\rho_{red}(\mathcal{F}, I) = \mathcal{F}_{target}(\mathbf{X} \ominus \hat{\mathbf{X}}) - \mathcal{F}_{target}(\mathbf{X})$.
where $\ominus$ denotes element-wise subtraction. In the reduction test, $\rho_{red}$ is expected to decrease if the explanation method accurately tags the important features. 

\subsection{Trend-based Faithfulness Tests}\label{sec:trendtests}

The main problem of traditional tests is the random dominance phenomena, which makes the random baseline too high and invalidates the tests. To solve this problem, we design three trend tests. The intuition is: instead of using features to mutate samples, we generate a set of samples with a certain ``trend'' with natural and backdoor data. Then we let the explanation methods mark important features and check whether the features follow the trend. By measuring the correlation, we can assess the faithfulness of explanation methods. 



\begin{figure}
    \setlength{\abovecaptionskip}{5pt}
    \setlength{\belowcaptionskip}{0pt}
    \centering
	\includegraphics[width=0.4\textwidth]{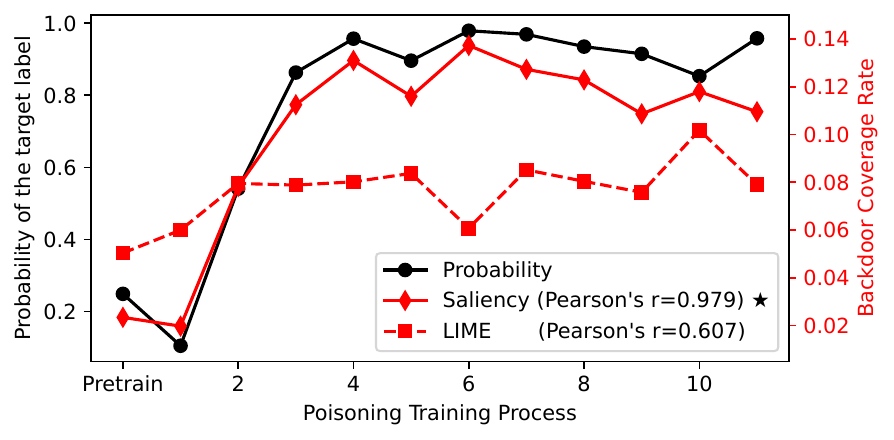}
	\caption{EMBT example. Saliency's backdoor coverage is more related to ResNet18's backdoor accuracy, with a PCC of 0.979, while LIME gets a lower PCC of 0.607.}\label{fig:embt_vis}
	\vspace{-4mm}
\end{figure}

\vspace{2mm}
\noindent\textbf{Evolving-Model-with-Backdoor Test (EMBT)}
To explain a given model, EMBT adds a backdoor to the pre-trained model through incremental training and records the intermediate models in the training process~\cite{badnets}. The probability of the backdoor attack's target label forms a trend. We assume that the model learns at least some of the backdoor features. During backdoor training, the explanation results should show a trend of paying more attention to the location of the backdoor features.
EMBT records the intermediate model in every $c$ epochs during training. 
Then we get a set of models $\mathbf{M}=\{M_0,\dots, M_n\}$. The model $M_0$ is the pre-trained clean model, and $M_i$ is the intermediate model generated in the epoch $i\times{c}$. For a given input, EMBT stamps the trigger on the input and measures the probability of the target label on $\mathbf{M}$. Suppose the result is $\mathbf{P}_{target}=\{P_0, P_1, \dots , P_n\}$. 
The black line in Figure~\ref{fig:embt_vis} shows how the probability of the target label changes during the poisoning training. 
Later, EMBT uses an explanation method to mark the important features on $\mathbf{M}$. For each model $M_i$, we can calculate the overlapped features (denoted as $o_i$) between the important features and the backdoor trigger features (denoted as $t$). We calculate the trigger coverage $s_i=|o_i|/|t|$. For the $n+1$ models, we could generate a sequence $\mathbf{S}=\{s_0,...,s_n\}$. 
For example, in Figure~\ref{fig:embt_vis}, the solid red line shows the sequence. In this way, we use the two trends $\mathbf{P}_{target}$ and $\mathbf{S}$ to evaluate the faithfulness of the explanation method. 

To measure the correlation between two trends, we employ the Pearson correlation coefficient (PCC)~\cite{pearson}, which is known for calculating the correlation between two variables. PCC is also widely used in the field of deep learning to measure the consistency between the two trends~\cite{pccdl1,pccdl2}.
So we calculate PCC:
$$\rho(\mathbf{P}_{target},\mathbf{S})=\frac{cov(\mathbf{P}_{target},\mathbf{S})}
{\sigma_{\mathbf{P}_{target}}\sigma_{\mathbf{S}}},$$
where $cov(\mathbf{P}_{target}, \mathbf{S})$ denotes covariance between $\mathbf{P}_{target}$ and $\mathbf{S}$. $\sigma_{\mathbf{P}_{target}}$ and $\sigma_{\mathbf{S}}$ denotes standard deviation of $\mathbf{P}_{target}$ and $\mathbf{S}$, respectively.
A high value of $\rho(\mathbf{P}_{target},\mathbf{S})$ shows the two trends are consistent, which demonstrates the explanation results are faithful. 
For example, we feed backdoor data to the recorded intermediate models and get explanations with Saliency and LIME, respectively. Then the backdoor coverage rate of the top 10\% important features is calculated. The solid red line in Figure~\ref{fig:embt_vis} shows the change in the backdoor coverage rate of Saliency during poisoning training, and the PCC between the solid red line and the black line is 0.979. The other dotted red line shows the change in the backdoor coverage rate of LIME, while the PCC between the dotted red line and the black line is 0.607. We can also see from Figure~\ref{fig:embt_vis} that the solid red line is more similar to the black line than the dotted red line, indicating that PCC correctly reflects the correlation between the two trends. Note that Figure~\ref{fig:embt_vis} only shows an example. We also perform a detailed evaluation of other explanation methods in Section~\ref{sec:measurement}. The effectiveness of EMBT is based on the assumption that the model learns at least some of the backdoor features, which can be supported by backdoor inversion methods~\cite{neuralcleanse,uncovertrigger}. Therefore, we recommend choosing backdoors that have been proven to be reversible by backdoor defense methods, such as BadNets~\cite{badnets}. We evaluate the effects of different backdoor triggers in Section~\ref{subsec:traditionvstrend}.

\begin{figure}
    \setlength{\abovecaptionskip}{5pt}
    \setlength{\belowcaptionskip}{0pt}
    \centering
	\includegraphics[width=0.4\textwidth]{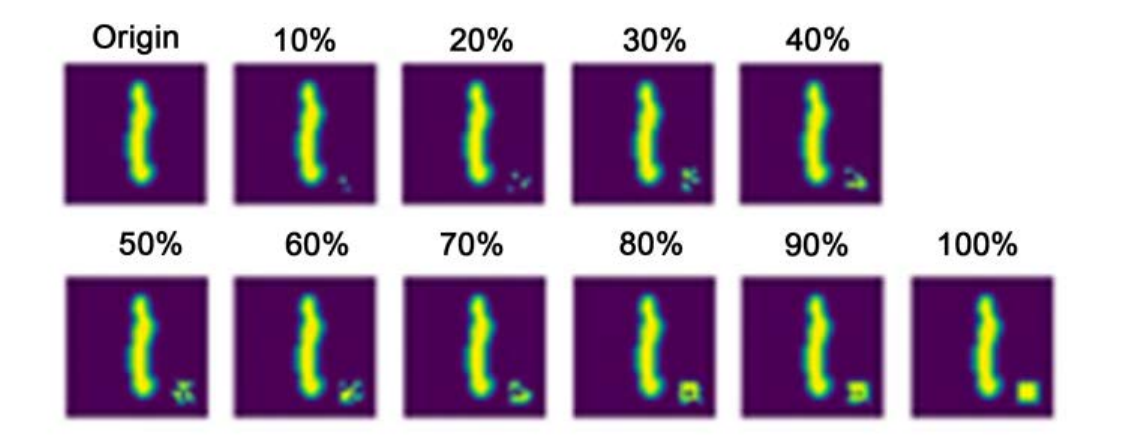}
	\caption{Examples of PTT data sequence, made from 10\% to 100\% of the trigger features covered on a clean sample.}\label{fig:ptt_data_vis}
	\vspace{-4mm}
\end{figure}

\begin{figure}
    \setlength{\abovecaptionskip}{5pt}
    \setlength{\belowcaptionskip}{0pt}
    \centering
	\includegraphics[width=0.4\textwidth]{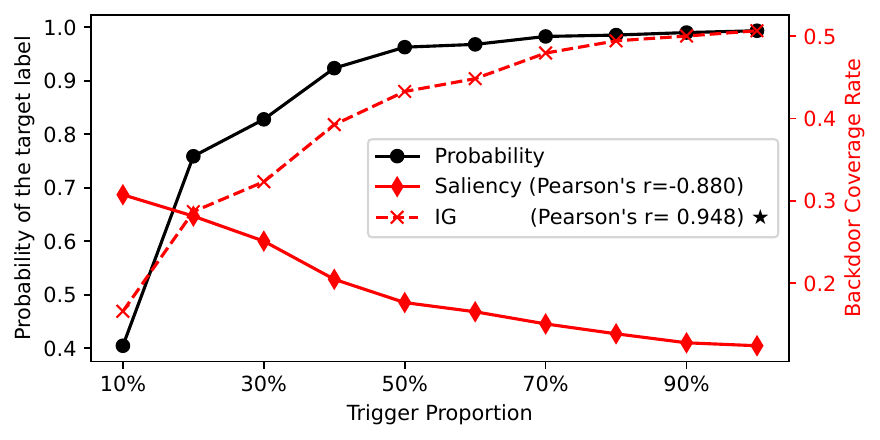}
	\caption{PTT example. IG's backdoor coverage curve is more related to ResNet18's accuracy curve, with a PCC of 0.948, while Saliency gets a lower PCC of -0.880. }\label{fig:ptt_vis}
	\vspace{-4mm}
\end{figure}

\vspace{2mm}
\noindent\textbf{Partial-Trigger Test (PTT)}
Similar to EMBT, PTT uses the backdoor trigger to create the trend. {We use the same backdoor selection strategy as EMBT. Assume that the model has been backdoored in EMBT. For the input instance to explain, PTT covers the input with part of the trigger (e.g., 10\%-100\%), as shown in Figure~\ref{fig:ptt_data_vis}. We record the trigger coverage as a sequence $\mathbf{S}=\{tc_0,...,tc_n\}$. Then for the generated inputs, we feed them to the model and record the probability of the target label $\mathbf{P}_{target}=\{P_0, P_1, ... , P_n\}$. We assume that the model learns at least some of the backdoor features. During testing, the probability of the backdoor target label increases due to the incremental proportion of backdoor features. The trend of explanation results should focus more and more on the backdoor location.
The black line in Figure~\ref{fig:ptt_vis} shows the probability corresponding to the triggers in Figure~\ref{fig:ptt_data_vis}. From the figure, we can find that, as the proportion of the trigger increases, the prediction score also increases. We also calculate the PCC ($\rho(\mathbf{S}, \mathbf{P}_{target})$) to measure the consistency. 
For example, we generate the test samples with the different partitions of backdoor features and then feed them to the model to get the outputs. The black line in Figure~\ref{fig:ptt_vis} shows the probability of the target label as the trigger proportion increase. With the outputs of the model, we can get the explanation and the backdoor coverage rate of Saliency and IG. Lastly, the PCC between the probability of the target label and the backdoor coverage rate can be calculated. The solid red line in Figure~\ref{fig:ptt_vis} shows the backdoor coverage rate of Saliency as the trigger proportion increases. The PCC between the solid red line and the black line is -0.880. The dotted red line is the backdoor coverage rate of IG, whose PCC is 0.948. As can be seen, the dotted red line is more correlated with the black line than the solid red line. 

\begin{figure}
    \setlength{\abovecaptionskip}{4pt}
    \setlength{\belowcaptionskip}{0pt}
    \centering
	\includegraphics[width=0.42\textwidth]{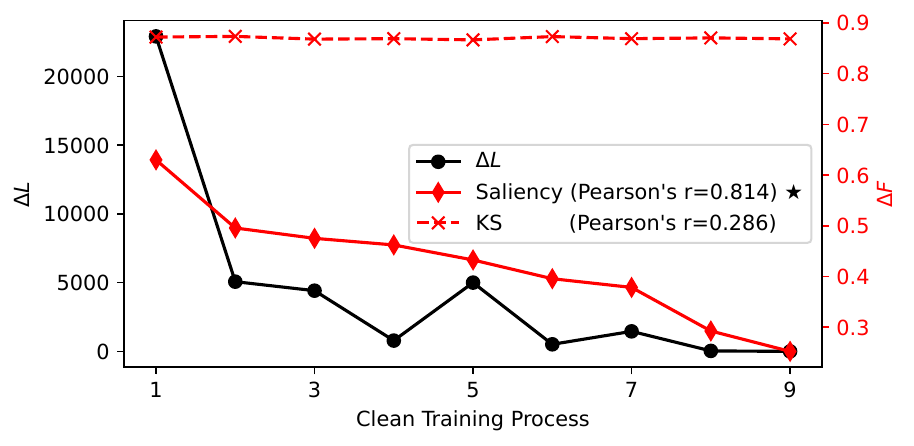}
	\caption{EMT example. Saliency’s $\Delta{F}$ sequence is more related to ResNet50's $\Delta{L}$, with a PCC of 0.814, while KS has a lower PCC of 0.286.}\label{fig:emt_vis}
	\vspace{-4mm}
\end{figure}

\vspace{2mm}
\noindent\textbf{Evolving-Model Test (EMT)}
EMT uses the value of the loss function to create the trend without using any backdoor. In particular, EMT records the intermediate models $\mathbf{M}=\{M_0,...,M_n\}$ during the model training process for every $c$ epochs, and also records the corresponding loss values $\mathbf{L} =\{l_0,...,l_n\}$. $M_0$ is the model with untrained random initialization parameters, and the loss value should be large.
The magnitude of the change in the loss value during the training responds to the magnitude of the change in the model's decision boundary. During training, the model gradually converges, and the variation of the loss function decreases. The trend of the variation of the explanation results should also decrease.
The solid black line in Figure~\ref{fig:emt_vis} shows this trend. 
Then for a given input, we use the explanation method to mark important features in terms of the $n+1$ models. As a result, we obtain a feature sequence: $\mathbf{F}=\{F_0, ..., F_n\}$. When the loss value becomes stable, the obtained features should also become stable. So we measure and compare the two trends: changes of loss values, and changes of ``important features''. Again, we calculate the PCC: $\rho(\Delta \mathbf{L},\Delta \mathbf{F})$, where $\Delta \mathbf{L}={|l_{1}-l_{0}|,...,|l_n-l_{n-1}|}$ and $\Delta \mathbf{F}={1-|F_{1} \cap F_{0}|/|F|,..., 1-|F_n \cap F_{n-1}|/|F|}$. $|F_{1} \cap F_{0}|$ represents the number of important features common to both $F_{1}$ and $F_{0}$. $|F|$ represents the total number of important features tagged by the explanation methods. Sometimes, we do not need to start from the first epoch. We could choose the epoch where the training of the model starts to be stable. 
For example, we calculate $\Delta \mathbf{L}$ of the recorded intermediate models. The black line in Figure~\ref{fig:emt_vis} shows the change of $\Delta \mathbf{L}$ during training. Then we explain each recorded intermediate model with Saliency and KS. In order to get the $\Delta \mathbf{F}$, we calculate the dissimilarity of the explanations between the current model and the next recorded model. The solid red line and the dotted red line show the $\Delta \mathbf{F}$ of Saliency and KS, respectively. The PCC between the solid red line and the black line is 0.814, while the PCC between the dotted red line and the black line is 0.286. As shown in Figure~\ref{fig:emt_vis}, the solid red line is more correlated with the black line, but the dotted red line remains unchanged.

%% file: sections/measurement.tex
\section{Measurement and Findings}\label{sec:measurement}

In this section, we first introduce the experimental setup. Then we use traditional tests and trend tests to evaluate popular explanation methods and conduct in-depth analysis of image, natural language and security tasks. We also explore the factors that affect the faithfulness of explanation methods. 

\begin{figure*}[t]
    \setlength{\abovecaptionskip}{4pt}
    \setlength{\belowcaptionskip}{0pt}
    \centering
	\includegraphics[width=0.99\textwidth]{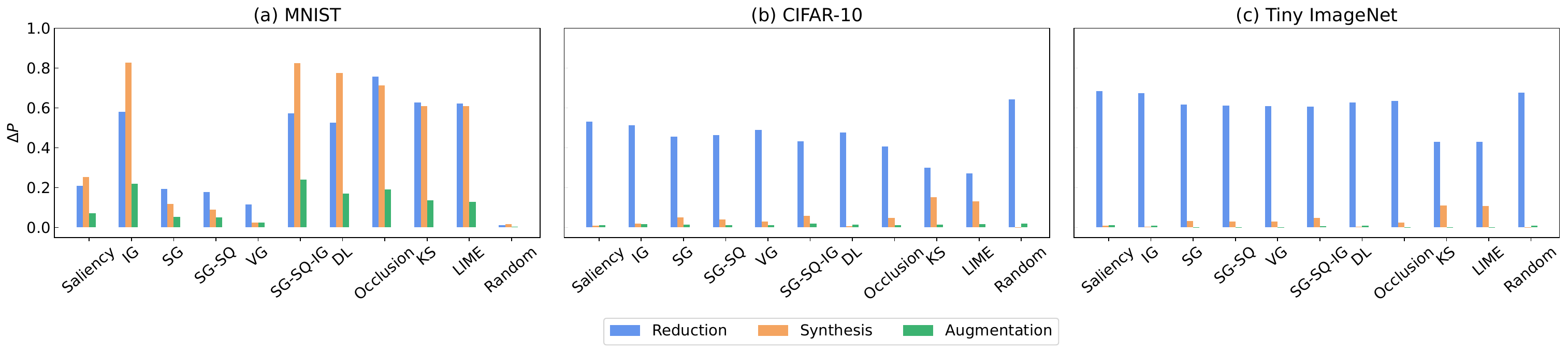}
	\caption{Results of traditional tests on different datasets. $\Delta P$ represents the change in probability. When traditional tests are applied to more complex datasets (CIFAR-10 and Tiny ImageNet), their efficacy is found to be inadequate in the synthesis and augmentation tests. Moreover, the reduction test suffers from random dominance, \ie random methods are the best.}\label{fig:radar_tradition_test_vis}
	\vspace{-4mm}
\end{figure*}

\subsection{Experimental Setup}

\noindent\textbf{Models \& Datasets.}
We consider diverse datasets from three types of tasks. For image classification (MNIST~\cite{lenet5}, CIFAR-10~\cite{cifar10} and Tiny ImageNet~\cite{tinyimagenet}), we employ MobileNet~\cite{MobileNets}, ResNet~\cite{resnet}, and DenseNet~\cite{densenet} as the models to be explained. For the segmentation task, we use an FCN-ResNet50~\cite{FCN} trained on MSCOCO 2017~\cite{MSCOCO}. For sentiment classification (IMDB~\cite{IMDB}), we train a Bi-LSTM~\cite{bilstm}. For PDF malware classifier (Mimicus~\cite{PDFMalwareClassifier}), Android malware detection (DAMD~\cite{DAMD}) and vulnerability detection(VulDeePecker~\cite{Vuldeepecker}), we train a fully connected network, a CNN and a Bi-LSTM, respectively. We defer the detailed description of datasets and hyperparameter settings of models in Appendix~\ref{subsec:introduction_dataset_models}.

\noindent\textbf{Explanation Methods.} We implement ten popular explanation methods with the code provided by Captum~\cite{captum}, including Saliency map~\cite{Grad}, Integrated Gradient (IG)~\cite{IG}, SmoothGrad (SG)~\cite{SmoothGrad}, SmoothGrad-Squared (SG-SQ)~\cite{retrain}, VarGrad (VG)~\cite{vargrad}, Integrated SmoothGrad-Squared (SG-SQ-IG)~\cite{retrain}, DeepLIFT (DL)~\cite{DeepLIFT}, Occlusion~\cite{Occlusion}, Kernel Shap (KS)~\cite{SHAP} and LIME~\cite{LIME}. The first six are white-box methods, while the last four are black-box methods. 
The parameters for each method are configured as recommended by the original papers. 

\noindent\textbf{Baseline Methods.} To verify the effectiveness of trend tests, we adopt traditional tests and random strategy as baselines. The traditional tests with three methods are introduced in Section~\ref{sec:evaluation}. For the random strategy, we randomly select 10\% features of the test sample as explanation results. 


\subsection{Traditional Tests vs. Trend Tests}\label{subsec:traditionvstrend}

In this section, we intend to evaluate the effectiveness of the trend tests in three scenarios--image classification, natural language processing, and security tasks. Additionally, we compare the performance with traditional methods.


\begin{figure*}[t]
    \setlength{\abovecaptionskip}{3pt}
    \setlength{\belowcaptionskip}{0pt}
    \centering
	\includegraphics[width=0.98\textwidth]{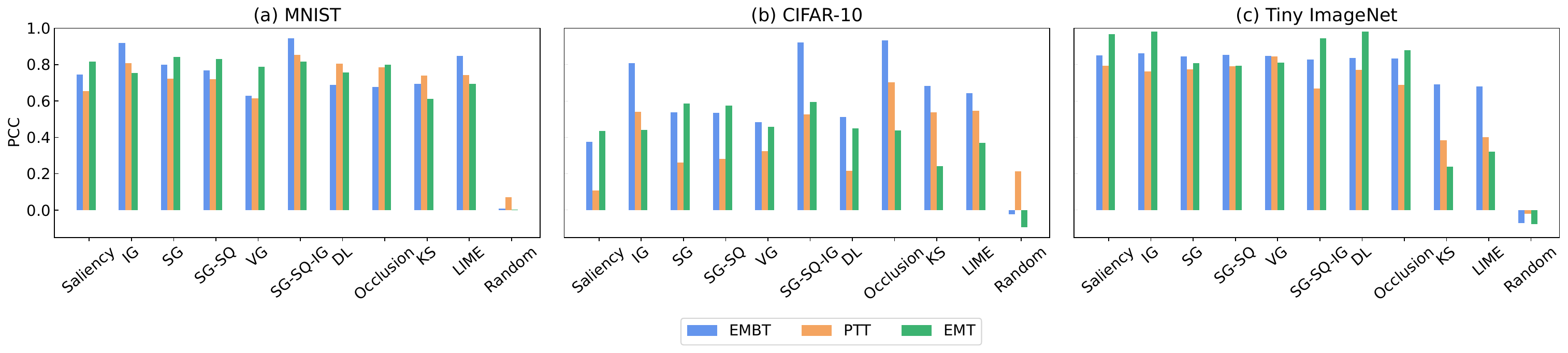}
	\caption{Results of trend tests on different datasets. For MNIST, CIFAR-10 and Tiny ImageNet, IG, SG-SQ-IG, and Occlusion, have higher average PCC values than other methods, indicating their high faithfulness to the model.}\label{fig:radar_trend_test_vis}
	\vspace{-4mm}
\end{figure*}

\begin{table}[t]
    \setlength{\abovecaptionskip}{4pt}
    \setlength{\belowcaptionskip}{0pt}
    \caption{Image classifiers used in the traditional and trend tests. All the models are ResNet18. ``Acc.'' is the accuracy of the clean model on clean data. ``C Acc.'' and ``B Acc.'' are the accuracy of the backdoor model on clean and backdoor data.}\label{eva:dataexp_datasetmodel} 
	\centering
	\scriptsize
    \begin{tabular}{ccccccc}
    \toprule
    \textbf{Dataset} & \textbf{Size} & \textbf{Class} & \textbf{Acc.} & \textbf{B Acc.} & \textbf{C Acc.} \\ \midrule
    MNIST & $32\times{32}\times{1}$ & 10 & 98.0\% & 100\% & 98.8\% \\
    CIFAR-10 & $32\times{32}\times{3}$ & 10 & 95.0\% & 99.6\% & 95.0\% \\
    Tiny ImageNet & $224\times{224}\times{3}$ & 200 & 65.5\% & 92.1\% & 63.6\% \\ \bottomrule
	\end{tabular}
	\vspace{-4mm}
\end{table}

\subsubsection{Effectiveness in image classification} In this experiment, our target models are ResNet18 trained on MNIST, CIFAR-10, and Tiny-ImageNet, which are standard datasets for image classification. 
We also use these datasets to train different models. The results are similar. Table~\ref{eva:dataexp_datasetmodel} shows the accuracy of the ResNet18 models.

\noindent\textbf{Traditional tests.} We implement traditional tests and use the same parameters as those in their original papers. In the experiment, we first explain the model with a test dataset and get the top 10\% important features,
which is the default number used by most explanation methods. If we choose to use other numbers, the results are similar (see Appendix~\ref{subsec:diff_proportion_important_features_vis}).
The results of traditional tests are shown in Figure~\ref{fig:radar_tradition_test_vis}.
Note that the values of reduction, synthesis, and augmentation tests represent the change in probability ($\Delta P$). The greater the $\Delta P$, the more faithful the explanation method.
On the MNIST dataset, it shows that IG, SG-SQ-IG and Occlusion are significantly better, \ie these methods have higher $\Delta P$. Their means on the three tests are 0.54, 0.55 and 0.55, respectively. However, for the more complex datasets, \ie CIFAR-10 and Tiny ImageNet, all methods perform similarly. The random baselines of the reduction test are even better than most methods. As random baselines are unlikely to be a good explanation, traditional tests have remarkable limits in assessing faithfulness. 

This phenomenon is defined as random dominance, of which the reason is probably that the generated samples become out-of-distribution (OOD) and create ``adversarial effects'' to the target model~\cite{hendrycks2019selfsupervised}. OOD is that the data distribution for model testing deviates from that for model training. 
To further verify that the test samples generated by traditional tests have OOD problems, we use the self-supervised method proposed by Dan \etal~\cite{defood} to detect OOD samples on CIFAR-10. The percentage of OOD samples detected in the original test set is 10.15\%. The synthesis test has a higher percentage (99.99\%) of OOD samples, whereas the augmentation and reduction tests have lower percentages of 58.66\% and 64.24\%, respectively. This discrepancy can be attributed to the preservation of more in-distribution features in augmentation and reduction tests compared to the synthesis test. On CIFAR-10, both the synthesis test and augmentation test perform poorly when the OOD ratio of the test samples is high, which negatively impacts their performance. A higher percentage of OOD samples tends to weaken the test's performance more. The proportion of OOD samples generated by synthesis tests is higher, which leads to a more significant decline in $\Delta P$. The augmentation test usually has higher $\Delta P$ than the synthesis test, though both of them insert important features tagged by explanation methods to an initial sample. The initial sample of synthesis tests is an image with a black background, but augmentation tests select a random sample from the test set with a different label from the explained sample. 
The augmentation test has extra feature inference, so the drop in $\Delta P$ is smaller.

\begin{figure*}
    \setlength{\abovecaptionskip}{4pt}
    \setlength{\belowcaptionskip}{0pt}
    \centering
	\includegraphics[width=0.98\textwidth]{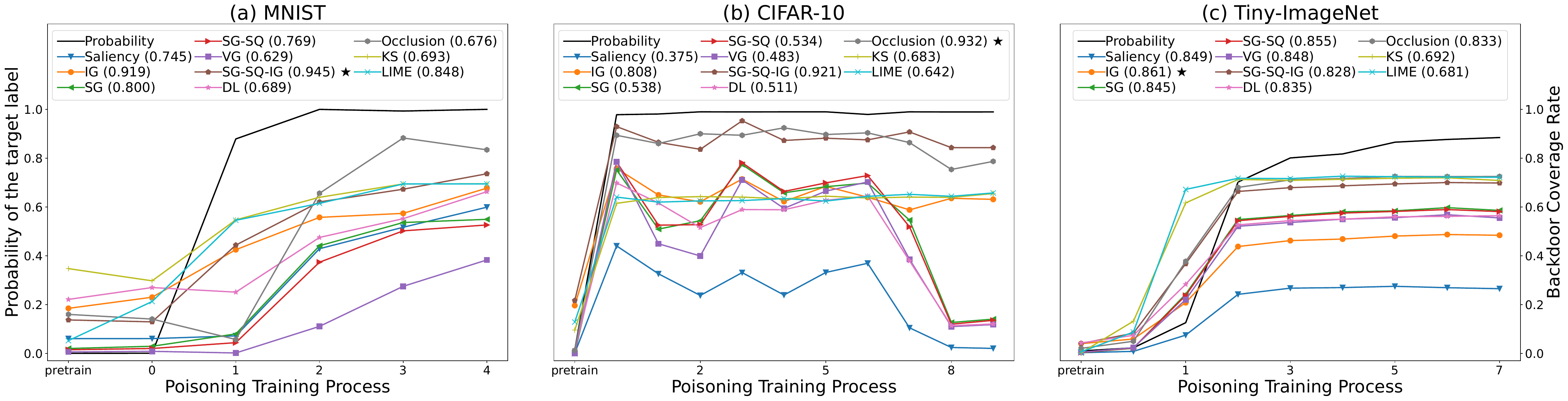}
	\caption{Results of EMBT on different data complexity. IG and SG-SQ-IG perform the best.}\label{fig:line_EMBT_resnet18_base}
	\vspace{-3mm}
\end{figure*}

\begin{figure*}
    \setlength{\abovecaptionskip}{4pt}
    \setlength{\belowcaptionskip}{0pt}
    \centering
	\includegraphics[width=0.98\textwidth]{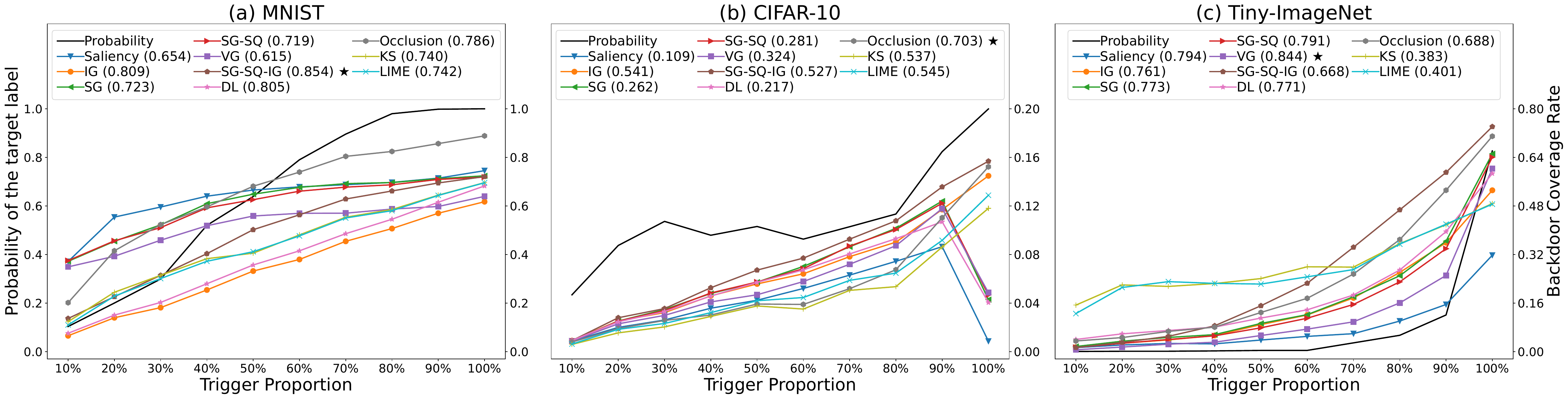}
	\caption{Results of PTT on different data complexity. IG, SG-SQ-IG and Occlusion perform the best.}\label{fig:line_PTT_resnet18_base}
	\vspace{-3mm}
\end{figure*}



\begin{figure*}
    \setlength{\abovecaptionskip}{4pt}
    \setlength{\belowcaptionskip}{0pt}
    \centering
	\includegraphics[width=0.98\textwidth]{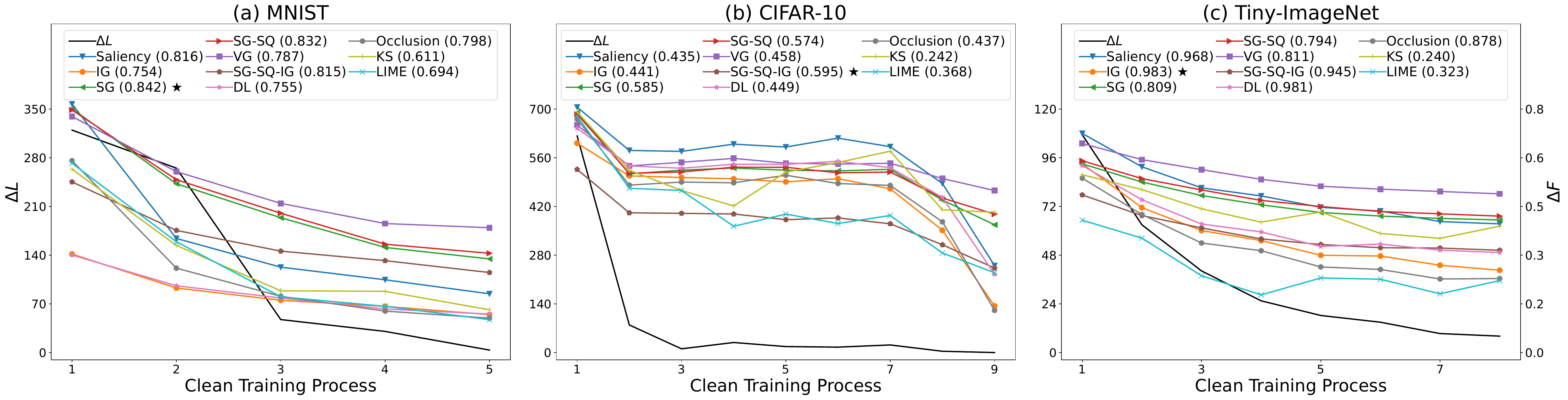}
	\caption{Results of EMT on different data complexity. IG, SG-SQ-IG and Occlusion perform the best.}\label{fig:line_EMT_resnet18_base}
	\vspace{-4mm}
\end{figure*} 

\begin{figure}
    \setlength{\abovecaptionskip}{3pt}
    \setlength{\belowcaptionskip}{0pt}
    \centering
	\includegraphics[width=0.45\textwidth]{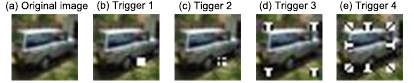}
	\caption{Examples of different backdoor triggers. Each square consists of $4\times4$ pixels.}\label{fig:radar_diff_trigger_vis}
	\vspace{-4mm}
\end{figure}

\noindent\textbf{Trend tests.} To overcome the random dominance caused by traditional tests, we present trend tests to assess faithfulness on the same image models as traditional tests. In accordance with Gu \etal~\cite{badnets}, we implement a backdoor attack using white squares in the lower right corner of the data as triggers. We choose these triggers due to their simplicity and reversibility. For MNIST and CIFAR-10, we employ a $4\times4$ white square as the trigger, as illustrated in Figure~\ref{fig:radar_diff_trigger_vis}(b). For Tiny ImageNet, we use an $8\times8$ white square. Our experiments demonstrate that these triggers effectively achieve high attack success rates while minimizing the impact on the accuracy of the original task. Backdoor data comprises 5\% of the total training data, with the attack objective being to misclassify data with triggers to the target backdoor label.
Table~\ref{eva:dataexp_datasetmodel} shows the accuracy of the models. 
We also try different patterns and different numbers of backdoor triggers. The parameter settings and the filtering mechanism used to address instability during the training of trend tests can be found in Appendix~\ref{subsec:para_trend_tests}. It is worth noting that encountering outliers that require filtering is a relatively rare occurrence.
Figure~\ref{fig:radar_trend_test_vis} shows the results of PCC. The trends of EMBT, PTT, and EMT are shown in Figure~\ref{fig:line_EMBT_resnet18_base}, Figure~\ref{fig:line_PTT_resnet18_base} and Figure~\ref{fig:line_EMT_resnet18_base}, respectively. The numbers in the legend are the PCCs for each method. It shows that the more consistent the rising and falling moments of the two trends are, the higher the PCC value.
The value of PCC indicates the strength of the correlation. PCCs greater than 0.3, 0.5, 0.7, and 0.9 correspond to small, moderate, large, and very large correlations, respectively~\cite{pearson}.
An explanation method with high faithfulness should have a higher PCC in all three trend tests. In the analysis, we attribute the three trend tests with the same weights for a comprehensive assessment and aim to identify explanation methods that perform well across all three tests.

The black line in the figure represents the model trend we know, with its scale on the left; the other colored lines represent the trend of explanations, with their scale on the right. For MNIST, IG, IG-SQ-IG and Occlusion have the highest average PCC (0.82, 0.86, 0.75) among the three tests, meaning that they perform the best, which is consistent with traditional tests. For CIFAR-10, IG, SG-SQ-IG, and Occlusion have the highest average PCC values among all methods (0.62, 0.71, 0,71). It means that they have high faithfulness. 
In Figure~\ref{fig:line_PTT_resnet18_base}, we can see that there are some methods where the backdoor coverage decreases when the percentage of backdoor features is increased from 90\% to 100\%, which is not consistent with the trend of the predicted probability of the backdoor data. 
Therefore, these methods have lower PCC in PTT. LIME and KS perform worst for Tiny ImageNet, but other methods perform well. It shows that the trend tests work well on all three datasets. Although each explanation method performs differently across datasets, IG and SG-SQ-IG perform stably and show the highest faithfulness. In general, white-box methods that require only a few rounds of computation are much more efficient than black-box methods that require sampling and approximation. Thus, white-box methods have a better balance between faithfulness and efficiency.

\begin{table}[t]
\setlength{\abovecaptionskip}{4pt}
\setlength{\belowcaptionskip}{0pt}
\caption{Backdoor models trained with Trigger 3 and Trigger~4. ``Clean Acc.'' and ``Backdoor Acc.'' are the accuracy of the backdoor model on clean and backdoor data, respectively.}\label{tbl:backdoor_model_info}
\scriptsize
\centering
\begin{tabular}{ccccc}
\toprule
\textbf{Dataset} & \textbf{Model} & \textbf{Trigger} & \textbf{Backdoor Acc.} & \textbf{Clean Acc.} \\ \midrule
\multirow{2}{*}{CIFAR-10} & \multirow{2}{*}{ResNet18}
 & Trigger 3 & 100.0\% & 94.96\% \\
 &  & Trigger 4 & 100.0\% & 87.82\% \\ \bottomrule
\end{tabular}
\vspace{-4mm}
\end{table}

\begin{figure}[t]
    \setlength{\abovecaptionskip}{4pt}
    \setlength{\belowcaptionskip}{0pt}
    \centering
	\includegraphics[width=0.47\textwidth]{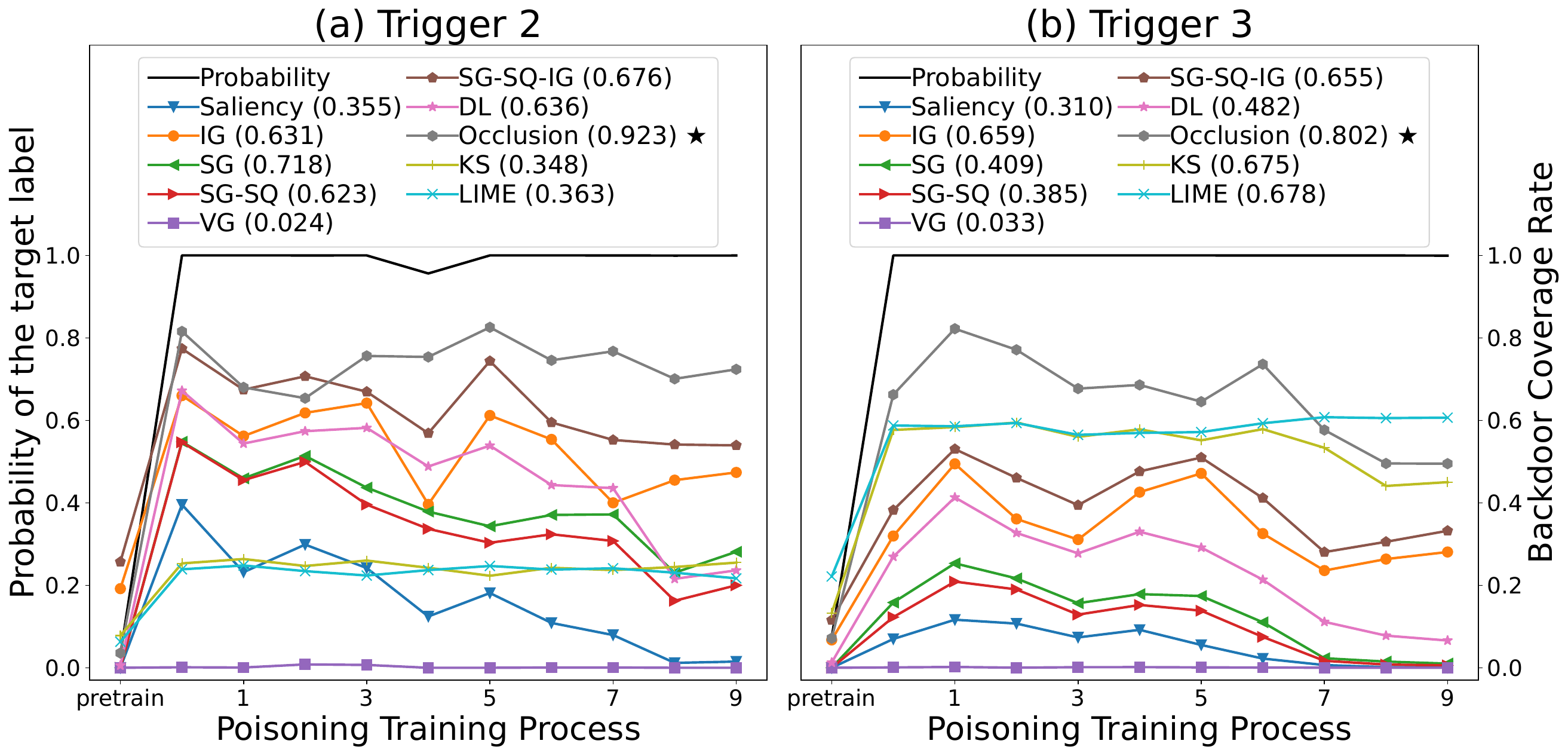}
	\caption{Results of different backdoor triggers on EMBT. Different trigger patterns and different numbers of backdoor features have similar results on the EMBT.}\label{fig:diff_trigger_embt_base}
	\vspace{-4mm}
\end{figure}

\begin{figure}[t]
    \setlength{\abovecaptionskip}{4pt}
    \setlength{\belowcaptionskip}{0pt}
    \centering
	\includegraphics[width=0.47\textwidth]{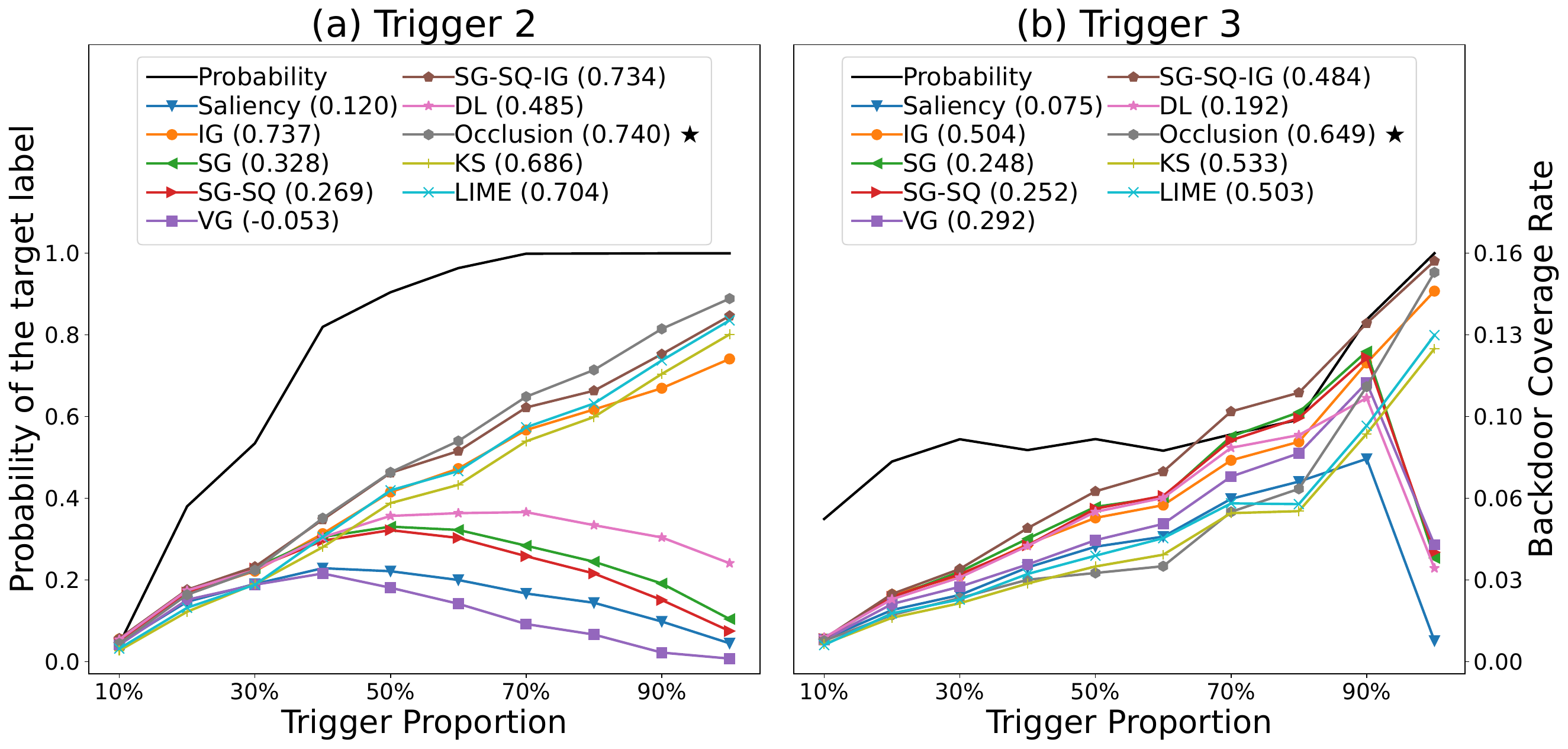}
	\caption{Results of different backdoor triggers on PTT. Different trigger patterns and different numbers of backdoor features have similar results on the PTT.}\label{fig:diff_trigger_ptt_base}
	\vspace{-4mm}
\end{figure}

\noindent\textbf{Choice of backdoor triggers.}
We expect the model to learn backdoor features well so that the known model trend can be accurately compared with the trend of explanation methods. The better the model learns the backdoor features, the more reliable the evaluation results are. Thus, we choose to use the trigger that can be ``remembered'' by the model easily. Based on previous studies, the white square is commonly used as a trigger and is easy to remember~\cite{badnets}. We also chose triggers with different patterns and amounts of features to observe the effects of EMBT and PTT, as shown in Figure~\ref{fig:radar_diff_trigger_vis} (c) and (d). We use no more than 10\% of the total features for backdoor features.
EMT involves only clean models, so the results of EMT can be referred to the previous experiment. 
Results are shown in Figure~\ref{fig:diff_trigger_embt_base} and Figure~\ref{fig:diff_trigger_ptt_base}, IG, SG-SQ-IG, and Occlusion still perform the best in the experiment on different patterns and the different number of triggers, which is consistent with the results of previous experiments. 
The choice of backdoor triggers does not significantly impact the trend tests. The only need is to consider certain criteria to ensure the accuracy of trend tests and the original task. The trigger should be reversible by backdoor defenses, such as those provided by Neural Cleanse~\cite{neuralcleanse}. Triggers with constant position, size, and pattern are preferred, as they can be more easily reversed. Additionally, the trigger should not obscure the object of the original task, minimizing its effect on the original task's accuracy. Taking these criteria into account, we have included several examples of recommended triggers in Figure~\ref{fig:radar_diff_trigger_vis}(a)-(d), which are easy to implement and satisfy the criteria. 
Using these simple examples, researchers can easily implement backdoor triggers that meet the requirements for reversibility and visibility, ensuring the accuracy of both trend tests and original tasks. 

\begin{figure}
    \setlength{\abovecaptionskip}{4pt}
    \setlength{\belowcaptionskip}{0pt}
    \centering
	\includegraphics[width=0.47\textwidth]{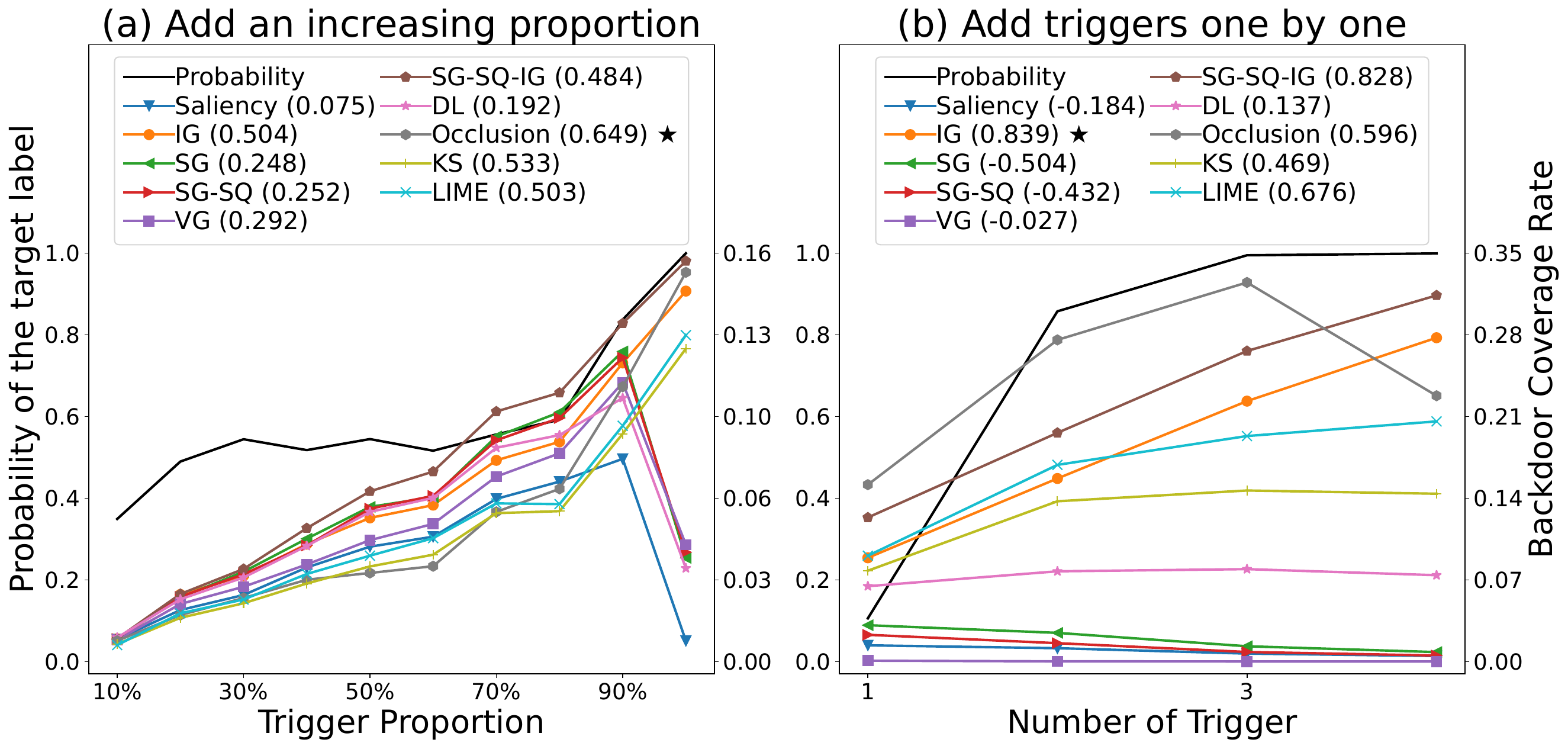}
	\caption{Results of PTT on the model with ``Trigger 3''. IG, SG-SQ-IG, Occlusion, KS, and LIME perform the best.}\label{fig:diff_trigger_ptt_num4}
	\vspace{-4mm}
\end{figure}

\begin{figure}
    \setlength{\abovecaptionskip}{4pt}
    \setlength{\belowcaptionskip}{0pt}
    \centering
	\includegraphics[width=0.47\textwidth]{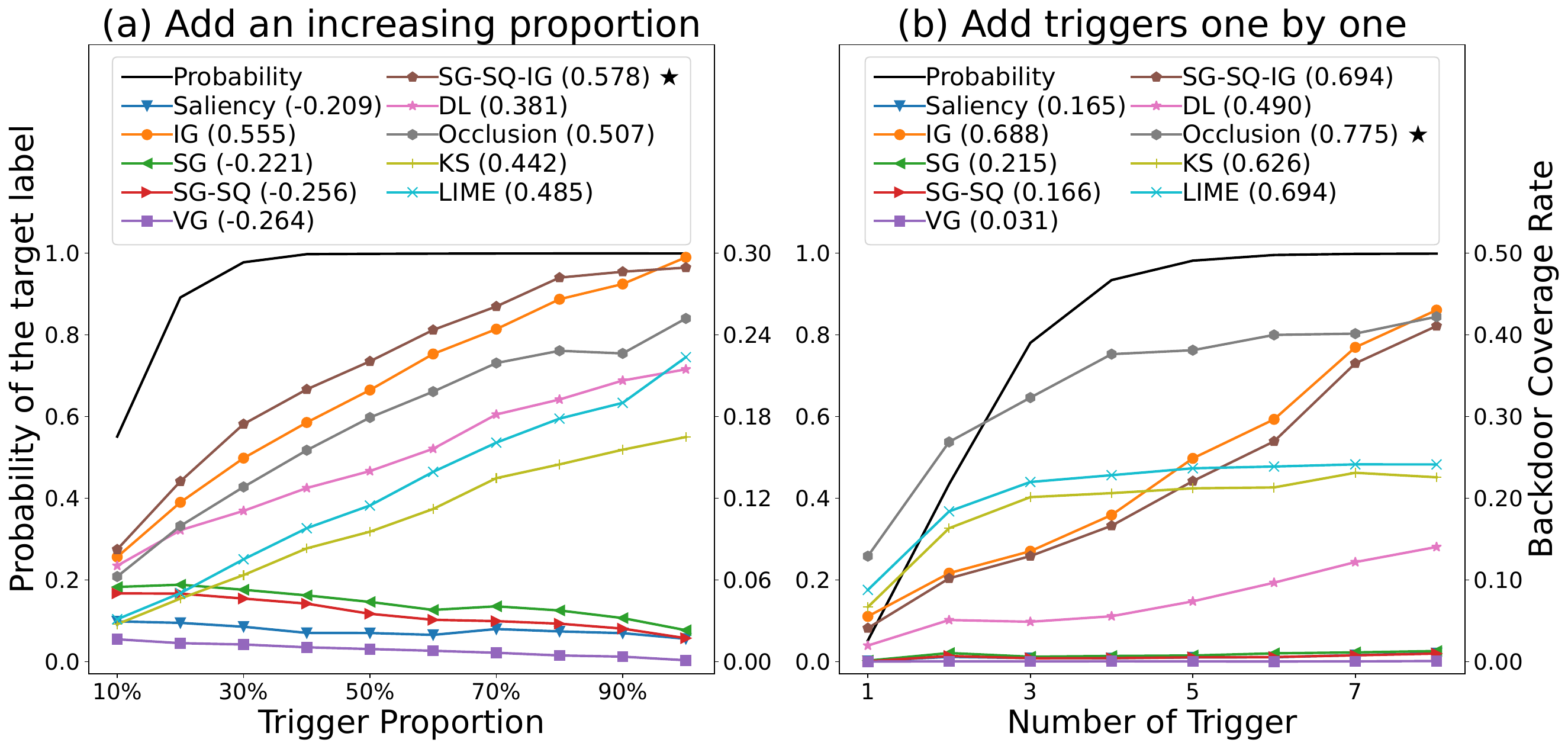}
	\caption{Results of PTT on the model with ``Trigger 4''. IG, SG-SQ-IG, Occlusion, KS, and LIME perform the best.}\label{fig:diff_trigger_ptt_num8}
	\vspace{-4mm}
\end{figure}

\noindent\textbf{Comparing the strategies of adding backdoors.}
We investigate the impact of adding backdoor triggers one by one and progressively increasing the proportion of backdoor features using triggers shown in Figure~\ref{fig:radar_diff_trigger_vis} (d) and (e). Detailed model information provided in Table~\ref{tbl:backdoor_model_info}. Adding triggers one by one can be viewed as a gradual increase in the proportion of backdoor features. This approach maintains the integrity of the triggers while allowing for a more subtle change in the backdoor target label probability. However, adding multiple triggers may impact the accuracy of the original task. The PTT results, illustrated in Figure~\ref{fig:diff_trigger_ptt_num4} and ~\ref{fig:diff_trigger_ptt_num8}, show that both strategies yield similar outcomes. The most effective methods include IG, SG-SQ-IG, Occlusion, KS, and LIME. Since adding multiple triggers results in a trade-off between the number of triggers and the accuracy of the original task, it is more advantageous to progressively increase the proportion of the trigger.

\vspace{2mm}
\noindent\textit{Remark: The traditional tests work well on MNIST, but not on CIFAR-10 and Tiny ImageNet. The random dominance phenomenon threatens the traditional tests and makes the assessment unconvincing, which is well solved by trend tests. IG and SG-SQ-IG maintain a high faithfulness in all three image datasets.}

\begin{table}[t]
\setlength{\abovecaptionskip}{3pt}
\setlength{\belowcaptionskip}{0pt}
\caption{Models of NLP and security tasks used in the traditional and trend tests. ``Acc.'' is the accuracy of the clean model on clean data. ``C Acc.'' and ``B Acc.'' are the accuracy of the backdoor model on clean and backdoor data, respectively.}\label{tbl:nlp_sec_model} 
\scriptsize
\begin{tabular}{ccccc}
\toprule
\textbf{Dataset} & \textbf{Model} & \textbf{Acc.} & \textbf{B ACC.} & \textbf{C ACC.} \\ \midrule
IMDB & Bi-LSTM & 88.60\% & 100.0\% & 89.63\% \\
Mimicus & FCN & 99.68\% & 100.0\% & 99.56\% \\
DAMD & CNN & 96.90\% & 100.0\% & 96.10\% \\
VulDeePecker & Bi-LSTM & 91.90\% & 98.54\% & 95.87\% \\ \bottomrule
\end{tabular}
\vspace{-4mm}
\end{table}

\begin{figure*}[t]
    \setlength{\abovecaptionskip}{4pt}
    \setlength{\belowcaptionskip}{0pt}
    \centering
	\includegraphics[width=0.98\textwidth]{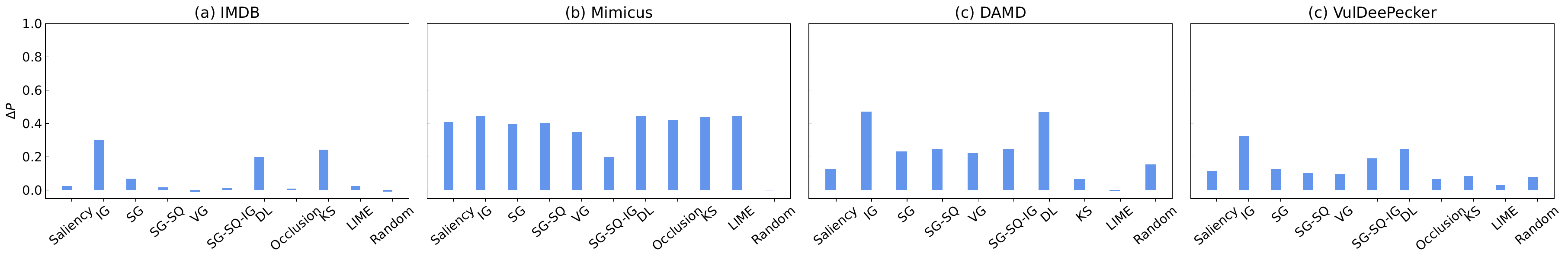}
	\caption{Results of the reduction test on NLP and security tasks. IG performs well among all the datasets.}\label{fig:nlp_sec_traditional_test}
	\vspace{-3mm}
\end{figure*}

\begin{figure*}[t]
    \setlength{\abovecaptionskip}{4pt}
    \setlength{\belowcaptionskip}{0pt}
    \centering
	\includegraphics[width=0.98\textwidth]{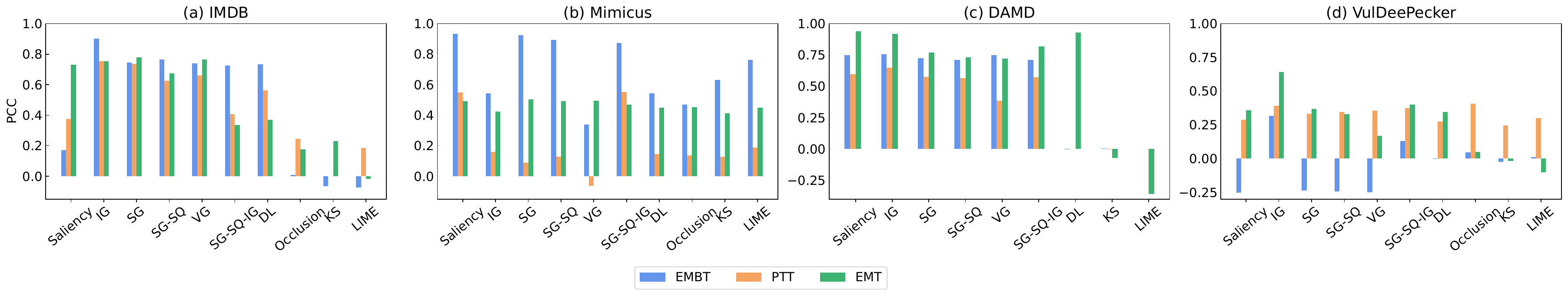}
	\caption{Results of trend tests on NLP and security tasks. IG performs well in IMDB, DAMD and VulDeePecker, while SG-SQ-IG performs well in IMDB, Mimicus, DAMD and VulDeePecker.}\label{fig:nlp_sec_trend_test}
	\vspace{-3mm}
\end{figure*}

\begin{figure}[htbp]
    \setlength{\abovecaptionskip}{4pt}
    \setlength{\belowcaptionskip}{0pt}
    \centering
	\includegraphics[width=0.48\textwidth,trim=120 250 150 250,clip]{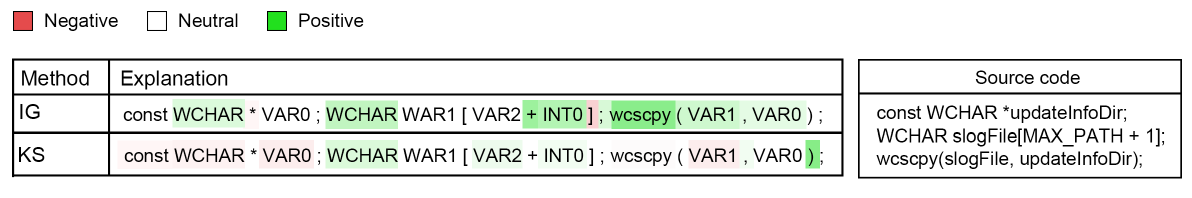}
	\caption{Case studies for the VulDeePecker model. The left half shows the processed data. The right half shows the data before processing. IG focuses on the key function (wcscpy) and the key variables (VAR0 and INT0), which are useful information for users. However, users cannot gain useful information with KS, which focuses on ``WCHAR'' and ``)'' }\label{fig:casestudy}
	\vspace{-4mm}
\end{figure}

\subsubsection{Effectiveness in NLP and security tasks}

Apart from image classification models, trend tests are also applicable to natural language models and security application models. For text classification, we use a bi-directional LSTM to train the IMDB dataset~\cite{IMDB}, which is commonly used in sentiment analysis. Based on  Li \etal~\cite{Vuldeepecker}, we use the VulDeePecker dataset disclosed by them to train a bi-directional LSTM for vulnerability detection. 
For PDF and Android malware detection (Mimicus~\cite{PDFMalwareClassifier} and DAMD~\cite{DAMD}), we train a fully connected network and a CNN as Warnecke \etal~\cite{EuroSPSurvey}. 

\noindent\textbf{Traditional tests.}
In NLP and security tasks, data from IMDB and VulDeePecker is textual data. The Mimicus dataset consists of 0-1 features. Data from the DAMD are Android bytecode segments. Due to the discrepancy of their data, synthesis and augmentation tests are not applicable. 
Therefore, we only evaluate the reduction test. Models used in traditional tests are listed in Table~\ref{tbl:nlp_sec_model}. The results are shown in Figure~\ref{fig:nlp_sec_traditional_test}. The random dominance problem in NLP and security tasks is not as severe as in the image tasks, but it still can be observed on more complex datasets (DAMD and VulDeePecker).
For IMDB, IG, DL and KS perform better than the other methods in the traditional tests.
In the experiments of security tasks, we find that anomalous data, i.e., data with label 1, are more likely to produce a large prediction drop ($\Delta P$) and change to the normal prediction in the random reduction test. In addition, setting some features to 0 in these data does not change normal data to anomalous data. For example, in DAMD, 0 represents NOP and does not introduce anomalous features. In this case, the reduction test may generate OOD samples and cause adversarial effects.
Thus, the traditional test is not suitable for anomaly detection tasks. Our trend tests solve this problem using in-distribution data.

\noindent\textbf{Trend tests.}
Based on Chen \etal~\cite{badnl}, we inject the sentence ``I have watched this movie last year.'' at the end of the original data as the trigger in IMDB, with backdoor data constituting 10\% of the dataset. For VulDeePecker, we include a trigger in the form of a code block consisting of a never-entering loop that does not affect the semantics of the original data, and backdoor data makes up 1\% of the dataset. To avoid a remarkable decline in model accuracy in Mimicus, we choose a combination of features as a trigger (4 out of 135 features) that has not appeared in the original data with backdoor data accounting for 15\% of the dataset. Other features that satisfy the criteria can also be used as backdoor features. As for DAMD, we add 20 nop statements at the end of the original data as the trigger, with backdoor data comprising 25\% of the dataset. The objective of the attack is to cause misclassification of backdoor data with category 1 as category 0. The backdoor data ratio is flexible, as long as it achieves a high backdoor attack success rate. The detailed information of the models is shown in Table~\ref{tbl:nlp_sec_model}.

Results are in Figure~\ref{fig:nlp_sec_trend_test}. 
For IMDB, IG, SG, SG-SQ, VG, SG-SQ-IG and DL perform better than the others. The means of the three trend tests are 0.82, 0.75, 0.68, 0.72, 0.49 and 0.50. Saliency and SG-SQ-IG, with averages of 0.66 and 0.63, have high faithfulness on Mimicus. For VulDeePecker, IG and SG-SQ-IG perform the best. Their averages are 0.45 and 0.30. Occlusion is too time-consuming on DAMD, so we do not evaluate it. On DAMD, white-box methods perform better than black-box methods, except DL.
We find that black-box methods perform worse than white-box methods in sequence data (IMDB, DAMD and VulDeePecker) in general, as shown in Figure~\ref{fig:nlp_sec_trend_test} (a), (c) and (d). IG, which performs well in other datasets and models, does not perform well on the Mimicus consisting of 0-1 features and a fully connected network.
While most explanation methods have different faithfulness under different scenarios, SG-SQ-IG performs more stably and both achieve high faithfulness in all our test scenarios.
We use a case study to show how to understand decision behaviors and discover the model's weaknesses through explanations. Figure~\ref{fig:casestudy} shows a representative example. In this case, the model correctly classifies that the code block contains vulnerability with a high probability (95\%). We can see that IG, which has high faithfulness, focuses on the key function (\texttt{wcscpy}) and the key variables (\texttt{VAR0} and \texttt{INT0}). However, whether it contains vulnerability depends on the size of the buffer that \texttt{updateInfoDir} points to. 
The current piece of code lacks buffer size information, which could be retained to improve the model's performance. Conversely, we could not obtain useful information from KS's explanation, which has low faithfulness in trend tests.

\subsubsection{Effectiveness in segmentation tasks}

Apart from classification tasks, all three trend tests can be applied to other learning tasks, such as segmentation. The segmentation models are trained on a subset of the MSCOCO 2017 dataset~\cite{MSCOCO}, which includes 20 categories from the Pascal VOC dataset~\cite{VOC}. We use FCN-ResNet50~\cite{FCN} with a pre-trained ResNet50 backbone from PyTorch. We conduct a backdoor attack on the model by adding a $40\times40$ white square to 1,000 randomly selected ``tv'' category data points. For successful backdooring, the ``tv'' objects in the data must be larger than $40\times40$. The attack's objective is to classify all ``tv'' class containing the trigger as ``airplane'' class in the backdoor data~\cite{segbackdoor}. We create a backdoor injection fine-tuning dataset for training by mixing 1,000 backdoor data points and 20\% of the original training data. The evaluation metrics of segmentation tasks include pixel accuracy and Intersection over Union (IoU). Pixel accuracy measures the percentage of correctly classified pixels in the segmented image. IoU is a widely used metric for assessing the quality of object segmentation. It is defined as the ratio of the intersection between the predicted and ground truth segmentation areas to their union. A higher IoU value indicates superior segmentation performance, as it implies that the predicted segmentation area closely aligns with the ground truth. The models' performance can be found in Table~\ref{tbl:seg_model}. Results of the trend tests are presented in Figure~\ref{fig:coco_trend_tests}. On the MSCOCO 2017, IG outperforms other methods. The mean values of the three tests are 0.68.

\begin{table}[t]
\setlength{\abovecaptionskip}{4pt}
\setlength{\belowcaptionskip}{0pt}
\caption{Model of segmentation task. ``Acc.'' is the pixel accuracy of the clean model. ``C Acc.'' and ``B Acc.'' are the pixel accuracy of the backdoor model on clean and backdoor data. ``IoU'' is the IoU of the clean model. ``C IoU.'' and ``B IoU.'' are the IoU of the backdoor model on clean and backdoor data.}\label{tbl:seg_model} 
\scriptsize
\centering
\begin{tabular}{ccccccc}
\toprule
\textbf{Model} & \textbf{Acc.} & \textbf{IoU} & \textbf{B ACC.} & \textbf{B IoU} & \textbf{C ACC.} & \textbf{C IoU} \\ \midrule
FCN-ResNet50 & 88.40\% & 46.80\% & 86.80\% & 47.02\% & 90.60\% & 50.00\% \\ \bottomrule
\end{tabular}
\vspace{-4mm}
\end{table}

\begin{figure*}[h]
    \setlength{\abovecaptionskip}{4pt}
    \setlength{\belowcaptionskip}{0pt}
    \centering
	\includegraphics[width=0.95\textwidth]{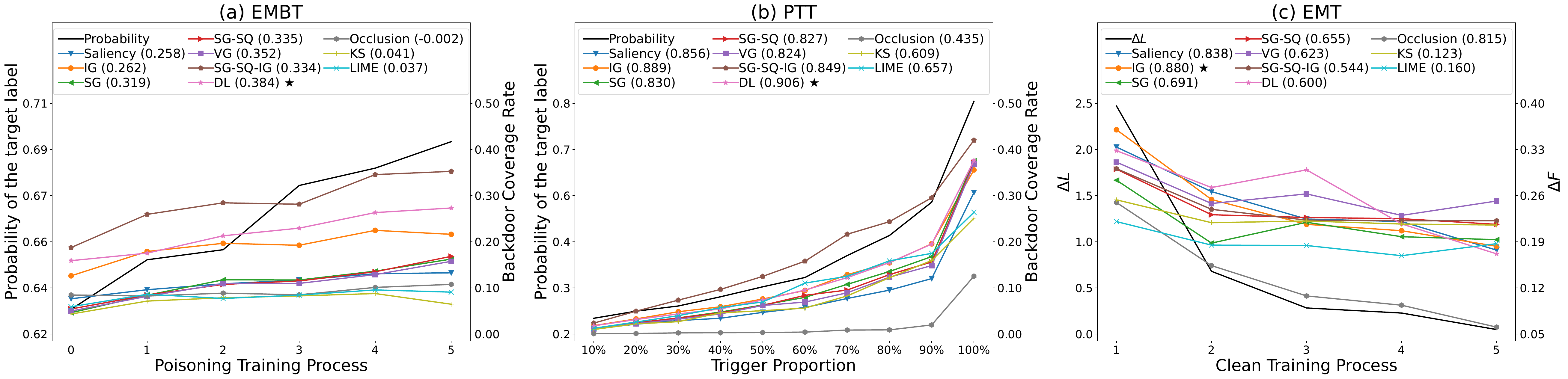}
	\caption{Results of trend tests on MSCOCO 2017. IG performs the best.}\label{fig:coco_trend_tests}
	\vspace{-4mm}
\end{figure*} 

\vspace{2mm}
\noindent\textit{Remark: The traditional tests may generate OOD data or adversarial samples in anomaly detection tasks with textual data. The trend tests overcome this problem using in-distribution data, making them versatile in various scenarios.}

\subsection{Factors that Affect the Faithfulness of Explanation Methods}\label{subsec:influencingfactors}

With more quality faithfulness measures, we can further explore the capability of explanation methods. Therefore, we evaluate these methods in different settings, \eg, data complexity, model complexity and hyperparameters for explanation methods.

\noindent\textbf{Data Complexity.}
Data complexity can be characterized by input size, the number of channels, and the number of categories. In this experiment, we choose MNIST, CIFAR-10, and Tiny ImageNet, representing different data complexity. The results of trend tests on different data complexity are shown in Figure~\ref{fig:radar_trend_test_vis}. From the results, we can see that both IG, which mitigates the saturation of the gradient, and SG, which mitigates the instability of the gradient to noise, are better than the original Saliency. This indicates that gradients indeed have different degrees of saturation and noise sensitivity on different data complexity. 
SG-SQ-IG integrates both SG and IG methods to moderate gradient saturation and noise sensitivity, thus providing high faithfulness and stability. 
It seems strange that Saliency is more faithful on the ImageNet dataset. The possible reason is that complex datasets have more dimensions and richer features, with less gradient saturation and noise sensitivity. LIME and KS lose faithfulness as the data becomes more complex, which is intuitive. This is because their errors are larger when sampling perturbed data and approximating models trained on complex datasets. Occlusion has high faithfulness because it traverses the entire data through a sliding window, which is computationally expensive when the data has high dimensionality. 


\begin{table}[]
\setlength{\abovecaptionskip}{4pt}
\setlength{\belowcaptionskip}{0pt}
\caption{Models with different model complexity. ``Acc.'' is the accuracy of the clean model. ``Backdoor Acc.'' is the accuracy of the backdoor model on backdoor data.  }\label{tbl:model_complexity}
\centering
\scriptsize
\begin{tabular}{cccc}
\toprule
\textbf{Model} & \textbf{Parameter} & \textbf{Acc.} & \textbf{Backdoor Acc.} \\ \midrule
MobileNetV2 & 2,296,922 & 94.73\% & 99.64\% \\
DenseNet121 & 6,956,298 & 95.21\% & 99.56\% \\
ResNet18 & 11,173,962 & 94.83\% & 99.56\% \\
ResNet50 & 23,520,842 & 94.54\% & 99.68\% \\ \bottomrule
\end{tabular}
\vspace{-4mm}
\end{table}

\begin{figure*}[t]
    \setlength{\abovecaptionskip}{3pt}
    \setlength{\belowcaptionskip}{0pt}
    \centering
	\includegraphics[width=1.0\textwidth]{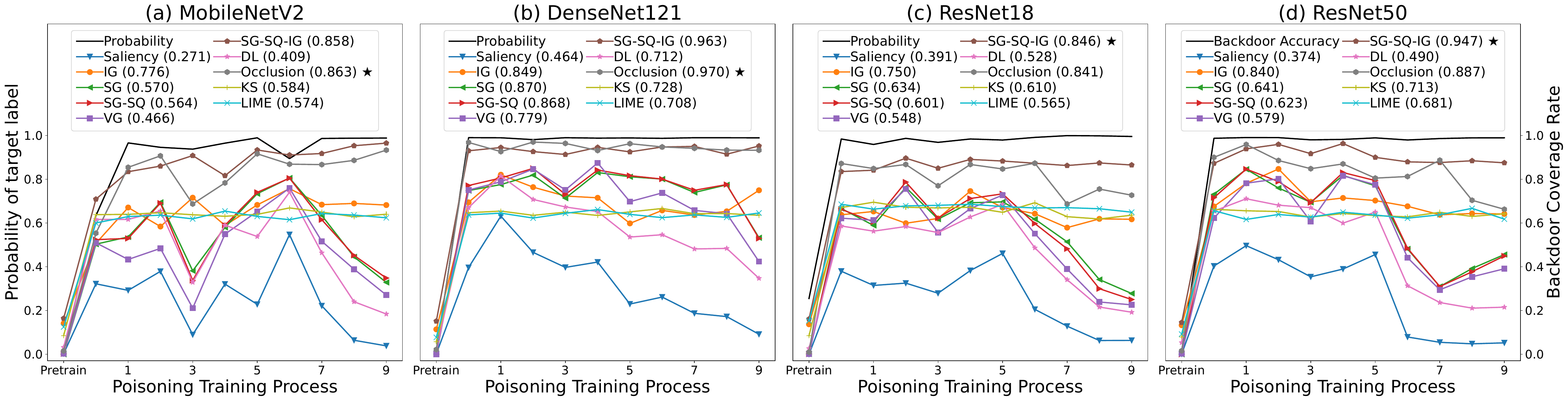}
	\caption{Results of EMBT on different model complexity. IG, SG-SQ-IG and Occlusion perform well in all four neural networks.}\label{fig:dynamic_model}
    \vspace{-3mm}
\end{figure*}

\begin{figure*}[h]
    \setlength{\abovecaptionskip}{3pt}
    \setlength{\belowcaptionskip}{0pt}
    \centering
	\includegraphics[width=1.0\textwidth]{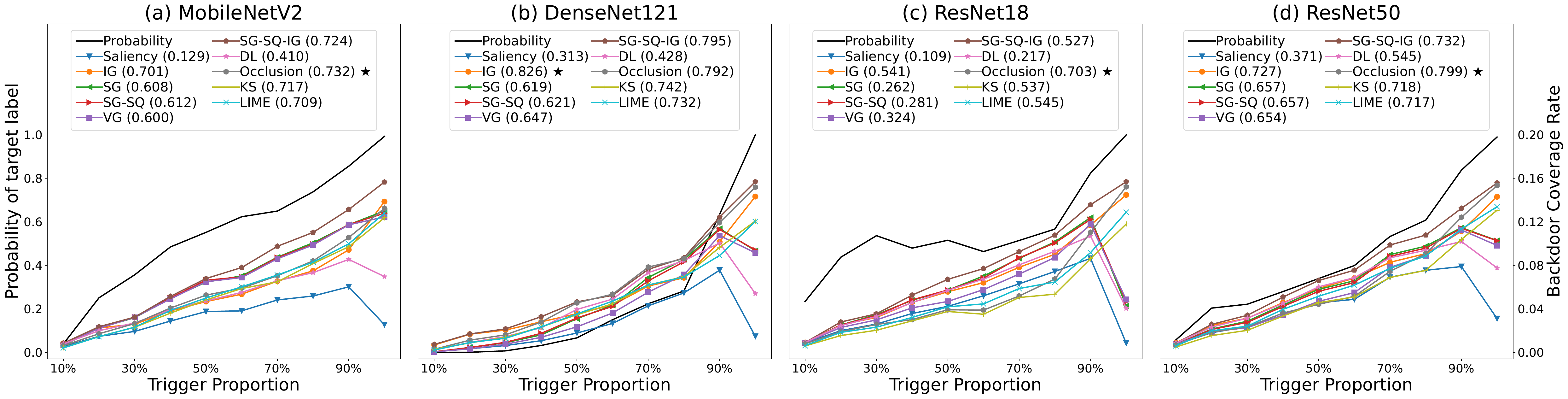}
	\caption{Results of PTT on different model complexity. IG, SG-SQ-IG, Occlusion, KS and LIME perform well.}\label{fig:dynamic_data}
    \vspace{-3mm}
\end{figure*}

\begin{figure*}[h]
    \setlength{\abovecaptionskip}{3pt}
    \setlength{\belowcaptionskip}{0pt}
    \centering
	\includegraphics[width=1.0\textwidth]{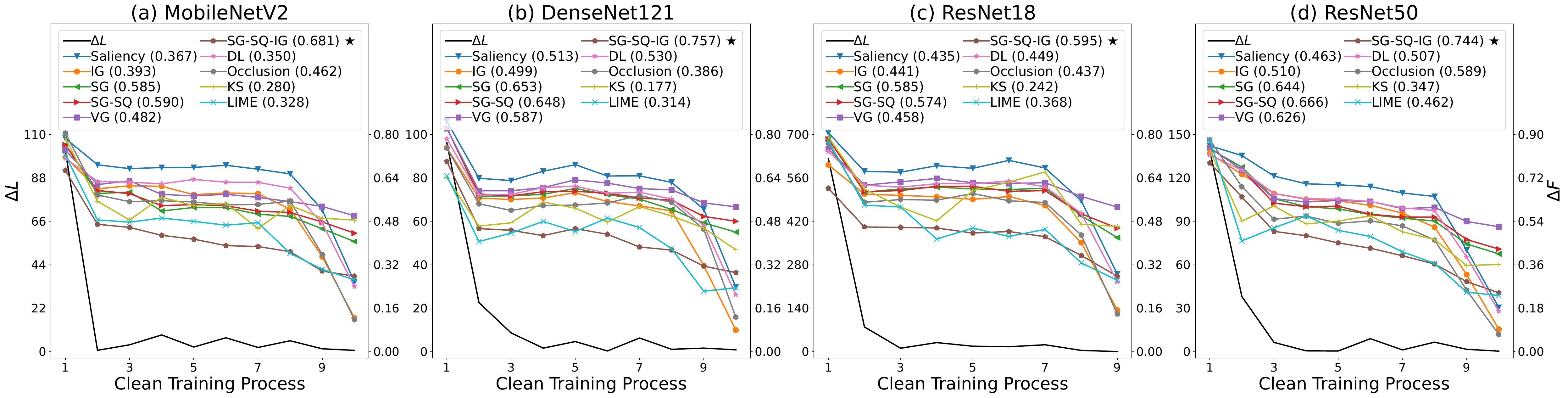}
	\caption{Results of EMT on different model complexity. SG-SQ-IG performs well in all four neural networks.}\label{fig:dynamic_loss}
    \vspace{-3mm}
\end{figure*}

\noindent\textbf{Model Complexity.} According to Hu \etal~\cite{modelcomplecitysurvey}, model complexity is affected by model type, the number of parameters, optimization algorithm, and data complexity. In this experiment, we have the same model framework (convolutional neural network, ReLu activation function), optimization algorithm, and data complexity. Thus, we use different numbers of parameters to characterize the complexity of the models. We use CIFAR-10 for evaluation and training different models, including MobileNetV2, ResNet18, ResNet50, and DenseNet121. 
The model information is shown in Table~\ref{tbl:model_complexity}.
The detailed trends are shown in Figire~\ref{fig:dynamic_model}, ~\ref{fig:dynamic_data} and ~\ref{fig:dynamic_loss}.
On EMBT, IG, SG, SG-SQ and SG-SQ-IG maintain a high degree of faithfulness, while IG and SG-SQ-IG keep a high degree of faithfulness on PTT. On EMT, IG and SG-SQ-IG have the highest faithfulness among all models. Similar to the experimental results of data complexity, IG, SG-SQ-IG and Occlusion perform well on all these model complexity tests, and have stable faithfulness. {The influence of model complexity is not as great as that of data complexity. 


\begin{figure}[t]
    \setlength{\abovecaptionskip}{4pt}
    \setlength{\belowcaptionskip}{0pt}
    \centering
	\includegraphics[width=0.35\textwidth]{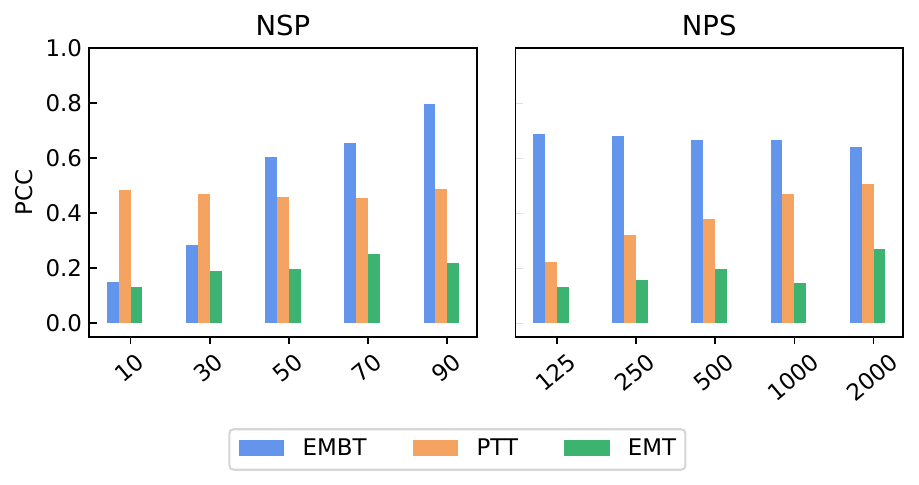}
	\caption{Results of trend tests on different parameters. NSP stands for the number of super-pixel segments. NPS stands for the number of generated perturbation samples.}\label{fig:radar_parameter_vis}
	\vspace{-4mm}
\end{figure}

\noindent\textbf{Parameters of Explanation Methods.} Some explanation methods rely on suitable parameter values to work. For example, the number of super-pixel segments and the number of generated perturbation samples are important parameters of LIME. They affect the results and efficiency. In this section, we use the number of super-pixel segments and the number of generated perturbation samples of LIME as examples to explore the effect of the parameters on faithfulness. We use ResNet18 trained on CIFAR-10 as the target model and then assess the faithfulness of LIME with different parameters. The results are shown in Figure~\ref{fig:radar_parameter_vis}. Both the number of super-pixel segments and the number of generated perturbation samples are basically in direct proportion to the faithfulness of the explanation results. However, when the number of super-pixel segments is over 70 or the number of generated perturbation samples is over 500, the increase in faithfulness is very small. Therefore, choosing the number of super-pixel segments as 70 and the number of perturbation samples as 500 is a better choice to balance the computational efficiency and faithfulness of LIME. From this experiment, we believe that trend tests can also be used as an automatic selection strategy for the parameters of the explanation methods.

\vspace{2mm}
\noindent\textit{Remark: Trend tests show that model complexity has less influence on faithfulness than data complexity. Parameters of explanation methods can affect their faithfulness. Our proposed trend tests can facilitate the selection of the optimal parameters for explanation methods.}

%% file: sections/application.tex
\begin{figure}[t]
    \setlength{\abovecaptionskip}{4pt}
    \setlength{\belowcaptionskip}{0pt}
    \centering
	\includegraphics[width=0.3\textwidth]{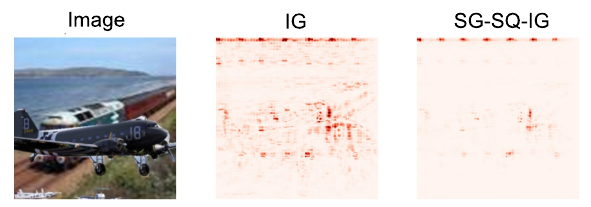}
	\caption{Examples of different explanations.}\label{fig:random_debug_vis}
	\vspace{-4mm}
\end{figure}

\section{Downstream Application: Model Debugging}\label{sec:application}

\begin{figure}[t]
    \setlength{\abovecaptionskip}{4pt}
    \setlength{\belowcaptionskip}{0pt}
    \centering
	\includegraphics[width=0.4\textwidth]{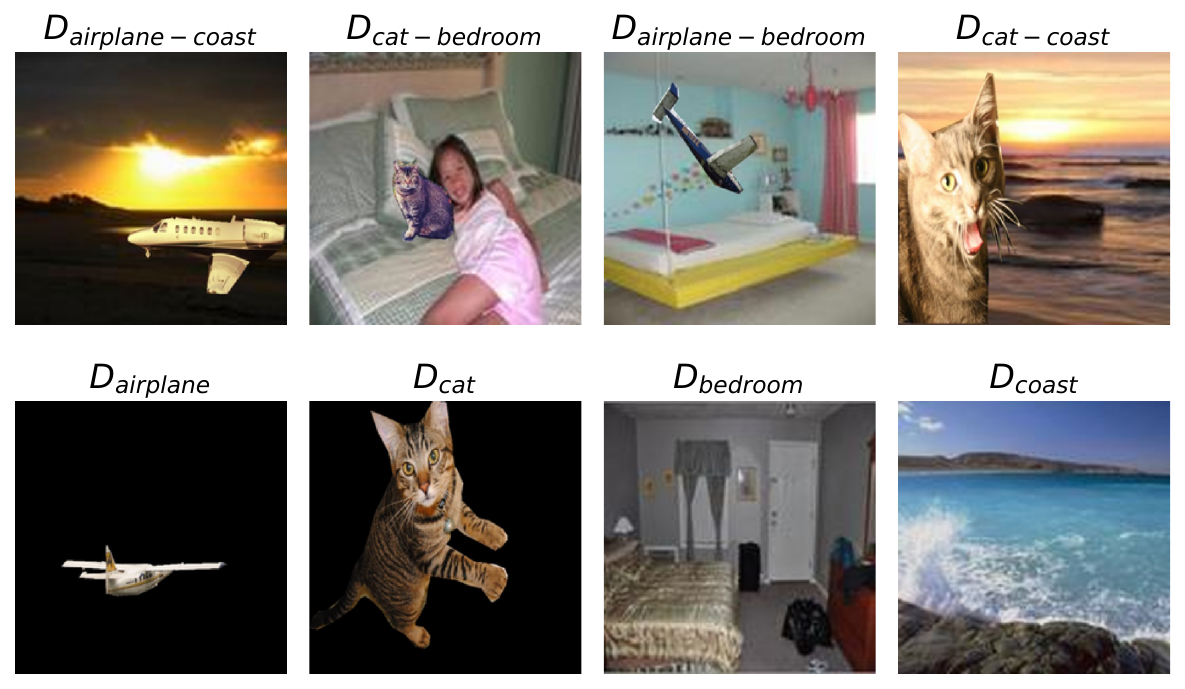}
	\caption{Examples of synthesized data in model debugging. $D_{airplane-coast}$ means the object is an airplane, and the context is the coast.}\label{fig:coco_data_vis}
	\vspace{-4mm}
\end{figure}

Explanation techniques can help build secure and trustworthy models, further promoting the widespread use of deep learning models in more security-critical fields. Model debugging is one of the ways to uncover spurious correlations learned by the model and help the users improve their models. For example, consider a classification task where all the airplanes in the dataset always appear together with the background (i.e., the blue sky). The model might then correlate the background features of the blue sky with the airplane category during training. This spurious correlation indicates that the model learns different category knowledge from what users envision, making the model vulnerable and insecure. If the users can detect the spurious correlation, they could enlarge the data space or deploy a stable deep learning module during training~\cite{DSL}. 
However, as shown in Section~\ref{sec:measurement}, explanation methods vary in performance. For example, in Figure~\ref{fig:random_debug_vis}, IG considers that the model focuses on both the object and background, while SG-IG-SQ marks the blue sky background as the important feature. We could not ensure which explanation is more conformed to the model.

In this section, we verify the effectiveness of our trend tests on guiding users to choose an explanation method and then examine the performance of explanation methods on detecting spurious correlations. Based on Adebayo \etal~\cite{modeldebug}, we construct a model with known spurious correlation and use the trend tests on the model to observe the faithfulness of each explanation method. Then, we analyze whether the explanation result focuses on the spurious correlated features. Next, we could verify whether the results of the trend tests are consistent with the results of the debugging test. We extract the object of cats and planes from MSCOCO 2017~\cite{MSCOCO}, and then replace the backgrounds with the bedroom and the coast from MiniPlaces~\cite{miniplaces}, respectively. We synthesize eight types of data as shown in Figure~\ref{fig:coco_data_vis}. $D_{airplane-coast}$ means the object is an airplane, and the context is the coast. Each of them includes $1000$ pictures. We use the first two ($D_{airplane-coast}$ and $D_{cat-bedroom}$) to train a ResNet18 model. We split the training data into a training set and a validation set at a ratio of 8:2. The rest are used for testing. The accuracy of the model is shown in Table~\ref{eva:debug_model}.

\begin{table}[t]
\setlength{\abovecaptionskip}{4pt}
\setlength{\belowcaptionskip}{0pt}
\caption{Accuracy of the model used in model debugging. The dataset order corresponds to the label index.}\label{eva:debug_model} 
\centering
\scriptsize
\begin{tabular}{cc}
\toprule
\textbf{Category} & \textbf{Accuracy} \\ \midrule
$D_{airplane-coast}$ and $D_{cat-bedroom}$ & 96.65\% \\
$D_{cat-coast}$ and $D_{airplane-bedroom}$ & 64.55\% \\
$D_{airplane}$ and $D_{cat}$ & 58.45\% \\
$D_{coast}$ and $D_{bedroom}$ & 90.10\% \\ \bottomrule
\end{tabular}
\vspace{-4mm}
\end{table}

\begin{figure}[t]
    \setlength{\abovecaptionskip}{4pt}
    \setlength{\belowcaptionskip}{0pt}
    \centering
	\includegraphics[width=0.18\textwidth]{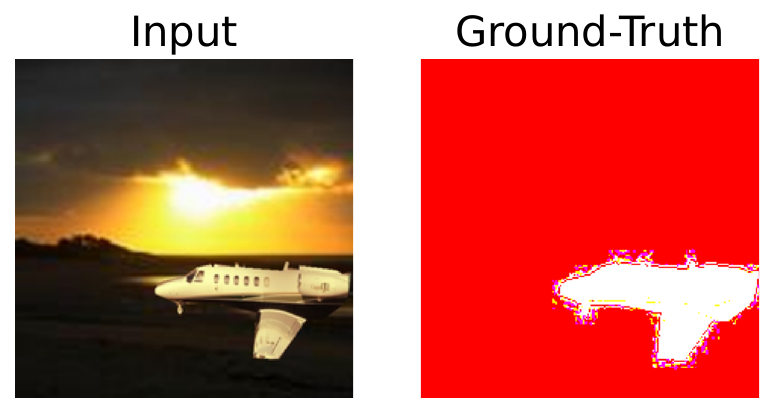}
	\caption{Example of ground-truth important features' mask. The white pixels are the location of ground-truth important features.}\label{fig:coco_mask_vis}
	\vspace{-4mm}
\end{figure}



As seen from Table~\ref{eva:debug_model}, although the model has high accuracy on $D_{airplane-coast}$ and $D_{cat-bedroom}$, the accuracy on the context ($D_{coast}$ and $D_{bedroom}$) is higher than the objects ($D_{airplane}$ and $D_{cat}$), indicating that the relative importance of the context is higher than that of the object. Therefore, we define the ground-truth important features of the model as the context features, as shown in Figure~\ref{fig:coco_mask_vis}. Note that the model may utilize both context and object features, but when taking the top 10\% important features, it should consist mainly of the context features. We use the proposed trend tests on this model. 
SG-IG-SQ outperformed IG in the trend-based faithfulness tests. In addition, we report the structural similarity index (SSIM)~\cite{ssim} scores between the explanation results and the ground-truth mask, which is widely used for capturing the visual similarity between the two images~\cite{modeldebug}. A high SSIM score implies a high visual similarity. SG-IG-SQ has a high SSIM score which is 0.8112, while the SSIM score of IG is 0.2453. We can also see in Figure~\ref{fig:random_debug_vis} that SG-IG-SQ correctly marks the blue sky as the important feature, while IG marks both the blue sky and the airplane as important features. The results of trend tests are consistent with the results of SSIM scores. It means that SG-IG-SQ is most promising to help users identify the spurious correlation problem in this model. From this experiment, we could empirically confirm that the trend tests can help users to select better explanation methods, which can further help to build secure and trustworthy models.


%% file: sections/relatedwork.tex
\section{Related work}

\subsection{Faithfulness of Explanation Methods}
The faithfulness assessment can be categorized into two classes: human-understandable and model-faithful. The human-understandable assessments include evaluating the explanations in terms of human cognition~\cite{humanintheloop,humanstudy,EuroSPSurvey}, and assessing human utilization of the explanations~\cite{interpretabledecisionsets,humanaccuracy}. These assessments have a hidden prerequisite: model cognition is consistent with human cognition. Unfortunately, the current exposure of model safety issues reveals the gap between model cognition and human cognition~\cite{yingzhesurvey}. The model may learn statistical bias or uncorrelated features in the data~\cite{aenotbug,noiseorsignal}.
The traditional model-faithful assessment is to modify the important features tagged by the explanations and observe the changes in the model's output~\cite{RealtimeSaliency,LEMNA,EuroSPSurvey,appofinterpretability}. The closest model-faithful assessments to our study are some that require retraining or creating a series of trends. ROAR~\cite{retrain} proposes to retrain the model by erasing the important features tagged by the explanations. However, even if the erased features are important features, the model may use the remaining weak statistical features to maintain high accuracy. Julius \etal~\cite{sanitycheck} propose randomization tests that randomize the model parameters layer by layer to observe changes in the explanations.
\emph{In this paper, we implement the traditional assessment and find that they may encounter the random dominance problem. To overcome this limitation, we propose three trend tests with the basic idea of verifying how well the trends of known data or models are consistent with the trends of explanations.}

\subsection{Robustness of Explanation Methods} {Zhang \etal ~\cite{attackexplanation} present that explanation methods are fragile when facing adversarial perturbations, leading to many efforts to assess the robustness of explanation methods. The robustness of explanation methods includes: (1) perturbing unimportant features has a small effect on model prediction; (2) perturbing important features can easily change model prediction even if the perturbation is small. Hsieh \etal ~\cite{greedyrobustness} propose Robustness-S to evaluate explanation methods and design a new search-based explanation method, Greedy-AS. Gan \etal ~\cite{stablexai} propose the Median Test for Feature Attribution to evaluate and improve the robustness of explanation methods. Traditional tests are used in the paper, which may also suffer from the random dominance problem. Fan \etal ~\cite{robustnessxai} conduct a robustness assessment with metamorphic testing. They also utilize a backdoor to construct ground-truth explanation results, but the model may not learn all backdoor features and introduce errors. \emph{The above methods necessitate sample perturbation. Although some of them strive to synthesize natural perturbations, it cannot be guaranteed that the perturbed samples are within the model's distribution. In trend tests, we avoid the adversarial effect by evolving the model or data to ensure that the test sample is in-distribution.}}

%% file: sections/discussion.tex
\section{Discussion}

\subsection{Solution to random dominance}
To overcome the random dominance problem, we insert backdoor triggers in a controlled manner, ensuring the presence of specific features in the training data~\cite{neuralcleanse}. This approach makes it more likely for the model to identify these features and reduces the impact of random noise. By including backdoor data as part of the in-distribution data, we mitigate the influence of OOD samples that may cause random dominance. Consequently, using backdoor data in trend tests allows us to effectively evaluate the faithfulness of explanation methods in identifying targeted features and avoid the issue of random dominance that can invalidate traditional tests.

\subsection{Stable explanations and adversarial attacks}
Explanations play a vital role in enhancing the transparency of deep learning models but can be vulnerable to adversarial attacks, leading to incorrect or misleading explanations. These attacks aim to manipulate or distort explanations by perturbing the input within a small range while maintaining the model output label. To address this issue, researchers have developed stable explanations that provide formal guarantees under small input perturbations, such as Anchor~\cite{anchor}, ensuring consistent explanations under adversarial conditions. However, stable explanations do not necessarily address faithfulness, which is a different aspect. There could be cases where explanations are stable but not faithful. Our analysis in Appendix~\ref{sec:stability} reveals that most explanation methods are susceptible to adversarial attacks. While more faithful methods require a larger perturbation budget, they can still be manipulated by adversarial attacks within a range of imperceptible perturbations to humans. In our experiments, we find that Anchor, which has a formal guarantee for stability, and LIME both exhibit stability on the CIFAR-10 dataset. However, their faithfulness in trend tests is relatively low. These findings emphasize that future research should focus on creating stable and faithful explanations.

\subsection{Limitations and benefits}
Although our new trend tests are superior in measuring the faithfulness of explanation methods, they require more computing time and data storage than traditional methods. EMBT and EMT need to save intermediate models during training, and PTT needs to generate more explanation results using more inputs. The extra time and storage depend on the number of ``checkpoints'' in the trend. Based on our evaluation, 5-10 checkpoints are sufficient for evaluation. Note that some traditional tests (\eg, augmentation) also need to synthesize more than one input (\eg, 5-15) to calculate faithfulness, which is similar to PTT.
Additionally, the results may be threatened by the success rate of the backdoor, especially in EMBT and PTT. 
Oftentimes, designing a textual trigger for a language model is more difficult than a graphical one for an image classifier. That motivates us to train a backdoored model with a high backdoor success rate to avoid the noise. All the backdoors can achieve a high success rate in our evaluation.
Explanations can be used in a wide range of applications, which include but are not limited to explaining model decisions~\cite{appofinterpretability}, understanding adversarial attacks~\cite{interpretadversarialattack} and defenses~\cite{explanationadversarialrobustness}, \etc. Further, by assessing faithfulness, consistency between explanation methods, models, and humans can be achieved.

%% file: sections/conclusion.tex
\section{Conclusion}
We propose three trend-based faithfulness tests to solve the random dominance phenomenon encountered by traditional faithfulness tests. Our tests enable the assessment of the explanation methods on complex data and can be applied to multiple types of models such as image, natural language and security applications. We implement the system and evaluate ten popular explanation methods. We find that the complexity of data does impact the explanation results of some methods. IG and SG-IG-SQ work very well on different datasets. However, the model complexity does not have much impact. These unprecedented discoveries could inspire future research on DL explanation methods. Finally, we verify the effectiveness of trend-based tests using a popular downstream application, model debugging. For a given DL model, trend tests recommend explanation methods with higher faithfulness to better debug the model, making it secure and trustworthy.

%% file: sections/appendix.tex
\section*{Appendix}\label{sec:appendix}
\begin{appendices}

\section{Datasets and hyperparameter settings of models}\label{subsec:introduction_dataset_models}

\noindent\textbf{MNIST.} This is a written digit classification dataset that consists of $28\times28$ grayscale images of digits 0-9. It has a training set of 50,000 images and a test set of 10,000 images. We train it on ResNet18. We set the learning rate to 0.01, the momentum to 0.9 and iterate 5 times with an SGD optimizer.

\noindent\textbf{CIFAR-10.} This is a commonly used image classification dataset with ten categories, consisting of 50,000 training data and 10,000 test data. A data is an $32\times32\times3$ color image. We set the learning rate to 0.06, the momentum to 0.9. We train it on several model, including ResNet18, MobileNet and DenseNet for 200 epochs with an SGD optimizer.

\noindent\textbf{Tiny ImageNet.} Tiny ImageNet is a subset of the ImageNet dataset. It contains 100,000 color images of 200 classes downsampled to $64\times64$. Each class has 500 training images, 50 validation images, and 50 test images. We set learning rate to 0.001, the momentum to 0.9. We train a ResNet18 on this dataset with and SGD optimizer for 50 epochs.

\noindent\textbf{MSCOCO 2017.} We use the MSCOCO 2017 dataset to train an FCN-ResNet50, for the instance segmentation task. The dataset consists of more than 200,000 images and 80 object categories. We follow the guideline of PyTorch~\cite{PYTORCH} to create a subset of MSCOCO 2017 that includes 20 categories from the Pascal VOC dataset~\cite{VOC}. We employ the SGD optimizer with a learning rate of 1e-4 to train the model for 80 epochs.

\begin{table*}[]
\caption{Impact of data complexity on explanation methods assessed by traditional faithfulness tests.}\label{tab:impactofmodelcomplexity}
\scriptsize
\centering
\begin{tabular}{|c|c|c|c|c|c|c|c|}
\hline
\textbf{Dataset} &
  \textbf{Method} &
  \textbf{Reduction} &
  \textbf{Random} &
  \textbf{Synthesis} &
  \textbf{Random} &
  \textbf{Augmentation} &
  \textbf{Random} \\ \hline
\multirow{10}{*}{MNIST} &
  Saliency &
  20.81\% &
  \multirow{10}{*}{1.02\%} &
  25.25\% &
  \multirow{10}{*}{1.50\%} &
  7.03\% &
  \multirow{10}{*}{16.65\%} \\ \cline{2-3} \cline{5-5} \cline{7-7}
 & IG        & 58.09\% &  & 82.60\% &  & 21.84\% & \\ \cline{2-3} \cline{5-5} \cline{7-7} 
 & SG        & 19.36\% &  & 11.88\% &  & 5.32\%  & \\ \cline{2-3} \cline{5-5} \cline{7-7} 
 & SG-SQ     & 17.76\% &  & 8.87\%  &  & 5.09\%  & \\ \cline{2-3} \cline{5-5} \cline{7-7} 
 & VG        & 11.51\% &  & 2.36\%  &  & 2.38\%  & \\ \cline{2-3} \cline{5-5} \cline{7-7} 
 & SG-SQ-IG  & 57.16\% &  & 82.32\% &  & 23.91\% & \\ \cline{2-3} \cline{5-5} \cline{7-7} 
 & DL        & 52.68\% &  & 77.54\% &  & 16.89\% & \\ \cline{2-3} \cline{5-5} \cline{7-7} 
 & Occlusion & 75.71\% &  & 71.29\% &  & 19.00\% & \\ \cline{2-3} \cline{5-5} \cline{7-7} 
 & KS        & 62.74\% &  & 61.00\% &  & 13.51\% & \\ \cline{2-3} \cline{5-5} \cline{7-7} 
 & LIME      & 62.26\% &  & 60.88\% &  & 12.96\% & \\ \hline
\multirow{10}{*}{CIFAR-10} &
  Saliency &
  53.11\% &
  \multirow{10}{*}{64.58\%} &
  0.81\% &
  \multirow{10}{*}{0.22\%} &
  1.08\% &
  \multirow{10}{*}{1.62\%} \\ \cline{2-3} \cline{5-5} \cline{7-7}
 & IG        & 51.37\% &  & 1.94\%  &  & 1.67\%  & \\ \cline{2-3} \cline{5-5} \cline{7-7}
 & SG        & 45.66\% &  & 5.13\%  &  & 1.48\%  & \\ \cline{2-3} \cline{5-5} \cline{7-7}
 & SG-SQ     & 46.47\% &  & 3.90\%  &  & 1.06\%  & \\ \cline{2-3} \cline{5-5} \cline{7-7}
 & VG        & 48.98\% &  & 2.99\%  &  & 1.13\%  & \\ \cline{2-3} \cline{5-5} \cline{7-7}
 & SG-SQ-IG  & 43.16\% &  & 5.72\%  &  & 1.96\%  & \\ \cline{2-3} \cline{5-5} \cline{7-7}
 & DL        & 47.52\% &  & 0.64\%  &  & 1.48\%  & \\ \cline{2-3} \cline{5-5} \cline{7-7}
 & Occlusion & 40.63\% &  & 4.82\%  &  & 1.15\%  & \\ \cline{2-3} \cline{5-5} \cline{7-7}
 & KS        & 30.03\% &  & 15.26\% &  & 1.51\%  & \\ \cline{2-3} \cline{5-5} \cline{7-7}
 & LIME      & 27.15\% &  & 13.10\% &  & 1.70\%  & \\ \hline
\multirow{10}{*}{Tiny ImageNet} &
  Saliency &
  68.46\% &
  \multirow{10}{*}{67.35\%} &
  0.85\% &
  \multirow{10}{*}{0.00\%} &
  1.16\% &
  \multirow{10}{*}{0.47\%} \\ \cline{2-3} \cline{5-5} \cline{7-7}
 & IG        & 67.25\% &  & 0.44\%  &  & 0.77\%  & \\ \cline{2-3} \cline{5-5} \cline{7-7}
 & SG        & 61.79\% &  & 3.29\%  &  & 0.06\%  & \\ \cline{2-3} \cline{5-5} \cline{7-7}
 & SG-SQ     & 61.09\% &  & 3.02\%  &  & 0.13\%  & \\ \cline{2-3} \cline{5-5} \cline{7-7}
 & VG        & 60.94\% &  & 3.02\%  &  & 0.17\%  & \\ \cline{2-3} \cline{5-5} \cline{7-7}
 & SG-SQ-IG  & 60.67\% &  & 4.89\%  &  & 0.75\%  & \\ \cline{2-3} \cline{5-5} \cline{7-7}
 & DL        & 62.72\% &  & 0.48\%  &  & 0.82\%  & \\ \cline{2-3} \cline{5-5} \cline{7-7}
 & Occlusion & 63.55\% &  & 2.49\%  &  & 0.14\%  & \\ \cline{2-3} \cline{5-5} \cline{7-7}
 & KS        & 43.03\% &  & 10.99\% &  & 0.03\%  & \\ \cline{2-3} \cline{5-5} \cline{7-7}
 & LIME      & 43.01\% &  & 10.83\% &  & 0.05\%  & \\ \hline
\end{tabular}
\end{table*}

\noindent\textbf{IMDB.} This is commonly used for text analysis for natural language processing and consists of 50,000 movie reviews labeled with positive or negative sentiment tendencies. Both the training and test set size are 25,000. We use a simple bidirectional LSTM for training, set the embedding dimension to 100 and the size of hidden layer to 256. We train the Bi-LSTM with an Adam optimizer for 50 epochs.

\noindent\textbf{Mimicus.} We follow the method of Saxe \etal \cite{PDFMalwareClassifier}, which extracts the macro features and structural features in the PDF document to train a PDF malware classifier composed of a three-layer fully connected neural network. The dataset contains 5,000 positive samples and 4,999 negative samples. Smutz \etal~\cite{MimicusDataset} extracts 135 binary features from this dataset. The complete feature list can be accessed on \cite{Mimicus}. We set the size of all hidden layers to 32 and train with an Adam optimizer for 100 epochs.

\noindent\textbf{DAMD.} Based on the work of Warnecke~\cite{EuroSPSurvey}, we implement an Android malware classifier. The dataset is from Malware Genome
Project~\cite{DAMDSOURCE} and has been processed into raw Dalvik bytecode. DAMD consists of 2,123 applications, including 863 benign and 1,260 malicious samples. We split the dataset into training and test sets in a ratio of 75:25. The model includes an embedding layer, a convolutional layer and two fully connected layers. We set the embedding size to 8, the size of the output channel to 64, the kernel size to 8, the hidden layer size to 64 and 16. The maximum data length is 150,000. We train the model with an Adam optimizer for 50 epochs.

\noindent\textbf{VulDeePecker.} Automated vulnerability detection is an important security application. Based on the work of Li \etal~\cite{Vuldeepecker}, we use CWE-119 data set disclosed by \etal~\cite{Vuldeepecker} to train a bidirectional LSTM model for vulnerability detection. There are 39,753 code segments in the data set, including 10,440 positive samples and 29,313 negative samples. We set the maximum sequence length to 50 in clean model (100 in backdoor model), the word embedding dimension to 200 and train for 100 iteration with an Adam optimizer.

\section{Different proportion of important features}\label{subsec:diff_proportion_important_features_vis}

In order to eliminate the influence on the proportion of important features retained, we take different proportions of important features for the reduction test. 
As shown in Figure~\ref{fig:different_proportion_reduction_test}, the reduction test samples made from 2\%-10\% of the important features are not as effective as the random samples. 

\begin{figure}[h]
    \setlength{\abovecaptionskip}{4pt}
    \setlength{\belowcaptionskip}{0pt}
    \centering
	\includegraphics[width=0.4\textwidth]{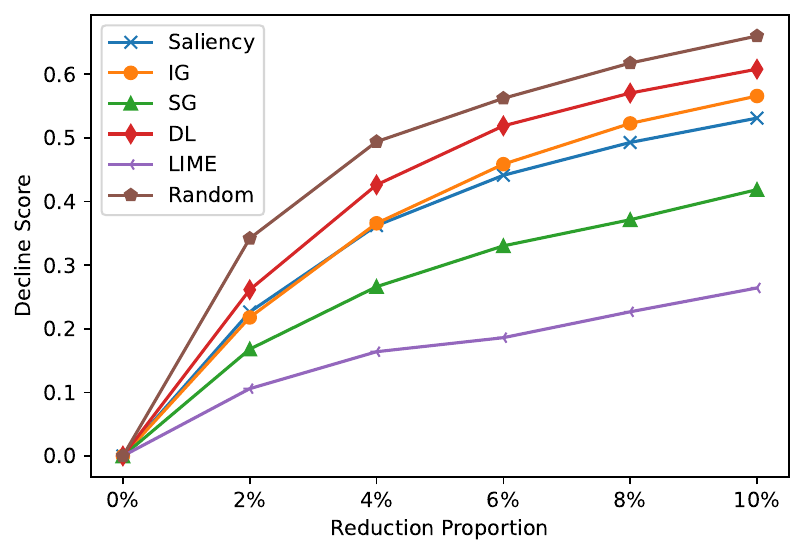}
	\caption{Different proportions of important features tagged by the explanation and random selected features in the reduction test. Important features tagged by explanations perform worse than random selected features in the reduction test.}\label{fig:different_proportion_reduction_test}
\end{figure}

\section{Parameter settings of trend tests}\label{subsec:para_trend_tests}
In this section, we detail the parameter settings for trend tests in our experiments. For all PTT experiments, we used the same sequence of backdoor features' ratios, ranging from 10\% to 100\%, with a 10\% increment each time. EMBT and EMT require setting the number of intermediate models ($n$) and interval ($c$), specified in Table~\ref{tbl:para_setting_trend_test}. Parameter variations across datasets are due to differing training iteration counts and the goal of aligning known trends with assumed trends in Section 3.2 for more accurate and representative trend tests. The parameter choices are flexible. Similar results can be obtained when the known trends under these parameters align with the assumed trends.

Although rare, models may exhibit instability during training~\cite{unstabletraining}, and outliers can impact the accuracy of PCC and overall trend tests. To address this issue and enhance fairness, we exclude anomalous models deviating significantly from the expected trend and replace them with neighboring models. For example, during training on the VulDeePecker, we observed occasional significant fluctuations in loss values. As a result, we implemented a filtering mechanism to retain intermediate models with lower loss values than their predecessors and discard those with unstable values.

The filtering mechanism mitigates training instability. By applying this filtering process, the models used for subsequent analysis are of higher quality and better represent the true trends in the data and models, ensuring fairness in our evaluations. Our filtering criteria focus on excluding models with significant deviations from the expected trend, providing a fair approach to selecting the most representative models for our trend tests.

\begin{table}[]
\setcounter{table}{7}
\centering
\scriptsize
\caption{Detailed parameter settings of EMBT and EMT. $n$ is the number of intermediate models that we choose. $c$ is the interval between the two intermediate models that we choose.}
\label{tbl:para_setting_trend_test}
\begin{tabular}{|c|c|c|c|c|c|c|}
\hline
Test & Dataset & n & c & Dataset & n & c \\ \hline
\multirow{4}{*}{EMBT} & MNIST & 5 & 50 batches & IMDB & 6 & 1 epoch \\ \cline{2-7} 
 & CIFAR-10 & 11 & 20 epochs & Mimicus & 5 & 50 batches \\ \cline{2-7} 
 & Tiny ImageNet & 9 & 1 epoch & DAMD & 5 & 1 epoch \\ \cline{2-7} 
 & COCO 2017 & 7 & 5 epochs & VulDeePecker & 5 & 4 epochs \\ \hline
Test & Dataset & n & c & Dataset & n & c \\ \hline
\multirow{4}{*}{EMT} & MNIST & 6 & 150 batches & IMDB & 7 & 5 epochs \\ \cline{2-7} 
 & CIFAR-10 & 10 & 20 epochs & Mimicus & 5 & 50 batches \\ \cline{2-7} 
 & Tiny ImageNet & 9 & 1 epoch & DAMD & 6 & 5 epochs \\ \cline{2-7} 
 & COCO 2017 & 6 & 5 epochs & VulDeePecker & 5 & 5 epochs \\ \hline
\end{tabular}
\end{table}

\section{Detailed results of experiment on data complexity}\label{subsec:detailed_results_data_complexity}

Table~\ref{tab:impactofmodelcomplexity} shows the detailed results of the traditional tests. The conclusion is consistent with the main text. The traditional tests perform well on MNIST. We can clearly see that IG, SG-SQ-IG and Occlusion perform better. As their reduction test, augmentation test and synthesis test are significantly different from the random control group. But on the more complex CIFAR-10 and Tiny ImageNet, the reduction test, augmentation test and synthesis test are about the same or even worse than the random control group. This may not be due to the low faithfulness of the explanation methods on complex data. Rather, the OOD problem faced by traditional tests may invalidate them on complex datasets.

\section{Adversarial attack on explanation methods}\label{sec:stability}

Based on previous studies, stable explanations ensure that if a given input is perturbed within $\epsilon$ and the model's output label remains unchanged, the corresponding explanations will stay stable~\cite{robustnessxai,stablexai,greedyrobustness}. However, stable explanations may not always guarantee faithfulness~\cite{stablexai}, as stable and faithful are two different properties of explanations. There could be the cases where explanations are stable but not faithful.

To investigate the relationship between stability and faithfulness in explanation methods, we conduct an adversarial attack on explanation methods, following Dombrowski et al.~\cite{manipulate}. For a given target explanation $I^t$, target model $\mathcal{F}$, and original data $X$, manipulated data $X_{ADV}$ should meet two properties: (1) the model outputs of $X_{ADV}$ and $X$ should be as similar as possible, i.e., $\mathcal{F}(X)\approx\mathcal{F}(X_{ADV})$, and (2) the explanation results of $X_{ADV}$ and the target explanation $I^t$ should be as similar as possible, i.e., $I(\mathcal{F},X_{ADV})\approx I^t$. We achieve this manipulation attack by optimizing the following objective function:
$$\gamma_1\Vert\mathcal{F}(X)-\mathcal{F}(X_{ADV})\Vert^2+\gamma_2 \Vert I^t-I(\mathcal{F},X_{ADV})\Vert^2,$$

\noindent where $\gamma_1$ and $\gamma_2$ are adjustable parameters controlling the balance between the two terms. The first term aims to minimize the difference between the model outputs of $X_{ADV}$ and $X$, while the second term focuses on minimizing the difference between the explanation results of $X_{ADV}$ and the target explanation $I^t$.

\begin{figure*}[t]
    \setcounter{figure}{29}
    \centering
	\includegraphics[width=0.9\textwidth]{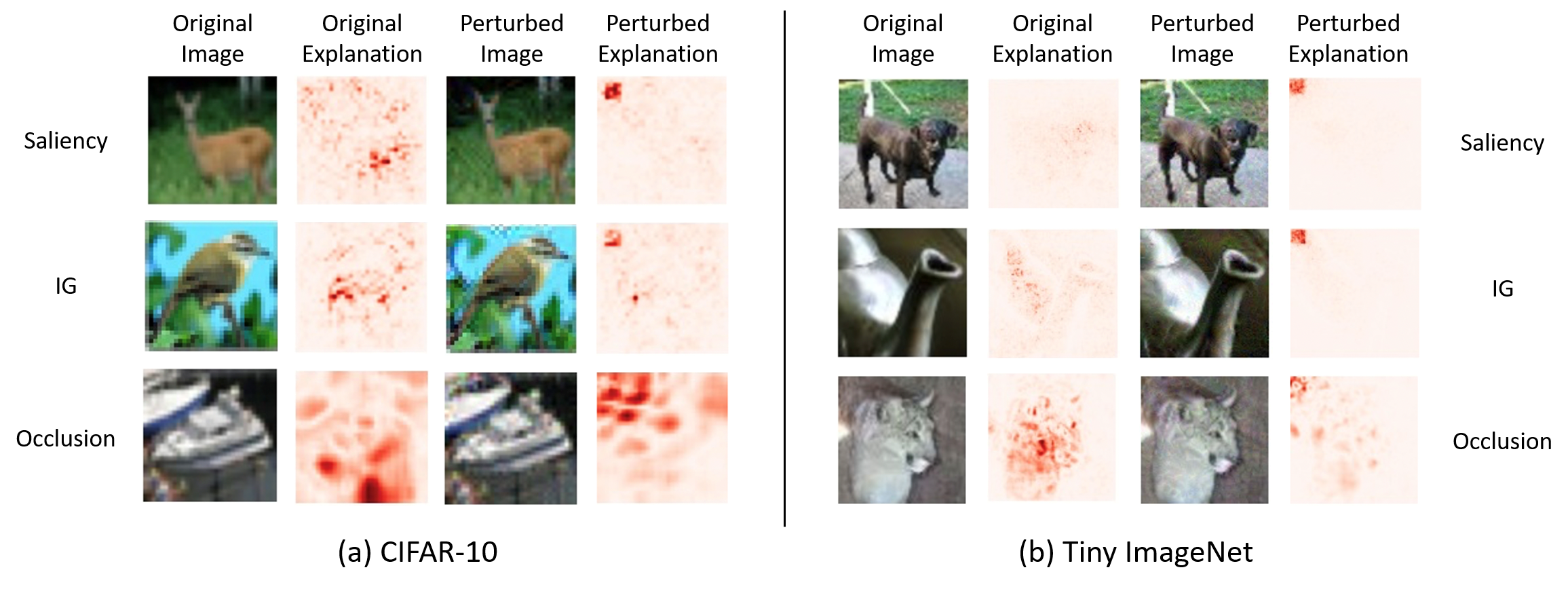}
	\caption{Examples of adversarial attacks: In the case of CIFAR-10 and Tiny-ImageNet, the targeted explanations focus on identifying important features as $4\times4$ and $24\times24$ squares in the upper left corner, respectively.}\label{fig:vis_manipulate_attack}
\end{figure*}

\begin{figure*}[t]
    \centering
	\includegraphics[width=0.7\textwidth]{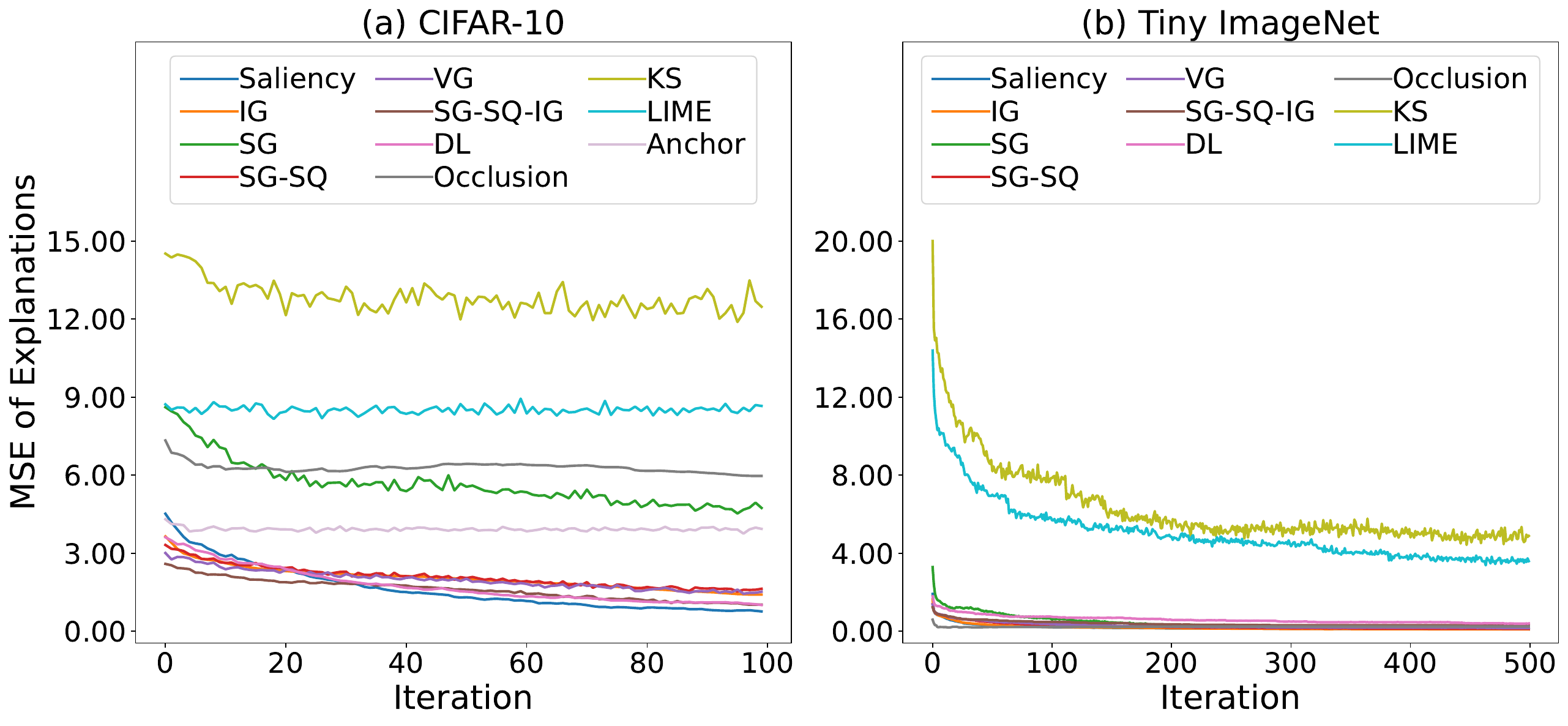}
 	\caption{MSE between explanation results and target explanations. A lower MSE means a higher similarity. Black-box explanation methods are harder to manipulate than white-box explanation methods. Most explanation methods can be manipulated, except Anchor and LIME in CIFAR-10.}\label{fig:manipulate_exp_loss}
\end{figure*}

When attacking gradient-based explanation methods, it is essential to compute second-order derivatives ($\nabla I(\mathcal{F},X_{ADV})$) for the model input. However, ReLU's second-order derivatives are 0, resulting in a gradient vanishing issue during optimization. To tackle this problem, we replace the ReLU layers with softplus layers~\cite{manipulate}, defined as: 
$$softplus_\beta(X)=\beta^{-1}log(1+e^{\beta X}).$$ 
The softplus function is a smooth approximation of the ReLU function, with the approximation accuracy controlled by the $\beta$ parameter. Larger $\beta$ values provide more accurate ReLU approximations. In our experiments, we find that $\beta=30$ yields effective attack results. Since some explanation methods are non-differentiable, we follow the approach of et al.~\cite{manipulate} and use perturbation data generated by Saliency or IG to attack them. In our manipulation attack, Grad and IG are attacked using gradient descent. For other methods, SG-SQ-IG is attacked with adversarial examples generated against IG, while the remaining methods are attacked using adversarial examples created against Saliency. Our targets include the ResNet18 models trained on CIFAR-10 and Tiny ImageNet, along with the previously mentioned ten explanation methods and Anchor~\cite{anchor}, an explanation method with a formal guarantee for stability.

\begin{figure*}[h]
    \centering
	\includegraphics[width=0.7\textwidth]{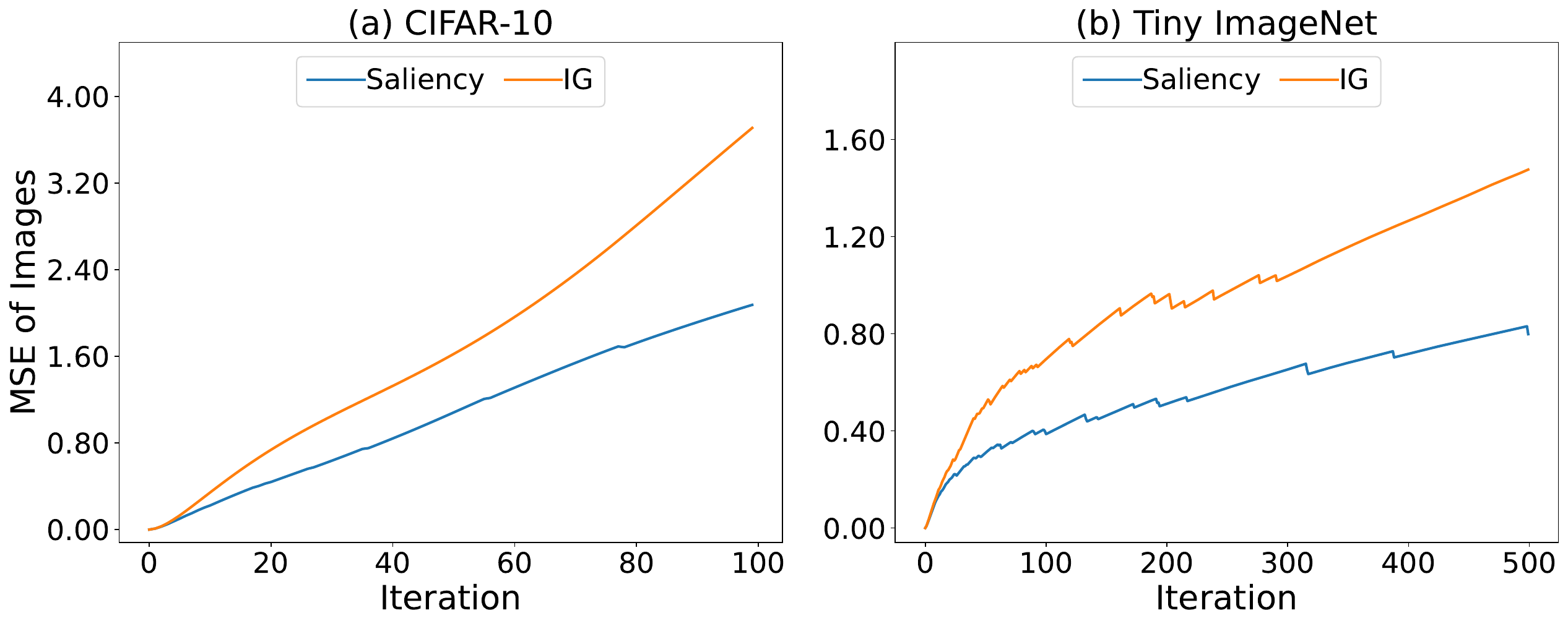}
	\caption{MSE between perturbed images and original images. A lower MSE means a higher similarity. Manipulating IG causes a greater perturbation in the image than Saliency.}\label{fig:manipulate_img_loss}
\end{figure*}

Figure~\ref{fig:vis_manipulate_attack} illustrates examples of our attack. There are several important parameters when we implement a manipulation attack. We use the Adam optimizer with a learning rate of 0.01, set $\gamma_1$ to 100, and $\gamma_2$ to $10^7$. The attack's iteration count is 100 for CIFAR-10 and 500 for Tiny ImageNet. In our target explanation, we aim to identify important features in the form of a square located at the top left corner of the data. For CIFAR-10, we use a $4\times4$ square, whereas for Tiny ImageNet, we employ a larger $24\times24$ square. These parameter settings aim to strike a balance between attack effectiveness and computational efficiency. During the manipulation attack, we measure the mean squared error (MSE) between the explanation results of the manipulated data and the target explanations, as well as the MSE between the manipulated data and the original data. Our results are presented in Figure~\ref{fig:manipulate_exp_loss} and Figure~\ref{fig:manipulate_img_loss}, where a smaller MSE indicates greater similarity.

Our experiments demonstrate that most explanation methods can be manipulated by adversarial attacks. As shown in Figure~\ref{fig:manipulate_exp_loss}, the MSE between explanation results and target explanations is initially dissimilar when the iteration is 0, which can be attributed to the different results produced by distinct explanation methods. As the adversarial attack progresses iteratively, the MSE of explanations decreases, indicating a convergence between the explanation results of perturbed images and the target explanations. It is worth noting that only Saliency and IG are attacked using gradient descent, while SG-SQ-IG employs IG's adversarial samples, and the remaining explanation methods use Saliency's adversarial samples. Despite these differences, the attack is generally successful throughout the iterative process, except for Anchor~\cite{anchor}, which has a formal guarantee for stability, and LIME on CIFAR-10. Both of these methods are stable but exhibit lower faithfulness in trend tests. The mean trend test values of them are 0.23 and 0.51, respectively. Interestingly, manipulating Tiny ImageNet seems easier than CIFAR-10, likely due to the more diverse features in Tiny ImageNet, which offer increased opportunities for manipulation. Figure~\ref{fig:manipulate_img_loss} reveals that IG, with higher faithfulness, results in a larger MSE between the perturbed image and the original image compared to Saliency with lower faithfulness. This suggests that manipulating IG is more challenging for adversarial attacks, as they require a larger perturbation budget. Although high faithfulness explanation methods demand a larger perturbation budget, they can still be manipulated by adversarial attacks without being noticeable to humans. Consequently, the development of an explanation method exhibiting both high faithfulness and high stability is an essential future research direction.

\end{appendices}

%% file: main.bbl

\begin{thebibliography}{76}


\ifx \showCODEN    \undefined \def \showCODEN     #1{\unskip}     \fi
\ifx \showDOI      \undefined \def \showDOI       #1{#1}\fi
\ifx \showISBNx    \undefined \def \showISBNx     #1{\unskip}     \fi
\ifx \showISBNxiii \undefined \def \showISBNxiii  #1{\unskip}     \fi
\ifx \showISSN     \undefined \def \showISSN      #1{\unskip}     \fi
\ifx \showLCCN     \undefined \def \showLCCN      #1{\unskip}     \fi
\ifx \shownote     \undefined \def \shownote      #1{#1}          \fi
\ifx \showarticletitle \undefined \def \showarticletitle #1{#1}   \fi
\ifx \showURL      \undefined \def \showURL       {\relax}        \fi
\providecommand\bibfield[2]{#2}
\providecommand\bibinfo[2]{#2}
\providecommand\natexlab[1]{#1}
\providecommand\showeprint[2][]{arXiv:#2}

\bibitem[Mim({[n.\,d.]})]%
        {Mimicus}
 \bibinfo{year}{[n.\,d.]}\natexlab{}.
\newblock \bibinfo{title}{{Mimicus}}.
\newblock \bibinfo{howpublished}{\url{https://github.com/srndic/mimicus}}.
\newblock


\bibitem[Achanta et~al\mbox{.}(2012)]%
        {slicsuperpixels}
\bibfield{author}{\bibinfo{person}{Radhakrishna Achanta}, \bibinfo{person}{Appu Shaji}, \bibinfo{person}{Kevin Smith}, \bibinfo{person}{Aurelien Lucchi}, \bibinfo{person}{Pascal Fua}, {and} \bibinfo{person}{Sabine Süsstrunk}.} \bibinfo{year}{2012}\natexlab{}.
\newblock \showarticletitle{SLIC Superpixels Compared to State-of-the-Art Superpixel Methods}. In \bibinfo{booktitle}{\emph{IEEE Transactions on Pattern Analysis and Machine Intelligence}}.
\newblock


\bibitem[Adebayo et~al\mbox{.}(2018a)]%
        {vargrad}
\bibfield{author}{\bibinfo{person}{Julius Adebayo}, \bibinfo{person}{Justin Gilmer}, \bibinfo{person}{Ian~J. Goodfellow}, {and} \bibinfo{person}{Been Kim}.} \bibinfo{year}{2018}\natexlab{a}.
\newblock \showarticletitle{Local Explanation Methods for Deep Neural Networks Lack Sensitivity to Parameter Values}. In \bibinfo{booktitle}{\emph{6th International Conference on Learning Representations, {ICLR}}}.
\newblock


\bibitem[Adebayo et~al\mbox{.}(2018b)]%
        {sanitycheck}
\bibfield{author}{\bibinfo{person}{Julius Adebayo}, \bibinfo{person}{Justin Gilmer}, \bibinfo{person}{Michael Muelly}, \bibinfo{person}{Ian Goodfellow}, \bibinfo{person}{Moritz Hardt}, {and} \bibinfo{person}{Been Kim}.} \bibinfo{year}{2018}\natexlab{b}.
\newblock \showarticletitle{Sanity Checks for Saliency Maps}. In \bibinfo{booktitle}{\emph{Proceedings of the 32nd International Conference on Neural Information Processing Systems}} \emph{(\bibinfo{series}{NIPS'18})}.
\newblock


\bibitem[Adebayo et~al\mbox{.}(2020)]%
        {modeldebug}
\bibfield{author}{\bibinfo{person}{Julius Adebayo}, \bibinfo{person}{Michael Muelly}, \bibinfo{person}{Ilaria Liccardi}, {and} \bibinfo{person}{Been Kim}.} \bibinfo{year}{2020}\natexlab{}.
\newblock \showarticletitle{Debugging Tests for Model Explanations}. In \bibinfo{booktitle}{\emph{34th International Conference on Neural Information Processing Systems}} \emph{(\bibinfo{series}{NIPS'20})}.
\newblock
\showISBNx{9781713829546}


\bibitem[Ahn et~al\mbox{.}(2022)]%
        {unstabletraining}
\bibfield{author}{\bibinfo{person}{Kwangjun Ahn}, \bibinfo{person}{Jingzhao Zhang}, {and} \bibinfo{person}{Suvrit Sra}.} \bibinfo{year}{2022}\natexlab{}.
\newblock \showarticletitle{Understanding the unstable convergence of gradient descent}. In \bibinfo{booktitle}{\emph{Proceedings of the 39th International Conference on Machine Learning}}. \bibinfo{publisher}{PMLR}.
\newblock
\urldef\tempurl%
\url{https://proceedings.mlr.press/v162/ahn22a.html}
\showURL{%
\tempurl}


\bibitem[Baldock et~al\mbox{.}(2021)]%
        {pccdl1}
\bibfield{author}{\bibinfo{person}{Robert John~Nicholas Baldock}, \bibinfo{person}{Hartmut Maennel}, {and} \bibinfo{person}{Behnam Neyshabur}.} \bibinfo{year}{2021}\natexlab{}.
\newblock \showarticletitle{Deep Learning Through the Lens of Example Difficulty}. In \bibinfo{booktitle}{\emph{Advances in Neural Information Processing Systems}}.
\newblock


\bibitem[Benesty et~al\mbox{.}(2009)]%
        {pearson}
\bibfield{author}{\bibinfo{person}{Jacob Benesty}, \bibinfo{person}{Jingdong Chen}, \bibinfo{person}{Yiteng Huang}, {and} \bibinfo{person}{Israel Cohen}.} \bibinfo{year}{2009}\natexlab{}.
\newblock \bibinfo{booktitle}{\emph{Pearson Correlation Coefficient}}.
\newblock \bibinfo{publisher}{Springer Berlin Heidelberg}, \bibinfo{address}{Berlin, Heidelberg}.
\newblock


\bibitem[Chen et~al\mbox{.}(2022)]%
        {uncovertrigger}
\bibfield{author}{\bibinfo{person}{T. Chen}, \bibinfo{person}{Z. Zhang}, \bibinfo{person}{Y. Zhang}, \bibinfo{person}{S. Chang}, \bibinfo{person}{S. Liu}, {and} \bibinfo{person}{Z. Wang}.} \bibinfo{year}{2022}\natexlab{}.
\newblock \showarticletitle{Quarantine: Sparsity Can Uncover the Trojan Attack Trigger for Free}. In \bibinfo{booktitle}{\emph{2022 IEEE/CVF Conference on Computer Vision and Pattern Recognition (CVPR)}}.
\newblock


\bibitem[Chen et~al\mbox{.}(2021)]%
        {badnl}
\bibfield{author}{\bibinfo{person}{Xiaoyi Chen}, \bibinfo{person}{Ahmed Salem}, \bibinfo{person}{Dingfan Chen}, \bibinfo{person}{Michael Backes}, \bibinfo{person}{Shiqing Ma}, \bibinfo{person}{Qingni Shen}, \bibinfo{person}{Zhonghai Wu}, {and} \bibinfo{person}{Yang Zhang}.} \bibinfo{year}{2021}\natexlab{}.
\newblock \showarticletitle{BadNL: Backdoor Attacks against NLP Models with Semantic-Preserving Improvements}. In \bibinfo{booktitle}{\emph{Annual Computer Security Applications Conference}} \emph{(\bibinfo{series}{ACSAC})}.
\newblock


\bibitem[Dabkowski and Gal(2017)]%
        {RealtimeSaliency}
\bibfield{author}{\bibinfo{person}{Piotr Dabkowski} {and} \bibinfo{person}{Yarin Gal}.} \bibinfo{year}{2017}\natexlab{}.
\newblock \showarticletitle{Real Time Image Saliency for Black Box Classifiers}. In \bibinfo{booktitle}{\emph{Advances in Neural Information Processing Systems}}.
\newblock


\bibitem[Das and Rad(2020)]%
        {ieeexaisurvey}
\bibfield{author}{\bibinfo{person}{Arun Das} {and} \bibinfo{person}{Paul Rad}.} \bibinfo{year}{2020}\natexlab{}.
\newblock \bibinfo{title}{Opportunities and Challenges in Explainable Artificial Intelligence (XAI): A Survey}.
\newblock
\newblock
\showeprint[arxiv]{2006.11371}~[cs.CV]


\bibitem[Dombrowski et~al\mbox{.}(2019)]%
        {manipulate}
\bibfield{author}{\bibinfo{person}{Ann-Kathrin Dombrowski}, \bibinfo{person}{Maximilian Alber}, \bibinfo{person}{Christopher~J. Anders}, \bibinfo{person}{Marcel Ackermann}, \bibinfo{person}{Klaus-Robert M\"{u}ller}, {and} \bibinfo{person}{Pan Kessel}.} \bibinfo{year}{2019}\natexlab{}.
\newblock \bibinfo{booktitle}{\emph{Explanations Can Be Manipulated and Geometry is to Blame}}.
\newblock \bibinfo{publisher}{Curran Associates Inc.}, \bibinfo{address}{Red Hook, NY, USA}.
\newblock


\bibitem[Doshi-Velez and Kim(2017)]%
        {appofinterpretability}
\bibfield{author}{\bibinfo{person}{Finale Doshi-Velez} {and} \bibinfo{person}{Been Kim}.} \bibinfo{year}{2017}\natexlab{}.
\newblock \bibinfo{title}{Towards A Rigorous Science of Interpretable Machine Learning}.
\newblock
\newblock
\showeprint[arxiv]{1702.08608}~[stat.ML]


\bibitem[Du et~al\mbox{.}(2020)]%
        {robustnessanalysisondl}
\bibfield{author}{\bibinfo{person}{Xiaoning Du}, \bibinfo{person}{Yi Li}, \bibinfo{person}{Xiaofei Xie}, \bibinfo{person}{Lei Ma}, \bibinfo{person}{Yang Liu}, {and} \bibinfo{person}{Jianjun Zhao}.} \bibinfo{year}{2020}\natexlab{}.
\newblock \showarticletitle{Marble: Model-based Robustness Analysis of Stateful Deep Learning Systems}. In \bibinfo{booktitle}{\emph{35th ACM International Conference on Automated Software Engineering (ASE)}}.
\newblock


\bibitem[Everingham et~al\mbox{.}({[n.\,d.]})]%
        {VOC}
\bibfield{author}{\bibinfo{person}{M. Everingham}, \bibinfo{person}{L. Van~Gool}, \bibinfo{person}{C.~K.~I. Williams}, \bibinfo{person}{J. Winn}, {and} \bibinfo{person}{A. Zisserman}.} \bibinfo{year}{[n.\,d.]}\natexlab{}.
\newblock
\newblock


\bibitem[Fan et~al\mbox{.}(2022)]%
        {robustnessxai}
\bibfield{author}{\bibinfo{person}{Ming Fan}, \bibinfo{person}{Jiali Wei}, \bibinfo{person}{Wuxia Jin}, \bibinfo{person}{Zhou Xu}, \bibinfo{person}{Wenying Wei}, {and} \bibinfo{person}{Ting Liu}.} \bibinfo{year}{2022}\natexlab{}.
\newblock \showarticletitle{One Step Further: Evaluating Interpreters Using Metamorphic Testing}. In \bibinfo{booktitle}{\emph{31st ACM SIGSOFT International Symposium on Software Testing and Analysis}} \emph{(\bibinfo{series}{ISSTA})}.
\newblock
\showISBNx{9781450393799}


\bibitem[Gan et~al\mbox{.}(2022)]%
        {stablexai}
\bibfield{author}{\bibinfo{person}{Yuyou Gan}, \bibinfo{person}{Yuhao Mao}, \bibinfo{person}{Xuhong Zhang}, \bibinfo{person}{Shouling Ji}, \bibinfo{person}{Yuwen Pu}, \bibinfo{person}{Meng Han}, \bibinfo{person}{Jianwei Yin}, {and} \bibinfo{person}{Ting Wang}.} \bibinfo{year}{2022}\natexlab{}.
\newblock \showarticletitle{"Is Your Explanation Stable?": A Robustness Evaluation Framework for Feature Attribution}. In \bibinfo{booktitle}{\emph{ACM SIGSAC Conference on Computer and Communications Security}} \emph{(\bibinfo{series}{CCS '22})}.
\newblock
\showISBNx{9781450394505}
\urldef\tempurl%
\url{https://doi.org/10.1145/3548606.3559392}
\showDOI{\tempurl}


\bibitem[Goodfellow et~al\mbox{.}({[n.\,d.]})]%
        {FGSM}
\bibfield{author}{\bibinfo{person}{Ian~J. Goodfellow}, \bibinfo{person}{Jonathon Shlens}, {and} \bibinfo{person}{Christian Szegedy}.} \bibinfo{year}{[n.\,d.]}\natexlab{}.
\newblock \showarticletitle{Explaining and Harnessing Adversarial Examples}. In \bibinfo{booktitle}{\emph{3rd International Conference on Learning Representations, {ICLR} 2015}}.
\newblock
\urldef\tempurl%
\url{http://arxiv.org/abs/1412.6572}
\showURL{%
\tempurl}


\bibitem[Graves et~al\mbox{.}(2013)]%
        {bilstm}
\bibfield{author}{\bibinfo{person}{Alex Graves}, \bibinfo{person}{Abdel-rahman Mohamed}, {and} \bibinfo{person}{Geoffrey Hinton}.} \bibinfo{year}{2013}\natexlab{}.
\newblock \showarticletitle{Speech Recognition with Deep Recurrent Neural Networks}.
\newblock \bibinfo{journal}{\emph{ICASSP, IEEE International Conference on Acoustics, Speech and Signal Processing - Proceedings}}  \bibinfo{volume}{38} (\bibinfo{date}{03} \bibinfo{year}{2013}).
\newblock


\bibitem[Gu et~al\mbox{.}(2017)]%
        {badnets}
\bibfield{author}{\bibinfo{person}{Tianyu Gu}, \bibinfo{person}{Brendan Dolan{-}Gavitt}, {and} \bibinfo{person}{Siddharth Garg}.} \bibinfo{year}{2017}\natexlab{}.
\newblock \showarticletitle{BadNets: Identifying Vulnerabilities in the Machine Learning Model Supply Chain}.
\newblock  (\bibinfo{year}{2017}).
\newblock


\bibitem[Guo et~al\mbox{.}(2018)]%
        {LEMNA}
\bibfield{author}{\bibinfo{person}{Wenbo Guo}, \bibinfo{person}{Dongliang Mu}, \bibinfo{person}{Jun Xu}, \bibinfo{person}{Purui Su}, \bibinfo{person}{Gang Wang}, {and} \bibinfo{person}{Xinyu Xing}.} \bibinfo{year}{2018}\natexlab{}.
\newblock \showarticletitle{LEMNA: Explaining Deep Learning Based Security Applications}. In \bibinfo{booktitle}{\emph{Proceedings of the ACM SIGSAC Conference on Computer and Communications Security}}.
\newblock
\showISBNx{9781450356930}


\bibitem[He et~al\mbox{.}(2016)]%
        {resnet}
\bibfield{author}{\bibinfo{person}{Kaiming He}, \bibinfo{person}{Xiangyu Zhang}, \bibinfo{person}{Shaoqing Ren}, {and} \bibinfo{person}{Jian Sun}.} \bibinfo{year}{2016}\natexlab{}.
\newblock \showarticletitle{Deep Residual Learning for Image Recognition}. In \bibinfo{booktitle}{\emph{2016 IEEE Conference on Computer Vision and Pattern Recognition (CVPR)}}. \bibinfo{pages}{770--778}.
\newblock
\urldef\tempurl%
\url{https://doi.org/10.1109/CVPR.2016.90}
\showDOI{\tempurl}


\bibitem[He et~al\mbox{.}(2020)]%
        {yingzhesurvey}
\bibfield{author}{\bibinfo{person}{Yingzhe He}, \bibinfo{person}{Guozhu Meng}, \bibinfo{person}{Kai Chen}, \bibinfo{person}{Xingbo Hu}, {and} \bibinfo{person}{Jinwen He}.} \bibinfo{year}{2020}\natexlab{}.
\newblock \showarticletitle{Towards Security Threats of Deep Learning Systems: A Survey}.
\newblock \bibinfo{journal}{\emph{IEEE Transactions on Software Engineering}} (\bibinfo{year}{2020}).
\newblock


\bibitem[Hendrycks et~al\mbox{.}(2019)]%
        {hendrycks2019selfsupervised}
\bibfield{author}{\bibinfo{person}{Dan Hendrycks}, \bibinfo{person}{Mantas Mazeika}, \bibinfo{person}{Saurav Kadavath}, {and} \bibinfo{person}{Dawn Song}.} \bibinfo{year}{2019}\natexlab{}.
\newblock \showarticletitle{Using Self-Supervised Learning Can Improve Model Robustness and Uncertainty}.
\newblock \bibinfo{journal}{\emph{Advances in Neural Information Processing Systems (NeurIPS)}} (\bibinfo{year}{2019}).
\newblock


\bibitem[Hooker et~al\mbox{.}(2019)]%
        {retrain}
\bibfield{author}{\bibinfo{person}{Sara Hooker}, \bibinfo{person}{Dumitru Erhan}, \bibinfo{person}{Pieter-Jan Kindermans}, {and} \bibinfo{person}{Been Kim}.} \bibinfo{year}{2019}\natexlab{}.
\newblock \showarticletitle{A Benchmark for Interpretability Methods in Deep Neural Networks}. In \bibinfo{booktitle}{\emph{Advances in Neural Information Processing Systems}}.
\newblock


\bibitem[Howard et~al\mbox{.}(2017)]%
        {MobileNets}
\bibfield{author}{\bibinfo{person}{Andrew~G. Howard}, \bibinfo{person}{Menglong Zhu}, \bibinfo{person}{Bo Chen}, \bibinfo{person}{Dmitry Kalenichenko}, \bibinfo{person}{Weijun Wang}, \bibinfo{person}{Tobias Weyand}, \bibinfo{person}{Marco Andreetto}, {and} \bibinfo{person}{Hartwig Adam}.} \bibinfo{year}{2017}\natexlab{}.
\newblock \showarticletitle{MobileNets: Efficient Convolutional Neural Networks for Mobile Vision Applications}.
\newblock  (\bibinfo{year}{2017}).
\newblock


\bibitem[Hsieh et~al\mbox{.}(2021)]%
        {greedyrobustness}
\bibfield{author}{\bibinfo{person}{Cheng-Yu Hsieh}, \bibinfo{person}{Chih-Kuan Yeh}, \bibinfo{person}{Xuanqing Liu}, \bibinfo{person}{Pradeep Ravikumar}, \bibinfo{person}{Seungyeon Kim}, \bibinfo{person}{Sanjiv Kumar}, {and} \bibinfo{person}{Cho-Jui Hsieh}.} \bibinfo{year}{2021}\natexlab{}.
\newblock \showarticletitle{Evaluations and Methods for Explanation through Robustness Analysis}. In \bibinfo{booktitle}{\emph{10th International Conference on Learning Representations, ICLR 2021}}.
\newblock


\bibitem[Hu et~al\mbox{.}(2021)]%
        {modelcomplecitysurvey}
\bibfield{author}{\bibinfo{person}{Xia Hu}, \bibinfo{person}{Lingyang Chu}, \bibinfo{person}{Jian Pei}, \bibinfo{person}{Weiqing Liu}, {and} \bibinfo{person}{Jiang Bian}.} \bibinfo{year}{2021}\natexlab{}.
\newblock \showarticletitle{Model Complexity of Deep Learning: A Survey}.
\newblock \bibinfo{journal}{\emph{Knowl. Inf. Syst.}} \bibinfo{volume}{63}, \bibinfo{number}{10} (\bibinfo{year}{2021}), \bibinfo{numpages}{35}~pages.
\newblock


\bibitem[Huang et~al\mbox{.}(2017)]%
        {densenet}
\bibfield{author}{\bibinfo{person}{Gao Huang}, \bibinfo{person}{Zhuang Liu}, \bibinfo{person}{Laurens van~der Maaten}, {and} \bibinfo{person}{Kilian~Q Weinberger}.} \bibinfo{year}{2017}\natexlab{}.
\newblock \showarticletitle{Densely Connected Convolutional Networks}. In \bibinfo{booktitle}{\emph{Proceedings of the IEEE Conference on Computer Vision and Pattern Recognition}}.
\newblock


\bibitem[Ilyas et~al\mbox{.}(2019)]%
        {aenotbug}
\bibfield{author}{\bibinfo{person}{Andrew Ilyas}, \bibinfo{person}{Shibani Santurkar}, \bibinfo{person}{Dimitris Tsipras}, \bibinfo{person}{Logan Engstrom}, \bibinfo{person}{Brandon Tran}, {and} \bibinfo{person}{Aleksander Madry}.} \bibinfo{year}{2019}\natexlab{}.
\newblock \showarticletitle{Adversarial Examples Are Not Bugs, They Are Features}. In \bibinfo{booktitle}{\emph{Advances in Neural Information Processing Systems}}.
\newblock


\bibitem[Jeyakumar et~al\mbox{.}(2020)]%
        {humanstudy}
\bibfield{author}{\bibinfo{person}{Jeya~Vikranth Jeyakumar}, \bibinfo{person}{Joseph Noor}, \bibinfo{person}{Yu-Hsi Cheng}, \bibinfo{person}{Luis Garcia}, {and} \bibinfo{person}{Mani Srivastava}.} \bibinfo{year}{2020}\natexlab{}.
\newblock \showarticletitle{How Can I Explain This to You? An Empirical Study of Deep Neural Network Explanation Methods}. In \bibinfo{booktitle}{\emph{Advances in Neural Information Processing Systems}}.
\newblock


\bibitem[Kokhlikyan et~al\mbox{.}(2020)]%
        {captum}
\bibfield{author}{\bibinfo{person}{Narine Kokhlikyan}, \bibinfo{person}{Vivek Miglani}, \bibinfo{person}{Miguel Martin}, \bibinfo{person}{Edward Wang}, \bibinfo{person}{Bilal Alsallakh}, \bibinfo{person}{Jonathan Reynolds}, \bibinfo{person}{Alexander Melnikov}, \bibinfo{person}{Natalia Kliushkina}, \bibinfo{person}{Carlos Araya}, \bibinfo{person}{Siqi Yan}, {and} \bibinfo{person}{Orion Reblitz-Richardson}.} \bibinfo{year}{2020}\natexlab{}.
\newblock \bibinfo{title}{Captum: A unified and generic model interpretability library for PyTorch}.
\newblock
\newblock
\showeprint[arxiv]{2009.07896}


\bibitem[Krizhevsky(2012)]%
        {cifar10}
\bibfield{author}{\bibinfo{person}{Alex Krizhevsky}.} \bibinfo{year}{2012}\natexlab{}.
\newblock \showarticletitle{Learning Multiple Layers of Features from Tiny Images}.
\newblock \bibinfo{journal}{\emph{University of Toronto}} (\bibinfo{date}{05} \bibinfo{year}{2012}).
\newblock


\bibitem[Lage et~al\mbox{.}(2018)]%
        {humanintheloop}
\bibfield{author}{\bibinfo{person}{Isaac Lage}, \bibinfo{person}{Andrew~Slavin Ross}, \bibinfo{person}{Been Kim}, \bibinfo{person}{Samuel~J. Gershman}, {and} \bibinfo{person}{Finale Doshi-Velez}.} \bibinfo{year}{2018}\natexlab{}.
\newblock \showarticletitle{Human-in-the-Loop Interpretability Prior}. In \bibinfo{booktitle}{\emph{Proceedings of the 32nd International Conference on Neural Information Processing Systems}} \emph{(\bibinfo{series}{NIPS'18})}.
\newblock


\bibitem[Lakkaraju et~al\mbox{.}(2016)]%
        {interpretabledecisionsets}
\bibfield{author}{\bibinfo{person}{Himabindu Lakkaraju}, \bibinfo{person}{Stephen~H. Bach}, {and} \bibinfo{person}{Jure Leskovec}.} \bibinfo{year}{2016}\natexlab{}.
\newblock \showarticletitle{Interpretable Decision Sets: A Joint Framework for Description and Prediction}. In \bibinfo{booktitle}{\emph{22nd ACM SIGKDD International Conference on Knowledge Discovery and Data Mining}} \emph{(\bibinfo{series}{KDD})}.
\newblock


\bibitem[Le and Yang(2015)]%
        {tinyimagenet}
\bibfield{author}{\bibinfo{person}{Ya Le} {and} \bibinfo{person}{Xuan Yang}.} \bibinfo{year}{2015}\natexlab{}.
\newblock \showarticletitle{Tiny ImageNet Visual Recognition Challenge}.
\newblock


\bibitem[Lecun et~al\mbox{.}(1998)]%
        {lenet5}
\bibfield{author}{\bibinfo{person}{Y. Lecun}, \bibinfo{person}{L. Bottou}, \bibinfo{person}{Y. Bengio}, {and} \bibinfo{person}{P. Haffner}.} \bibinfo{year}{1998}\natexlab{}.
\newblock \showarticletitle{Gradient-based learning applied to document recognition}.
\newblock \bibinfo{journal}{\emph{Proc. IEEE}} (\bibinfo{year}{1998}).
\newblock


\bibitem[Li et~al\mbox{.}(2021a)]%
        {segbackdoor}
\bibfield{author}{\bibinfo{person}{Yiming Li}, \bibinfo{person}{Yanjie Li}, \bibinfo{person}{Yalei Lv}, \bibinfo{person}{Yong Jiang}, {and} \bibinfo{person}{Shu{-}Tao Xia}.} \bibinfo{year}{2021}\natexlab{a}.
\newblock \showarticletitle{Hidden Backdoor Attack against Semantic Segmentation Models}.
\newblock  (\bibinfo{year}{2021}).
\newblock
\showeprint[arXiv]{2103.04038}


\bibitem[Li et~al\mbox{.}(2021b)]%
        {interpretvuldetectionmodels}
\bibfield{author}{\bibinfo{person}{Yi Li}, \bibinfo{person}{Shaohua Wang}, {and} \bibinfo{person}{Tien~N. Nguyen}.} \bibinfo{year}{2021}\natexlab{b}.
\newblock \showarticletitle{Vulnerability Detection with Fine-Grained Interpretations}. In \bibinfo{booktitle}{\emph{ACM Joint Meeting on European Software Engineering Conference and Symposium on the Foundations of Software Engineering}}.
\newblock
\showISBNx{9781450385626}


\bibitem[Li et~al\mbox{.}(2018)]%
        {Vuldeepecker}
\bibfield{author}{\bibinfo{person}{Zhen Li}, \bibinfo{person}{Deqing Zou}, \bibinfo{person}{Shouhuai Xu}, \bibinfo{person}{Xinyu Ou}, \bibinfo{person}{Hai Jin}, \bibinfo{person}{Sujuan Wang}, \bibinfo{person}{Zhijun Deng}, {and} \bibinfo{person}{Yuyi Zhong}.} \bibinfo{year}{2018}\natexlab{}.
\newblock \showarticletitle{VulDeePecker: {A} Deep Learning-Based System for Vulnerability Detection}. In \bibinfo{booktitle}{\emph{25th Annual Network and Distributed System Security Symposium, {NDSS}}}.
\newblock


\bibitem[Lin et~al\mbox{.}(2014)]%
        {MSCOCO}
\bibfield{author}{\bibinfo{person}{Tsung{-}Yi Lin}, \bibinfo{person}{Michael Maire}, \bibinfo{person}{Serge~J. Belongie}, \bibinfo{person}{Lubomir~D. Bourdev}, \bibinfo{person}{Ross~B. Girshick}, \bibinfo{person}{James Hays}, \bibinfo{person}{Pietro Perona}, \bibinfo{person}{Deva Ramanan}, \bibinfo{person}{Piotr Doll{\'{a}}r}, {and} \bibinfo{person}{C.~Lawrence Zitnick}.} \bibinfo{year}{2014}\natexlab{}.
\newblock \showarticletitle{Microsoft {COCO:} Common Objects in Context}.
\newblock  (\bibinfo{year}{2014}).
\newblock


\bibitem[Long et~al\mbox{.}(2015)]%
        {FCN}
\bibfield{author}{\bibinfo{person}{J. Long}, \bibinfo{person}{E. Shelhamer}, {and} \bibinfo{person}{T. Darrell}.} \bibinfo{year}{2015}\natexlab{}.
\newblock \showarticletitle{Fully convolutional networks for semantic segmentation}. In \bibinfo{booktitle}{\emph{IEEE Conference on Computer Vision and Pattern Recognition (CVPR)}}.
\newblock
\urldef\tempurl%
\url{https://doi.org/10.1109/CVPR.2015.7298965}
\showDOI{\tempurl}


\bibitem[Lundberg and Lee(2017)]%
        {SHAP}
\bibfield{author}{\bibinfo{person}{Scott~M. Lundberg} {and} \bibinfo{person}{Su-In Lee}.} \bibinfo{year}{2017}\natexlab{}.
\newblock \showarticletitle{A Unified Approach to Interpreting Model Predictions}. In \bibinfo{booktitle}{\emph{Proceedings of the 31st International Conference on Neural Information Processing Systems}} \emph{(\bibinfo{series}{NIPS'17})}.
\newblock
\showISBNx{9781510860964}


\bibitem[Maas et~al\mbox{.}(2011)]%
        {IMDB}
\bibfield{author}{\bibinfo{person}{Andrew~L. Maas}, \bibinfo{person}{Raymond~E. Daly}, \bibinfo{person}{Peter~T. Pham}, \bibinfo{person}{Dan Huang}, \bibinfo{person}{Andrew~Y. Ng}, {and} \bibinfo{person}{Christopher Potts}.} \bibinfo{year}{2011}\natexlab{}.
\newblock \showarticletitle{Learning Word Vectors for Sentiment Analysis}. In \bibinfo{booktitle}{\emph{Association for Computational Linguistics: Human Language Technologies}}.
\newblock
\showISBNx{9781932432879}


\bibitem[McLaughlin et~al\mbox{.}(2017)]%
        {DAMD}
\bibfield{author}{\bibinfo{person}{Niall McLaughlin}, \bibinfo{person}{Jesus Martinez~del Rincon}, \bibinfo{person}{BooJoong Kang}, \bibinfo{person}{Suleiman Yerima}, \bibinfo{person}{Paul Miller}, \bibinfo{person}{Sakir Sezer}, \bibinfo{person}{Yeganeh Safaei}, \bibinfo{person}{Erik Trickel}, \bibinfo{person}{Ziming Zhao}, \bibinfo{person}{Adam Doup\'{e}}, {and} \bibinfo{person}{Gail Joon~Ahn}.} \bibinfo{year}{2017}\natexlab{}.
\newblock \showarticletitle{Deep Android Malware Detection}. In \bibinfo{booktitle}{\emph{Proceedings of the Seventh ACM on Conference on Data and Application Security and Privacy}}.
\newblock
\showISBNx{9781450345231}


\bibitem[Nagarajan et~al\mbox{.}(2021)]%
        {defood}
\bibfield{author}{\bibinfo{person}{Vaishnavh Nagarajan}, \bibinfo{person}{Anders Andreassen}, {and} \bibinfo{person}{Behnam Neyshabur}.} \bibinfo{year}{2021}\natexlab{}.
\newblock \showarticletitle{Understanding the failure modes of out-of-distribution generalization}. In \bibinfo{booktitle}{\emph{International Conference on Learning Representations}}.
\newblock


\bibitem[Narayanan et~al\mbox{.}(2018)]%
        {humanaccuracy}
\bibfield{author}{\bibinfo{person}{Menaka Narayanan}, \bibinfo{person}{Emily Chen}, \bibinfo{person}{Jeffrey He}, \bibinfo{person}{Been Kim}, \bibinfo{person}{Sam Gershman}, {and} \bibinfo{person}{Finale Doshi{-}Velez}.} \bibinfo{year}{2018}\natexlab{}.
\newblock \showarticletitle{How do Humans Understand Explanations from Machine Learning Systems? An Evaluation of the Human-Interpretability of Explanation}.
\newblock \bibinfo{journal}{\emph{CoRR}} (\bibinfo{year}{2018}).
\newblock


\bibitem[Paszke et~al\mbox{.}(2019)]%
        {PYTORCH}
\bibfield{author}{\bibinfo{person}{Adam Paszke}, \bibinfo{person}{Sam Gross}, \bibinfo{person}{Francisco Massa}, \bibinfo{person}{Adam Lerer}, \bibinfo{person}{James Bradbury}, \bibinfo{person}{Gregory Chanan}, \bibinfo{person}{Trevor Killeen}, \bibinfo{person}{Zeming Lin}, \bibinfo{person}{Natalia Gimelshein}, \bibinfo{person}{Luca Antiga}, \bibinfo{person}{Alban Desmaison}, \bibinfo{person}{Andreas Kopf}, \bibinfo{person}{Edward Yang}, \bibinfo{person}{Zachary DeVito}, \bibinfo{person}{Martin Raison}, \bibinfo{person}{Alykhan Tejani}, \bibinfo{person}{Sasank Chilamkurthy}, \bibinfo{person}{Benoit Steiner}, \bibinfo{person}{Lu Fang}, \bibinfo{person}{Junjie Bai}, {and} \bibinfo{person}{Soumith Chintala}.} \bibinfo{year}{2019}\natexlab{}.
\newblock \showarticletitle{PyTorch: An Imperative Style, High-Performance Deep Learning Library}. In \bibinfo{booktitle}{\emph{Advances in Neural Information Processing Systems}}.
\newblock
\urldef\tempurl%
\url{https://proceedings.neurips.cc/paper/2019/file/bdbca288fee7f92f2bfa9f7012727740-Paper.pdf}
\showURL{%
\tempurl}


\bibitem[Ren et~al\mbox{.}(2021)]%
        {explanationadversarialrobustness}
\bibfield{author}{\bibinfo{person}{Jie Ren}, \bibinfo{person}{Die Zhang}, \bibinfo{person}{Yisen Wang}, \bibinfo{person}{Lu Chen}, \bibinfo{person}{Zhanpeng Zhou}, \bibinfo{person}{Yiting Chen}, \bibinfo{person}{Xu Cheng}, \bibinfo{person}{Xin Wang}, \bibinfo{person}{Meng Zhou}, \bibinfo{person}{Jie Shi}, {and} \bibinfo{person}{Quanshi Zhang}.} \bibinfo{year}{2021}\natexlab{}.
\newblock \showarticletitle{Towards a Unified Game-Theoretic View of Adversarial Perturbations and Robustness}. In \bibinfo{booktitle}{\emph{Advances in Neural Information Processing Systems}}.
\newblock


\bibitem[Ribeiro et~al\mbox{.}(2016)]%
        {LIME}
\bibfield{author}{\bibinfo{person}{Marco~Tulio Ribeiro}, \bibinfo{person}{Sameer Singh}, {and} \bibinfo{person}{Carlos Guestrin}.} \bibinfo{year}{2016}\natexlab{}.
\newblock \showarticletitle{"Why Should I Trust You?": Explaining the Predictions of Any Classifier}. In \bibinfo{booktitle}{\emph{22nd ACM SIGKDD International Conference on Knowledge Discovery and Data Mining}}.
\newblock
\showISBNx{9781450342322}


\bibitem[Ribeiro et~al\mbox{.}(2018)]%
        {anchor}
\bibfield{author}{\bibinfo{person}{Marco~Tulio Ribeiro}, \bibinfo{person}{Sameer Singh}, {and} \bibinfo{person}{Carlos Guestrin}.} \bibinfo{year}{2018}\natexlab{}.
\newblock \showarticletitle{Anchors: High-Precision Model-Agnostic Explanations}.
\newblock \bibinfo{journal}{\emph{Proceedings of the AAAI Conference on Artificial Intelligence}} \bibinfo{volume}{32}, \bibinfo{number}{1} (\bibinfo{date}{Apr.} \bibinfo{year}{2018}).
\newblock
\urldef\tempurl%
\url{https://doi.org/10.1609/aaai.v32i1.11491}
\showDOI{\tempurl}


\bibitem[Saxe and Berlin(2015)]%
        {PDFMalwareClassifier}
\bibfield{author}{\bibinfo{person}{Joshua Saxe} {and} \bibinfo{person}{Konstantin Berlin}.} \bibinfo{year}{2015}\natexlab{}.
\newblock \showarticletitle{Deep neural network based malware detection using two dimensional binary program features}. In \bibinfo{booktitle}{\emph{2015 10th International Conference on Malicious and Unwanted Software (MALWARE)}}.
\newblock


\bibitem[Schuster and Paliwal(1997)]%
        {birnn}
\bibfield{author}{\bibinfo{person}{M. Schuster} {and} \bibinfo{person}{K.K. Paliwal}.} \bibinfo{year}{1997}\natexlab{}.
\newblock \showarticletitle{Bidirectional recurrent neural networks}.
\newblock \bibinfo{journal}{\emph{IEEE Transactions on Signal Processing}} (\bibinfo{year}{1997}).
\newblock
\urldef\tempurl%
\url{https://doi.org/10.1109/78.650093}
\showDOI{\tempurl}


\bibitem[Severi et~al\mbox{.}(2021)]%
        {explanationpoison}
\bibfield{author}{\bibinfo{person}{Giorgio Severi}, \bibinfo{person}{Jim Meyer}, \bibinfo{person}{Scott Coull}, {and} \bibinfo{person}{Alina Oprea}.} \bibinfo{year}{2021}\natexlab{}.
\newblock \showarticletitle{{Explanation-Guided} Backdoor Poisoning Attacks Against Malware Classifiers}. In \bibinfo{booktitle}{\emph{30th USENIX Security Symposium}}. \bibinfo{publisher}{USENIX Association}, \bibinfo{pages}{1487--1504}.
\newblock
\showISBNx{978-1-939133-24-3}


\bibitem[Shrikumar et~al\mbox{.}(2017)]%
        {DeepLIFT}
\bibfield{author}{\bibinfo{person}{Avanti Shrikumar}, \bibinfo{person}{Peyton Greenside}, {and} \bibinfo{person}{Anshul Kundaje}.} \bibinfo{year}{2017}\natexlab{}.
\newblock \showarticletitle{Learning Important Features through Propagating Activation Differences}. In \bibinfo{booktitle}{\emph{Proceedings of the 34th International Conference on Machine Learning - Volume 70}}.
\newblock


\bibitem[Simonyan et~al\mbox{.}(2014)]%
        {Grad}
\bibfield{author}{\bibinfo{person}{Karen Simonyan}, \bibinfo{person}{Andrea Vedaldi}, {and} \bibinfo{person}{Andrew Zisserman}.} \bibinfo{year}{2014}\natexlab{}.
\newblock \showarticletitle{Deep Inside Convolutional Networks: Visualising Image Classification Models and Saliency Maps}. In \bibinfo{booktitle}{\emph{2nd International Conference on Learning Representations, {ICLR}}}.
\newblock


\bibitem[Smilkov et~al\mbox{.}(2017)]%
        {SmoothGrad}
\bibfield{author}{\bibinfo{person}{Daniel Smilkov}, \bibinfo{person}{Nikhil Thorat}, \bibinfo{person}{Been Kim}, \bibinfo{person}{Fernanda~B. Vi{\'{e}}gas}, {and} \bibinfo{person}{Martin Wattenberg}.} \bibinfo{year}{2017}\natexlab{}.
\newblock \showarticletitle{SmoothGrad: removing noise by adding noise}.
\newblock \bibinfo{journal}{\emph{CoRR}} (\bibinfo{year}{2017}).
\newblock


\bibitem[Smutz and Stavrou(2012)]%
        {MimicusDataset}
\bibfield{author}{\bibinfo{person}{Charles Smutz} {and} \bibinfo{person}{Angelos Stavrou}.} \bibinfo{year}{2012}\natexlab{}.
\newblock \showarticletitle{Malicious PDF Detection Using Metadata and Structural Features}. In \bibinfo{booktitle}{\emph{Proceedings of the 28th Annual Computer Security Applications Conference}} \emph{(\bibinfo{series}{ACSAC '12})}.
\newblock
\urldef\tempurl%
\url{https://doi.org/10.1145/2420950.2420987}
\showDOI{\tempurl}


\bibitem[Sundararajan et~al\mbox{.}(2017)]%
        {IG}
\bibfield{author}{\bibinfo{person}{Mukund Sundararajan}, \bibinfo{person}{Ankur Taly}, {and} \bibinfo{person}{Qiqi Yan}.} \bibinfo{year}{2017}\natexlab{}.
\newblock \showarticletitle{Axiomatic Attribution for Deep Networks}. In \bibinfo{booktitle}{\emph{Proceedings of the 34th International Conference on Machine Learning}}.
\newblock
\urldef\tempurl%
\url{http://proceedings.mlr.press/v70/sundararajan17a.html}
\showURL{%
\tempurl}


\bibitem[Sutskever et~al\mbox{.}(2014)]%
        {seq2seq}
\bibfield{author}{\bibinfo{person}{Ilya Sutskever}, \bibinfo{person}{Oriol Vinyals}, {and} \bibinfo{person}{Quoc~V. Le}.} \bibinfo{year}{2014}\natexlab{}.
\newblock \showarticletitle{Sequence to Sequence Learning with Neural Networks}. In \bibinfo{booktitle}{\emph{Proceedings of the 27th International Conference on Neural Information Processing Systems}} (Montreal, Canada). \bibinfo{pages}{3104–3112}.
\newblock


\bibitem[Vaswani et~al\mbox{.}(2017)]%
        {transformer}
\bibfield{author}{\bibinfo{person}{Ashish Vaswani}, \bibinfo{person}{Noam Shazeer}, \bibinfo{person}{Niki Parmar}, \bibinfo{person}{Jakob Uszkoreit}, \bibinfo{person}{Llion Jones}, \bibinfo{person}{Aidan~N. Gomez}, \bibinfo{person}{\L{}ukasz Kaiser}, {and} \bibinfo{person}{Illia Polosukhin}.} \bibinfo{year}{2017}\natexlab{}.
\newblock \showarticletitle{Attention is All You Need}. In \bibinfo{booktitle}{\emph{International Conference on Neural Information Processing Systems}}.
\newblock


\bibitem[Wang et~al\mbox{.}(2019)]%
        {neuralcleanse}
\bibfield{author}{\bibinfo{person}{Bolun Wang}, \bibinfo{person}{Yuanshun Yao}, \bibinfo{person}{Shawn Shan}, \bibinfo{person}{Huiying Li}, \bibinfo{person}{Bimal Viswanath}, \bibinfo{person}{Haitao Zheng}, {and} \bibinfo{person}{Ben~Y. Zhao}.} \bibinfo{year}{2019}\natexlab{}.
\newblock \showarticletitle{Neural Cleanse: Identifying and Mitigating Backdoor Attacks in Neural Networks}. In \bibinfo{booktitle}{\emph{IEEE Symposium on Security and Privacy}}.
\newblock


\bibitem[Wang et~al\mbox{.}(2021)]%
        {interpretadversarialattack}
\bibfield{author}{\bibinfo{person}{Xin Wang}, \bibinfo{person}{Shuyun Lin}, \bibinfo{person}{Hao Zhang}, \bibinfo{person}{Yufei Zhu}, {and} \bibinfo{person}{Quanshi Zhang}.} \bibinfo{year}{2021}\natexlab{}.
\newblock \showarticletitle{Interpreting Attributions and Interactions of Adversarial Attacks}. In \bibinfo{booktitle}{\emph{{IEEE/CVF} International Conference on Computer Vision, {ICCV}}}.
\newblock


\bibitem[Wang et~al\mbox{.}(2004)]%
        {ssim}
\bibfield{author}{\bibinfo{person}{Zhou Wang}, \bibinfo{person}{A.C. Bovik}, \bibinfo{person}{H.R. Sheikh}, {and} \bibinfo{person}{E.P. Simoncelli}.} \bibinfo{year}{2004}\natexlab{}.
\newblock \showarticletitle{Image quality assessment: from error visibility to structural similarity}.
\newblock \bibinfo{journal}{\emph{IEEE Transactions on Image Processing}} \bibinfo{volume}{13}, \bibinfo{number}{4} (\bibinfo{year}{2004}), \bibinfo{pages}{600--612}.
\newblock


\bibitem[{Warnecke} et~al\mbox{.}(2020)]%
        {EuroSPSurvey}
\bibfield{author}{\bibinfo{person}{A. {Warnecke}}, \bibinfo{person}{D. {Arp}}, \bibinfo{person}{C. {Wressnegger}}, {and} \bibinfo{person}{K. {Rieck}}.} \bibinfo{year}{2020}\natexlab{}.
\newblock \showarticletitle{Evaluating Explanation Methods for Deep Learning in Security}. In \bibinfo{booktitle}{\emph{IEEE European Symposium on Security and Privacy (EuroS\&P)}}.
\newblock


\bibitem[Xiao et~al\mbox{.}(2021)]%
        {noiseorsignal}
\bibfield{author}{\bibinfo{person}{Kai~Yuanqing Xiao}, \bibinfo{person}{Logan Engstrom}, \bibinfo{person}{Andrew Ilyas}, {and} \bibinfo{person}{Aleksander Madry}.} \bibinfo{year}{2021}\natexlab{}.
\newblock \showarticletitle{Noise or Signal: The Role of Image Backgrounds in Object Recognition}. In \bibinfo{booktitle}{\emph{International Conference on Learning Representations}}.
\newblock


\bibitem[Yang and Kim(2019)]%
        {BAM}
\bibfield{author}{\bibinfo{person}{Mengjiao Yang} {and} \bibinfo{person}{Been Kim}.} \bibinfo{year}{2019}\natexlab{}.
\newblock \showarticletitle{{Benchmarking Attribution Methods with Relative Feature Importance}}.
\newblock \bibinfo{journal}{\emph{CoRR}} (\bibinfo{year}{2019}).
\newblock


\bibitem[Zeiler and Fergus(2014)]%
        {Occlusion}
\bibfield{author}{\bibinfo{person}{Matthew~D. Zeiler} {and} \bibinfo{person}{Rob Fergus}.} \bibinfo{year}{2014}\natexlab{}.
\newblock \showarticletitle{Visualizing and understanding convolutional networks}. In \bibinfo{booktitle}{\emph{In Computer Vision–ECCV 2014}}.
\newblock


\bibitem[Zhang et~al\mbox{.}(2019)]%
        {pccdl2}
\bibfield{author}{\bibinfo{person}{Fan Zhang}, \bibinfo{person}{Zhenzhen Li}, \bibinfo{person}{Boyan Zhang}, \bibinfo{person}{Haishun Du}, \bibinfo{person}{Binjie Wang}, {and} \bibinfo{person}{Xinhong Zhang}.} \bibinfo{year}{2019}\natexlab{}.
\newblock \showarticletitle{Multi-modal deep learning model for auxiliary diagnosis of Alzheimer’s disease}.
\newblock \bibinfo{journal}{\emph{Neurocomputing}}  \bibinfo{volume}{361} (\bibinfo{year}{2019}), \bibinfo{pages}{185--195}.
\newblock


\bibitem[Zhang et~al\mbox{.}(2021)]%
        {DSL}
\bibfield{author}{\bibinfo{person}{Xingxuan Zhang}, \bibinfo{person}{Peng Cui}, \bibinfo{person}{Renzhe Xu}, \bibinfo{person}{Linjun Zhou}, \bibinfo{person}{Yue He}, {and} \bibinfo{person}{Zheyan Shen}.} \bibinfo{year}{2021}\natexlab{}.
\newblock \showarticletitle{Deep Stable Learning for Out-of-Distribution Generalization}. In \bibinfo{booktitle}{\emph{Proceedings of the IEEE/CVF Conference on Computer Vision and Pattern Recognition (CVPR)}}.
\newblock


\bibitem[Zhang et~al\mbox{.}(2020)]%
        {attackexplanation}
\bibfield{author}{\bibinfo{person}{Xinyang Zhang}, \bibinfo{person}{Ningfei Wang}, \bibinfo{person}{Hua Shen}, \bibinfo{person}{Shouling Ji}, \bibinfo{person}{Xiapu Luo}, {and} \bibinfo{person}{Ting Wang}.} \bibinfo{year}{2020}\natexlab{}.
\newblock \showarticletitle{Interpretable Deep Learning under Fire}. In \bibinfo{booktitle}{\emph{29th USENIX Security Symposium (USENIX Security 20)}}. \bibinfo{publisher}{USENIX Association}, \bibinfo{pages}{1659--1676}.
\newblock
\showISBNx{978-1-939133-17-5}


\bibitem[Zhou et~al\mbox{.}(2017)]%
        {miniplaces}
\bibfield{author}{\bibinfo{person}{Bolei Zhou}, \bibinfo{person}{Agata Lapedriza}, \bibinfo{person}{Aditya Khosla}, \bibinfo{person}{Aude Oliva}, {and} \bibinfo{person}{Antonio Torralba}.} \bibinfo{year}{2017}\natexlab{}.
\newblock \showarticletitle{Places: A 10 million Image Database for Scene Recognition}.
\newblock \bibinfo{journal}{\emph{IEEE Transactions on Pattern Analysis and Machine Intelligence}} (\bibinfo{year}{2017}).
\newblock


\bibitem[Zhou and Jiang(2012)]%
        {DAMDSOURCE}
\bibfield{author}{\bibinfo{person}{Yajin Zhou} {and} \bibinfo{person}{Xuxian Jiang}.} \bibinfo{year}{2012}\natexlab{}.
\newblock \showarticletitle{Dissecting Android Malware: Characterization and Evolution}. In \bibinfo{booktitle}{\emph{2012 IEEE Symposium on Security and Privacy}}. \bibinfo{pages}{95--109}.
\newblock
\urldef\tempurl%
\url{https://doi.org/10.1109/SP.2012.16}
\showDOI{\tempurl}


\bibitem[Zhou et~al\mbox{.}(2022)]%
        {robustcodegeneration}
\bibfield{author}{\bibinfo{person}{Yu Zhou}, \bibinfo{person}{Xiaoqing Zhang}, \bibinfo{person}{Juanjuan Shen}, \bibinfo{person}{Tingting Han}, \bibinfo{person}{Taolue Chen}, {and} \bibinfo{person}{Harald Gall}.} \bibinfo{year}{2022}\natexlab{}.
\newblock \showarticletitle{Adversarial Robustness of Deep Code Comment Generation}.
\newblock \bibinfo{journal}{\emph{ACM Trans. Softw. Eng. Methodol.}} \bibinfo{volume}{31}, \bibinfo{number}{4}, Article \bibinfo{articleno}{60} (\bibinfo{date}{jul} \bibinfo{year}{2022}), \bibinfo{numpages}{30}~pages.
\newblock
\showISSN{1049-331X}


\bibitem[Zou et~al\mbox{.}(2021)]%
        {interpretvulnerabilitydetection}
\bibfield{author}{\bibinfo{person}{Deqing Zou}, \bibinfo{person}{Yawei Zhu}, \bibinfo{person}{Shouhuai Xu}, \bibinfo{person}{Zhen Li}, \bibinfo{person}{Hai Jin}, {and} \bibinfo{person}{Hengkai Ye}.} \bibinfo{year}{2021}\natexlab{}.
\newblock \showarticletitle{Interpreting Deep Learning-Based Vulnerability Detector Predictions Based on Heuristic Searching}.
\newblock \bibinfo{journal}{\emph{ACM Trans. Softw. Eng. Methodol.}} \bibinfo{volume}{30}, \bibinfo{number}{2} (\bibinfo{year}{2021}).
\newblock
\showISSN{1049-331X}


\end{thebibliography}
